\documentclass{article}

\usepackage{iclr2027_conference,times}
\iclrfinalcopy

\newlength{\runinheadingspace}
\usepackage[utf8]{inputenc}
\usepackage[T1]{fontenc}
\usepackage{url}
\usepackage{booktabs}
\usepackage{amsfonts}
\usepackage{nicefrac}
\usepackage{microtype}
\usepackage{xcolor}
\definecolor{darkorange}{RGB}{190,95,0}
\usepackage{graphicx}
\graphicspath{{./}}
\usepackage{subcaption}
\usepackage{adjustbox}
\usepackage{comment}
\usepackage{pdflscape}
\usepackage{array}
\usepackage{enumitem}
\usepackage{verbatim}
\usepackage{amsmath}
\usepackage{amssymb}
\usepackage{mathrsfs}
\usepackage{mathtools}
\usepackage{amsthm}
\usepackage{algorithm}
\usepackage{algorithmic}
\usepackage{float}
\usepackage{wrapfig}
\usepackage{multirow}
\usepackage{tabularx}
\usepackage{makecell}
\usepackage{tocloft}
\usepackage{etoc}
\usepackage{appendix}
\usepackage{hyperref}
\hypersetup{hidelinks}

\newcommand{\Dkl}{D_{\mathrm{KL}}}

\newcommand{\Tau}{\mathcal{T}}
\newcommand{\dappo}{DA:~PPO}
\newcommand{\dareppo}{DA:~REPPO}
\newcommand{\dawpo}{DA:~WPO}

\newcommand{\thetar}{\theta_{\mathrm{roll}}}
\newcommand{\thetastop}{\theta^*}
\DeclareRobustCommand{\trudi}{TruDi}
\DeclareRobustCommand{\replaceReppoDime}{\trudi}

\usepackage[capitalize,noabbrev]{cleveref}

\theoremstyle{plain}
\newtheorem{theorem}{Theorem}[section]
\newtheorem{proposition}[theorem]{Proposition}
\newtheorem{lemma}[theorem]{Lemma}

\theoremstyle{definition}

\theoremstyle{remark}

\usepackage[textsize=tiny]{todonotes}

\newcommand{\eqsizesmall}{\fontsize{8pt}{10pt}\selectfont}
\newcommand{\eqsizetiny}{\fontsize{7pt}{8pt}\selectfont}

\title{Diffusion-Augmented Markov Decision \\Processes for Maximum Entropy \\Reinforcement Learning}

\author{
Sebastian Sanokowski$^{1}$ \qquad
Kaustubh Patil$^{2,3}$ \qquad
Majid Khadiv$^{1}$ \\[1ex]
$^{1}$Applied and Theoretical Aspects of Robot Intelligence (ATARI) Lab, \\
Munich Institute of Robotics and Machine Intelligence (MIRMI), \\
Technical University of Munich \\
$^{2}$Practical Project Student Exchange Program, Technical University of Munich \\
$^{3}$MIT World Peace University \\
\texttt{sebastian.sanokowski[at]tum.de}
}

\begin{document}
\etocdepthtag.toc{mainmatter}

\maketitle
\lhead{Under review}

\begin{abstract}
Diffusion models provide an expressive framework for sampling from complex, unnormalized distributions. In this work, we extend Maximum Entropy Reinforcement Learning (ME-RL) to diffusion-based policies by introducing \emph{Diffusion-Augmented Markov Decision Processes} (DA-MDPs). DA-MDPs interpret each reverse-diffusion transition as an individual reinforcement-learning decision, while only the final denoised action is executed in the environment. Our DA-MDPs follow from a principled derivation based on the variational-inference formulation of ME-RL. By augmenting policy and target trajectories with intermediate diffusion variables, we obtain a tractable reverse-KL upper bound via the data-processing inequality. This bound decomposes across denoising transitions, yielding diffusion-augmented variants of soft rewards, value functions, and local policy objectives. This provides a general framework for adapting ME-RL algorithms to diffusion policies while differentiating through only one diffusion transition at a time. We instantiate the framework with PPO, REPPO, and a maximum-entropy extension of WPO.
Experiments demonstrate improved continuous-control performance, benefits from additional diffusion steps, and memory-efficient training. On the StackCube and PushT manipulation tasks, DA-MDP methods learn alternative successful strategies from the same initial state and achieve higher success rates and generally higher success-weighted mode entropy than the Gaussian ME-RL baseline. We also demonstrate successful training when using action chunking.
Our implementations are publicly available in JAX\footnote{JAX code: \urlstyle{same}\url{https://github.com/sanokows/Diffusion-Augmented-Markov-Decision-Processes}.} and PyTorch\footnote{PyTorch code: \urlstyle{same}\url{https://github.com/Atarilab/DA_MDP_pytorch}.}.
\end{abstract}

\begin{figure}[!t]
    \centering
    \begin{subfigure}{\textwidth}
        \centering
        \includegraphics[width=\linewidth]{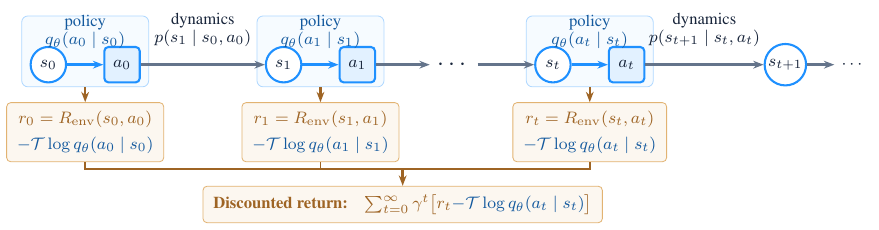}
        \caption{An ordinary maximum-entropy MDP: one policy sample per environment step.}
        \label{fig:ordinary_mdp_overview}
    \end{subfigure}
    \par\medskip
    \begin{subfigure}{\textwidth}
        \centering
        \includegraphics[width=\linewidth]{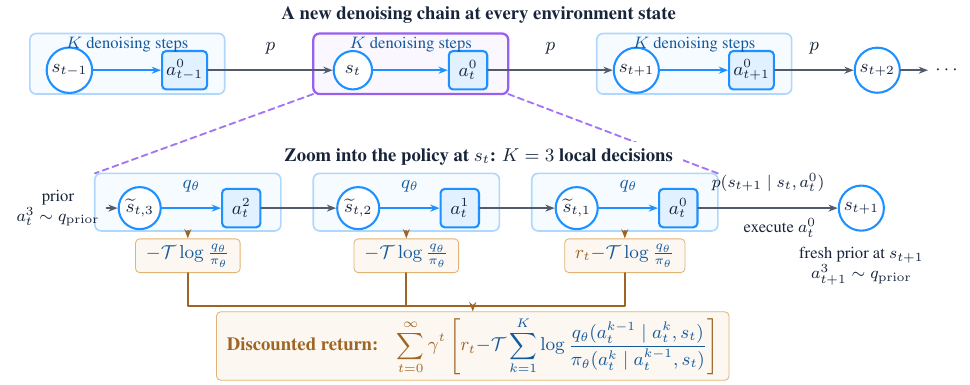}
        \caption{A DA-MDP across environment steps, with a zoom into $K=3$ local decisions at $s_t$.}
        \label{fig:denoising_mdp_overview}
    \end{subfigure}
    \caption{From environment trajectories to local denoising decisions. (a) Each action produces an environment reward and an entropy contribution before the environment advances. (b) A fresh denoising chain generates an action at each environment state. Dashed purple guides expand the highlighted policy call at $s_t$ into local DA-MDP decisions. Gray arrows between denoising decisions carry the sampled action into the next augmented state; only the final action $a_t^0$ changes the environment state. Gold boxes show the local contributions and their infinite-horizon discounted returns.}
    \label{fig:da_mdp_overview}
\end{figure}

\section{Introduction}

Diffusion models have emerged as powerful generative models for complex distributions \citep{DicksteinDiff, DenoisingDiffusionModels, song2021maximum}. Diffusion samplers extend this idea to distributions known only through an unnormalized density: they learn a reverse process that transports samples from a simple prior toward high-density regions of the target \citep{PathIntegralSampler, berner2022optimal, DDS, vargas2024transport, richter2023Imporved, sanokowski2024diffusion, Sanokowski2025scalable}. Maximum-entropy reinforcement learning (ME-RL) naturally fits this setting because it can be viewed as sampling trajectories from an unnormalized distribution that favors high cumulative reward \citep{todorov2008general, ziebart2010modeling, kappen2012optimal, levine2018reinforcement}. Diffusion policies are therefore a natural choice for ME-RL: they provide a flexible representation of complex, multimodal action distributions \citep{ren2025diffusion, celik2025dime, dong2025maximum, ma2025reinforcement, mcallister2025flow}.
The difficulty, however, in using diffusion models in ME-RL is the computation of the likelihood of the action executed in the environment. For a diffusion policy, obtaining this likelihood requires marginalizing all intermediate diffusion variables and is generally intractable. Existing full-path objectives avoid this marginalization, but require backpropagation through the complete diffusion chain \citep{celik2025dime, le2026trust}.

We introduce \textbf{Diffusion-Augmented Markov Decision Processes (DA-MDPs)} to address these challenges by treating each reverse-diffusion transition as an individual RL decision. This formulation replaces the intractable marginal action likelihood with tractable transition likelihoods and requires backpropagation through only one diffusion step at a time. At each denoising step, the policy samples a partially denoised action conditioned on the current noisy action and the environment state. Only the final denoised action is executed in the environment.

Each denoising transition receives a reward based on the log ratio of the forward and reverse transition densities. This reward encourages consistency with the prescribed noising process while promoting entropy maximization at each reverse step. To derive it, we augment the policy and target trajectories with intermediate diffusion variables and apply the data-processing inequality. The resulting upper bound on the reverse KL decomposes into local critic and actor objectives.
Figure~\ref{fig:da_mdp_overview} illustrates the construction; App.~\ref{app:graphical_overview} explains the diagram in detail.

\textbf{Our main contributions are as follows:}

\begin{itemize}[nolistsep, itemsep=0.5ex, leftmargin=*]
    \item We derive \textbf{DA-MDPs} from a reverse KL between augmented trajectory distributions, obtaining a principled local reward for each diffusion transition.
    \item We extend TRPO's monotonic-improvement bound \citep{schulman2015trust} to DA-MDPs, giving sufficient conditions for improved augmented return, including when forward kernels are learned.
    \item We instantiate \textbf{\dappo{}}, \textbf{\dareppo{}}, and \textbf{\dawpo{}} from PPO \citep{schulman_proximal_2017}, REPPO \citep{voelcker2025reppo}, and our maximum-entropy extension of WPO \citep{pfau2025wasserstein}.
    \item Across twelve continuous-control tasks, the DA variants improve on their corresponding base algorithms. On StackCube, DA policies learn both stacking orders, including when executing action chunks. On PushT, \dareppo{} maintains successful clockwise and counter-clockwise strategies near the direction-switching boundary, where stochastic Gaussian REPPO rarely succeeds. Further experiments demonstrate benefits from additional diffusion steps and faster training than a full-chain baseline under tight memory budgets.
\end{itemize}

\section{Setup and Background}
\label{sec:problem_description}

We first review the two ingredients required for our construction: the probabilistic-inference formulation of ME-RL and discrete-time diffusion policies. A summary of the notation used throughout the paper is provided in Table~\ref{tab:notation} in App.~\ref{app:notation}.

\subsection{Maximum-Entropy Reinforcement Learning as Variational Inference}
\label{sec:maxentRL}

Reinforcement learning can be interpreted as an inference problem in which the objective is to sample trajectories with large cumulative reward. Consider a finite-horizon MDP with initial-state distribution $p(s_0)$, transition dynamics $p(s_{t+1}\mid s_t,a_t)$, and reward function $R_{\mathrm{env}}(s_t,a_t)$. Here, $s_t=(x_t,t)\in\mathcal S$ denotes the time-dependent state, $a_t\in\mathcal A\subset\mathbb R^N$ denotes the action, where $\mathcal A$ is compact, and $T$ is the final environment-time index. We further assume that the reward is uniformly bounded above: $R_{\mathrm{env}}(s,a)\leq R_{\max}<\infty$. 

\begingroup
The unnormalized target over action trajectories $\widetilde{\pi}(a_{0:T})$ is obtained by marginalizing the unnormalized joint state--action target $\widetilde\pi(a_{0:T},s_{0:T+1})$:
\[
\begin{gathered}
\widetilde{\pi}(a_{0:T})
:=\int_{\mathcal S^{T+2}}
\widetilde{\pi}(a_{0:T},s_{0:T+1})\,d s_{0:T+1},
\quad
\pi(a_{0:T}):=\frac{\widetilde{\pi}(a_{0:T})}{\mathcal Z},\\
\widetilde{\pi}(a_{0:T},s_{0:T+1})
:=p(s_0)\prod_{t=0}^{T}
p(s_{t+1}\mid s_t,a_t)
\exp\!\left(\alpha R_{\mathrm{env}}(s_t,a_t)\right).
\end{gathered}
\]
Here, $\alpha:=1/\Tau>0$ is the inverse of the entropy temperature $\Tau>0$. The temperature $\Tau$ controls the trade-off between reward maximization and entropy: larger values place greater weight on entropy, whereas smaller values concentrate the target distribution on high-reward trajectories. A tilde denotes an unnormalized density, whereas the corresponding symbol without a tilde denotes the normalized distribution. The partition function is $\mathcal Z:=\int_{\mathcal A^{T+1}}\widetilde{\pi}(a_{0:T})\,d a_{0:T}$, and the compact-action and bounded-reward assumptions imply $0<\mathcal Z<\infty$ for every fixed finite horizon; the same $\mathcal Z$ normalizes the joint target. Because
$\prod_{t=0}^{T}\exp(\alpha R_{\mathrm{env}}(s_t,a_t))=\exp(\alpha\sum_{t=0}^{T}R_{\mathrm{env}}(s_t,a_t))$, trajectories with larger cumulative reward receive exponentially more probability mass.

A policy $q_\theta(a_t\mid s_t)$ analogously induces the action-trajectory marginal and the corresponding joint variational distribution
{\small
\[
q_\theta(a_{0:T})
:=\int_{\mathcal S^{T+2}}
q_\theta(a_{0:T},s_{0:T+1})\,d s_{0:T+1},
\quad
q_\theta(a_{0:T},s_{0:T+1})
:=p(s_0)\prod_{t=0}^{T}
p(s_{t+1}\mid s_t,a_t)q_\theta(a_t\mid s_t).
\]
}
In principle, variational inference would minimize the reverse KL divergence between $q_\theta(a_{0:T})$ and $\pi(a_{0:T})$. Evaluating this divergence is generally intractable, however, because both marginal densities require integration over all state trajectories. Applying the data-processing inequality (DPI) to this marginalization gives
\begin{equation}
D_{\mathrm{KL}}\!\left(q_\theta(a_{0:T})\,\middle\|\,\pi(a_{0:T})\right)
\leq
D_{\mathrm{KL}}\!\left(q_\theta(a_{0:T},s_{0:T+1})\,\middle\|\,
\pi(a_{0:T},s_{0:T+1})\right),
\label{eq:rl_DPI}
\end{equation}
and we can therefore minimize the tractable joint reverse KL on the right-hand side as an upper bound on the desired action-marginal objective. Its zero-divergence optimality condition is $q_{\theta^\star}(a_{0:T},s_{0:T+1})=\pi(a_{0:T},s_{0:T+1})$ and implies $q_{\theta^\star}(a_{0:T}) = \pi(a_{0:T})$. See App.~\ref{app:dpi_gap_derivation} for more details on the DPI.
\endgroup

Expanding the temperature-scaled finite-horizon joint KL yields the negative maximum-entropy return, up to a policy-independent constant. With stationary dynamics and policies, we introduce $\Tau_t=\gamma^t\Tau$ and $R_t=\gamma^tR_{\mathrm{env}}$, with $\gamma\in[0,1)$. Taking $T\to\infty$ yields the discounted ME-RL objective below. An equivalent joint trajectory-KL formulation uses geometric stopping (App.~\ref{app:infinite_horizon_gaussian}).
\begin{equation}
\mathcal J_\mathrm{ME}(\theta)
:=
\mathbb E_{q_\theta}\!\left[
\sum_{t=0}^{\infty}\gamma^t
\left(R_{\mathrm{env}}(s_t,a_t)-\Tau\log q_\theta(a_t\mid s_t)\right)
\right].
\label{eq:KL_loss}
\end{equation}
Here, the expectation is over trajectories initialized with $s_0\sim p(\cdot)$ and generated by the policy and environment dynamics. In expectation subscripts, we write $a\sim q_\theta$ for $a\sim q_\theta(\cdot\mid s)$ and $s'\sim p$ for $s'\sim p(\cdot\mid s,a)$. Maximizing $\mathcal J_\mathrm{ME}(\theta)$ maximizes the expected environment reward together with the entropy of the policy.
Applying the policy-gradient theorem to Eq.~\ref{eq:KL_loss}, followed by a reverse application of the log-derivative trick, gives the stationary local ME-RL surrogate. Let $\thetar$ denote the frozen parameters of the policy used to collect the rollout.
For a policy $q$, let $\rho_t^q(s)$ denote the
probability density of the continuous state $s_t$ evaluated at $s$,
starting from $s_0\sim p(s_0)$.
The normalized discounted state-visitation density is
$d^{q_{\thetar}}(s):=(1-\gamma)\sum_{t=0}^{\infty}\gamma^t
\rho_t^{q_{\thetar}}(s)$.
The resulting surrogate objective is
\begin{equation}
\mathcal L_{\mathrm{ME}}(\theta;\thetar)
=
\frac{\Tau}{1-\gamma}
\mathbb E_{s'\sim d^{q_{\thetar}}}
\left[
D_{\mathrm{KL}}\!\left(
q_\theta(\cdot\mid s')
\,\middle\|\,
\frac{\exp\!\left(\alpha Q^{q_{\thetar}}(s',\cdot)\right)}
{Z(s')}
\right)
\right].
\label{eq:surroga_loss}
\end{equation}

The stationary soft value functions satisfy
\begin{equation*}
\begin{aligned}
Q^{q_\theta}(s,a)
&=R_{\mathrm{env}}(s,a)+\gamma\,\mathbb E_{s'\sim p}[V^{q_\theta}(s')],
\quad&
V^{q_\theta}(s)
&=\mathbb E_{a\sim q_\theta}[Q^{q_\theta}(s,a)-\Tau\log q_\theta(a\mid s)].
\end{aligned}
\end{equation*}
Both $Q^{q_{\thetar}}$ and $d^{q_{\thetar}}$ are frozen during the actor update. More precisely, minimizing the surrogate matches maximizing the return at the rollout parameters, i.e.~ $
\left.\nabla_\theta\mathcal L_{\mathrm{ME}}(\theta;\thetar)\right|_{\theta=\thetar}
=
-\left.\nabla_\theta\mathcal J_{\mathrm{ME}}(\theta)\right|_{\theta=\thetar}.$
Although the gradient identity is exact only at the rollout parameters, it provides the local ascent direction underlying the trust-region update. The theoretical improvement bound of \citet{schulman2015trust} shows that a positive surrogate gain implies improvement of the true objective whenever the distribution shift induced by the policy update remains sufficiently small.

\subsection{Discrete-Time Diffusion Policies}
\label{sec:diff_samplers}

A diffusion sampler targets sampling from $\pi_0(x^0)=\widetilde\pi_0(x^0)/Z$, given its unnormalized density. The forward process progressively noises target samples toward a simple reference distribution $q_{\mathrm{prior}}$, while the learned reverse process denoises reference samples toward the target \citep{DDS,berner2022optimal}. Their path distributions are
\begin{equation}
\pi_\theta(x^{0:K})=\pi_0(x^0)\prod_{k=1}^{K}\pi_\theta(x^k\mid x^{k-1}),
\qquad
q_\theta(x^{0:K})=q_{\mathrm{prior}}(x^K)\prod_{k=1}^{K}q_\theta(x^{k-1}\mid x^k).
\label{eq:diffusion_sampler_paths}
\end{equation}
Evaluating the reverse clean-sample density $q_\theta(x^0)$ generally requires intractable marginalization over the intermediate diffusion variables. Applying the DPI again to the projection onto $x^0$ gives
\begin{equation}
D_{\mathrm{KL}}\!\left(q_\theta(x^0)\,\middle\|\,\pi_0(x^0)\right)
\leq
D_{\mathrm{KL}}\!\left(q_\theta(x^{0:K})\,\middle\|\,\pi_\theta(x^{0:K})\right).
\label{eq:diffusion_sampler_dpi}
\end{equation}
The path-space KL on the right can be estimated from reverse samples using the reference, transition, and unnormalized target densities, up to the parameter-independent constant $\log Z$.

For bounded action spaces, as commonly encountered in continuous-control RL, we consider a discretized variance-preserving SDE (VP-SDE) with Gaussian kernels and $a=\tanh(x^0)$, or a Beta diffusion process on $[0,1]^N$ with $a=2x^0-1$ \citep{avdeyev2023dirichlet}. In all DA-MDP experiments, we jointly learn the reverse policy and the diffusion coefficients governing the forward and reverse schedules. Below, $a^0$ denotes the executed action, including the terminal change of variables. We compare these diffusion processes in App.~\ref{app:action_transform_entropy_ablation}.

\section{Diffusion-Augmented Markov Decision Processes}
\label{sec:Method}

We first derive the tractable diffusion-path objective from ME-RL, then construct the augmented MDP and its local critic and actor objectives.

\subsection{Deriving the DA-MDP Objective from ME-RL}
\label{sec:da_mdp_derivation}

We begin with a proof sketch of the DA-MDP objective, which is detailed in App.~\ref{app:infinite_horizon_da_mdp}. Because the executed-action marginal of a diffusion policy is intractable, we augment both trajectory distributions from Sec.~\ref{sec:maxentRL} with all intermediate diffusion variables. The normalized augmented policy and target distributions are
\begin{equation}
\begin{aligned}
q_\theta(a^{0:K}_{0:T},s_{0:T+1})
&=p(s_0)\prod_{t=0}^{T}\!\left[
p(s_{t+1}\mid s_t,a_t^0)q_{\mathrm{prior}}(a_t^K)
\prod_{k=1}^{K}q_\theta(a_t^{k-1}\mid a_t^k,s_t)\right],\\
\pi_\theta(a^{0:K}_{0:T},s_{0:T+1})
&=\frac{p(s_0)}{\mathcal Z}\prod_{t=0}^{T}\!\left[
p(s_{t+1}\mid s_t,a_t^0)e^{\alpha R_{\mathrm{env}}(s_t,a_t^0)}
\prod_{k=1}^{K}\pi_\theta(a_t^k\mid a_t^{k-1},s_t)\right].
\end{aligned}
\label{eq:q_pi_diff_chain}
\end{equation}
Marginalizing $a^{1:K}_{0:T}$ recovers the state--executed-action distributions used in Eq.~\ref{eq:rl_DPI}, so the data-processing inequality gives
\begin{equation}
D_{\mathrm{KL}}\!\left(
q_\theta(a_{0:T},s_{0:T+1})\,\middle\|\,
\pi(a_{0:T},s_{0:T+1})\right)
\leq
D_{\mathrm{KL}}\!\left(
q_\theta(a^{0:K}_{0:T},s_{0:T+1})\,\middle\|\,
\pi_\theta(a^{0:K}_{0:T},s_{0:T+1})\right).
\label{eq:rl_DPI_fixed}
\end{equation}
This then yields a tractable reverse KL between the augmented policy and target distributions, which upper-bounds their reverse KL over executed-action sequences (Eqs.~\ref{eq:rl_DPI} and~\ref{eq:rl_DPI_fixed}). Its zero-divergence optimality condition is $q_{\theta^\star}(a^{0:K}_{0:T},s_{0:T+1})=\pi_{\theta^\star}(a^{0:K}_{0:T},s_{0:T+1})$. As with any DPI bound, decreasing the upper bound does not imply that the marginal objective decreases monotonically after every parameter update. See App.~\ref{app:dpi_gap_derivation} for more details.

Expanding the augmented joint KL cancels the shared environment dynamics, while the fixed prior contributes only a parameter-independent constant. Applying the reward and temperature scaling from Sec.~\ref{sec:maxentRL} and taking $T\to\infty$ yields the stationary discounted DA-MDP return
\begin{equation}
\begin{aligned}
\mathcal J_{\mathrm{DA}}(\theta)
&:=
\mathbb E_{q_\theta}\!\left[
\sum_{t=0}^{\infty}\gamma^t
\left(
R_{\mathrm{env}}(s_t,a_t^0)
-\Tau\sum_{k=1}^{K}\log
\frac{q_\theta(a_t^{k-1}\mid a_t^k,s_t)}
     {\pi_\theta(a_t^k\mid a_t^{k-1},s_t)}
\right)
\right].
\end{aligned}
\label{eq:kl_decomp_fixed}
\end{equation}
Here, the expectation is over augmented trajectories generated as in Eq.~\ref{eq:q_pi_diff_chain}, including the initial state, prior samples, all denoising steps, and environment transitions.
Equivalently, the temperature-scaled geometrically stopped augmented KL equals $-\mathcal J_{\mathrm{DA}}(\theta)$ up to parameter-independent constants (App.~\ref{app:infinite_horizon_da_mdp}). The expression in parentheses identifies the local objective for each denoising decision: the environment reward appears only at $k=1$, while the reverse--forward log ratio appears at every diffusion step.

\subsection{Augmented MDP and Local Objectives}
\label{sec:MDP}

Applying the policy-gradient theorem to $\mathcal J_{\mathrm{DA}}$ in Eq.~\ref{eq:kl_decomp_fixed} yields the local critic and actor objectives and the associated DA-MDP construction presented in this section (see App.~\ref{app:infinite_horizon_da_mdp}).
We treat each denoising transition as a decision in an augmented MDP. For environment time $t=0,1,\ldots$ and denoising phase $k=K,\ldots,1$, define
\[
\tilde t(t,k)=tK+(K-k),\qquad
\tilde s_{\tilde t(t,k)}=(s_t,a_t^k,k),\qquad
\tilde a_{\tilde t(t,k)}=a_t^{k-1}.
\]
The augmented state contains the environment state, noisy action, and diffusion index; the action is the next denoised sample. Writing a generic successor as $\tilde s'=(s',a',k')$, the transition kernel is
\[
p_{\mathrm{DA}}\!\left(\tilde s'\mid(s_t,a_t^k,k),a_t^{k-1}\right)
=
\begin{cases}
\delta\!\left(\tilde s'=(s_t,a_t^{k-1},k-1)\right),
&k>1,\\[0.4em]
p(s'\mid s_t,a_t^0)\,q_{\mathrm{prior}}(a')\,\delta(k'=K),
&k=1.
\end{cases}
\]
Here, $\delta(u=v)$ denotes the Dirac measure concentrated at $v$. For $k>1$, the successor is deterministic and the environment state remains fixed. At $k=1$, the environment executes $a_t^0$, transitions to $s_{t+1}$, and a fresh prior sample starts the next denoising chain.
The environment reward is $\widetilde R_{\mathrm{DA}}=\mathbf 1_{\{k=1\}}R_{\mathrm{env}}(s_t,a_t^0)$. We set $\bar\gamma(k)=1$ for $k>1$ and $\bar\gamma(1)=\gamma$, so discounting occurs only when the environment advances. Figure~\ref{fig:da_mdp_overview} illustrates the construction.

For the frozen rollout parameters $\thetar$, the soft Bellman equations are
\begin{equation}
\begin{aligned}
Q_{\mathrm{DA}}^{q_{\thetar}}(\tilde s_{\tilde t},a_t^{k-1})
&=
\widetilde R_{\mathrm{DA}}(\tilde s_{\tilde t},a_t^{k-1})
+\bar\gamma(k)
\mathbb E_{\tilde s_{\tilde t+1}}
\!\left[V_{\mathrm{DA}}^{q_{\thetar}}(\tilde s_{\tilde t+1})\right],\\
V_{\mathrm{DA}}^{q_{\thetar}}(\tilde s_{\tilde t})
&=
\mathbb E_{a_t^{k-1}\sim q_{\thetar}(\cdot\mid\tilde s_{\tilde t})}
\left[
Q_{\mathrm{DA}}^{q_{\thetar}}(\tilde s_{\tilde t},a_t^{k-1})
-\Tau\log
\frac{q_{\thetar}(a_t^{k-1}\mid a_t^k,s_t)}
     {\pi_{\thetar}(a_t^k\mid a_t^{k-1},s_t)}
\right].
\end{aligned}
\label{eq:diffmaxentvaluefunction}
\end{equation}
where $\tilde s_{\tilde t+1}\sim p_{\mathrm{DA}}(\cdot\mid\tilde s_{\tilde t},a_t^{k-1})$. The reverse--forward log ratio replaces the usual entropy term.

Let $\rho_{t,k}^{q}(s,a^k)$ denote the joint density of the environment state and noisy action under policy $q$ at environment time $t$ and denoising phase $k$. For $\tilde s=(s,a^k,k)$, the normalized discounted occupancy is $d_{\mathrm{DA}}^{q_{\thetar}}(s,a^k,k):=\frac{1-\gamma}{K}\sum_{t=0}^{\infty}\gamma^t\rho_{t,k}^{q_{\thetar}}(s,a^k)$, which weights all $K$ denoising phases equally. Following the local-surrogate construction in Eq.~\ref{eq:surroga_loss}, the actor loss is
\begin{equation}
\mathcal L_{\mathrm{DA}}(\theta;\thetar)
=
\frac{K\Tau}{1-\gamma}
\mathbb E_{\tilde s\sim d_{\mathrm{DA}}^{q_{\thetar}}}
\left[
D_{\mathrm{KL}}^{\mathrm u}\!\left(
q_\theta(\cdot\mid\tilde s)
\,\middle\|\,
\pi_\theta(a^k\mid\cdot,s)
\exp\!\left(\alpha Q_{\mathrm{DA}}^{q_{\thetar}}(\tilde s,\cdot)\right)
\right)
\right].
\label{eq:diffusion_surrogate}
\end{equation}
Here, $D_{\mathrm{KL}}^{\mathrm u}$ denotes a KL-like variational free energy with an unnormalized right-hand argument (see Eq.~\ref{eq:unnormalized_kl_definition}). At the rollout parameters, minimizing this loss is locally equivalent to maximizing the DA-MDP return: $\left.\nabla_\theta\mathcal L_{\mathrm{DA}}(\theta;\thetar)\right|_{\theta=\thetar}=-\left.\nabla_\theta\mathcal J_{\mathrm{DA}}(\theta)\right|_{\theta=\thetar}$. App.~\ref{app:numerical_surrogate_verification} verifies this identity numerically on an exactly enumerable finite MDP with learned forward diffusion kernels.

\begingroup

\begin{proposition}[DA-MDP return improvement]
\label{prop:da_return_improvement}
Under the assumptions in App.~\ref{app:da_monotonic_improvement_proof}, let
$\Delta\mathcal J_{\mathrm{DA}}:=\mathcal J_{\mathrm{DA}}(\theta)-\mathcal J_{\mathrm{DA}}(\thetar)$
and define the surrogate gain
$\Delta\mathcal J_{\mathrm{DA}}^{\mathrm{surrogate}}
:=\mathcal L_{\mathrm{DA}}(\thetar;\thetar)-\mathcal L_{\mathrm{DA}}(\theta;\thetar)$.
For $\tilde s=(s,a^k,k)$, define
$D_q:=\sup_{\tilde s}\Dkl\!\left(q_\theta(\cdot\mid\tilde s)\|q_{\thetar}(\cdot\mid\tilde s)\right)$,
$\epsilon_A:=\sup_{\tilde s,a^{k-1}}|A_{\mathrm{roll}}(\tilde s,a^{k-1})|$,
and $\epsilon_G:=\sup_{\tilde s}\left|
\mathbb E_{a^{k-1}\sim q_\theta(\cdot\mid\tilde s)}
\left[\log\frac{\pi_\theta(a^k\mid a^{k-1},s)}
                   {\pi_{\thetar}(a^k\mid a^{k-1},s)}\right]\right|$,
where $A_{\mathrm{roll}}$ is the soft rollout advantage
(Eq.~\ref{eq:da_local_rollout_advantage}). With
$B_{K,\gamma}:=\frac{K}{1-\gamma}\left(K-1+\frac{2\gamma K}{1-\gamma}\right)$,
the return change satisfies
\begin{equation}
\Delta\mathcal J_{\mathrm{DA}}
\geq\Delta\mathcal J_{\mathrm{DA}}^{\mathrm{surrogate}}
-\left(\epsilon_A B_{K,\gamma}+\frac{K\Tau}{1-\gamma}\right)D_q
-\Tau B_{K,\gamma}\epsilon_G\sqrt{D_q/2}.
\label{eq:da_return_improvement_main}
\end{equation}
\end{proposition}
Thus, the return improves whenever the surrogate gain exceeds
the displayed penalties. Fixed forward kernels give $\epsilon_G=0$.
App.~\ref{app:da_monotonic_improvement_proof} provides the proof.
\endgroup

\subsection{Diffusion-Augmented MDP Algorithms: Examples}
\label{sec:mexEntWPO}

Our DA-MDP algorithms limit changes to the reverse policy using update constraints adapted from their Gaussian-policy counterparts. These constraints aim to keep the penalties in Proposition~\ref{prop:da_return_improvement} small relative to the surrogate gain. We instantiate the framework with PPO, REPPO, and WPO by applying their actor and critic updates to stored local denoising transitions. Each DA-MDP diffusion variant is explained in detail in App.~\ref{app:DA_MDP_algos}  and shortly summarized below.

\par\vspace{\runinheadingspace}\noindent\textbf{\dappo{}} applies the log-derivative trick to $\mathcal L_{\mathrm{DA}}$ in Eq.~\ref{eq:diffusion_surrogate}, using importance weights between the current and rollout reverse kernels for stored local denoising transitions. It estimates their advantages using a value critic and generalized advantage estimation on the DA-MDP rewards. Learning the forward process requires an additional forward log-density gradient \citep{sanokowski2025rethinking}. PPO's clipped surrogate controls reverse-policy updates. {Setting $\Tau=0$ recovers a DPPO-style local PPO objective on the denoising-augmented MDP} \citep{ren2025diffusion}. See App.~\ref{app:diffppo} for details.
\par\vspace{\runinheadingspace}\noindent\textbf{\dareppo{}} applies the reparameterization trick to $\mathcal L_{\mathrm{DA}}$ in Eq.~\ref{eq:diffusion_surrogate}, using an action-value critic on the augmented MDP and KL-based trust-region constraints between the rollout and current reverse kernels. App.~\ref{app:da_reppo} provides the complete objective and implementation details.
\par\vspace{\runinheadingspace}\noindent\textbf{\dawpo{}.}
To instantiate the framework with WPO, we derive its maximum-entropy extension ME-WPO in App.~\ref{app:WPO_deriv}. The complete Algorithm is presented in App.~\ref{app:me_wpo_algorithm}. Projecting the Wasserstein gradient flow of the soft policy-improvement objective onto the policy parameters makes the update depend on both the action-value gradient and the policy's entropy score.
When extending this to DA-MDPs we obtain \dawpo{} by applying the same score-matching construction to the local diffusion target in Eq.~\ref{eq:diffusion_surrogate}. This yields
{\small
\begin{equation}
\mathcal L_{\mathrm{DA:WPO}}(\theta)
=
\frac{\Tau}{2}
\mathbb E_{\substack{
\tilde s_{\tilde t}\sim d_{\mathrm{DA}}^{q_{\thetar}}\\
a_t^{k-1}\sim q_{\thetastop}(\cdot\mid\tilde s_{\tilde t})
}}
\left[
\left\|
\nabla_{a_t^{k-1}}
\left(
\log
\frac{q_\theta(a_t^{k-1}\mid a_t^k,s_t)}
     {\pi_\theta(a_t^k\mid a_t^{k-1},s_t)}
-\alpha Q_{\mathrm{DA}}^{q_{\thetar}}(\tilde s_{\tilde t},a_t^{k-1})
\right)
\right\|^2
\right].
\label{eq:diffmaxent_wpo_main}
\end{equation}
}
The augmented state is sampled from the frozen discounted visitation distribution $d_{\mathrm{DA}}^{q_{\thetar}}$, and $a_t^{k-1}$ is then drawn from the current reverse kernel with gradients stopped. \dawpo{} otherwise uses the same augmented rollout, critic, and KL-based trust-region constraints on the reverse kernels as \dareppo{} and replaces only its policy loss by Eq.~\ref{eq:diffmaxent_wpo_main}. For variance-preserving Gaussian reverse kernels, we approximate full parameter-space inverse-Fisher preconditioning using adjusted per-coordinate blocks in the local control and diffusion-coefficient coordinates. App.~\ref{app:da_wpo} presents the \dawpo{} algorithm, and App.~\ref{app:diff_WPO_derivation} derives its objective $\mathcal L_{\mathrm{DA:WPO}}$.

\section{Related Work}

Diffusion and flow policies provide expressive alternatives to Gaussian actors and have been trained with supervised, policy-gradient, and variational objectives \citep{ren2025diffusion, celik2025dime, dong2025maximum, ma2025reinforcement, mcallister2025flow}. DA-MDPs build on the local policy-gradient view of scalable discrete diffusion samplers \citep{Sanokowski2025scalable} and extend it to continuous-control RL. 
\begingroup
DIME \citep{celik2025dime} is conceptually closest to our approach. It starts from the per-state ME-RL objective, expressed as
the action-marginal divergence $\mathcal D_{\mathrm{marg}}(s):=D_{\mathrm{KL}}\!\left(q_\theta(a^0\mid s)\,\middle\|\,\exp(\alpha Q(s,a^0))/Z(s)\right)$. Applying the DPI yields the tractable full-path upper bound $\mathcal D_{\mathrm{path}}(s):=D_{\mathrm{KL}}\!\left(q_\theta(a^{0:K}\mid s)\,\middle\|\,\pi_\theta(a^{1:K}\mid a^0,s)\exp(\alpha Q(s,a^0))/Z(s)\right)\geq\mathcal D_{\mathrm{marg}}(s)$, where $\pi_\theta(a^{1:K}\mid a^0,s)$ is the conditional forward path.
TruDi \citep{le2026trust} instantiates this full-path objective within REPPO and is therefore the closest algorithmic counterpart to \dareppo{}. DIME and TruDi backpropagate through the complete reverse and forward chains, which requires $\mathcal O(K)$ active memory. DA-MDPs instead optimize one reverse transition conditioned on $\tilde s_{\tilde t}=(s_t,a_t^k,k)$. Each sampled transition has $\mathcal O(1)$ cost, and a minibatch of $B$ local transitions uses $\mathcal O(B)$ memory. A dataset of $N$ environment transitions contains $NK$ local transitions; processing it once requires approximately $NK/B$ sequential minibatch updates. See App.~\ref{app:related_work} for a more detailed discussion about related work.
\endgroup

\section{Experiments}
\label{sec:experiments}
\label{sec:toy_example}
\label{sec:diffusion_step_ablation}
\label{sec:memory_runtime}

We evaluate the proposed diffusion-augmented algorithms on standard continuous-control benchmarks and tasks designed to assess multimodality and behavioral diversity. Our experiments address five questions: whether the DA-MDP variants improve their corresponding base algorithms, whether they are able to represent multimodal action distributions, whether this diversity extends to contact-rich manipulation, how performance depends on the number of diffusion steps, and how the local one-step objective trades memory for runtime compared with a full-chain method.
\par\vspace{\runinheadingspace}\noindent\textbf{Multimodality and Behavioral Diversity.}
To assess whether the DA-MDP algorithms can represent multimodal action distributions, we introduce a simple control task with a bimodal reward landscape (see App.~\ref{app:double_well} for task details). The reward has the same two symmetric optima in each state and therefore induces an inherently multimodal optimal policy. The transition dynamics discretize the state space into a small set of orientations, which allows us to visualize the complete learned action distribution at every state.

\begin{figure*}[!t]
    \centering
    \setlength{\parskip}{0pt}
    \captionsetup{skip=2pt}
    \captionsetup[subfigure]{skip=2pt,justification=centering}
    \begin{subfigure}[t]{\textwidth}
        \centering
        \resizebox{\linewidth}{!}{\shortstack{\includegraphics[trim=356.7445bp 286.7047bp 369.5087bp 0bp,clip]{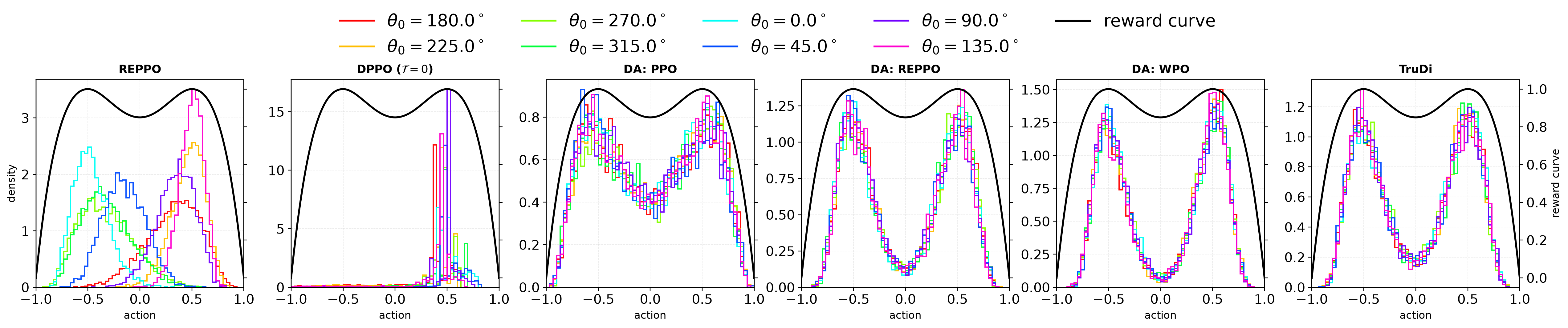}\\
            \includegraphics[trim=0bp 0bp 322.0519bp 62.1848bp,clip]{Figures/Ablations/multi_agent_hist_extended.png}\includegraphics[trim=1614.5142bp 0bp 0bp 62.1848bp,clip]{Figures/Ablations/multi_agent_hist_extended.png}}}
        \caption{Learned action distributions and bimodal reward landscape.}
        \label{fig:main_multimodal_distributions}
    \end{subfigure}
    \par\smallskip
    \begin{subfigure}[t]{\textwidth}
        \centering
        \includegraphics[width=\linewidth,trim=0bp 0bp 308.4149bp 0bp,clip]{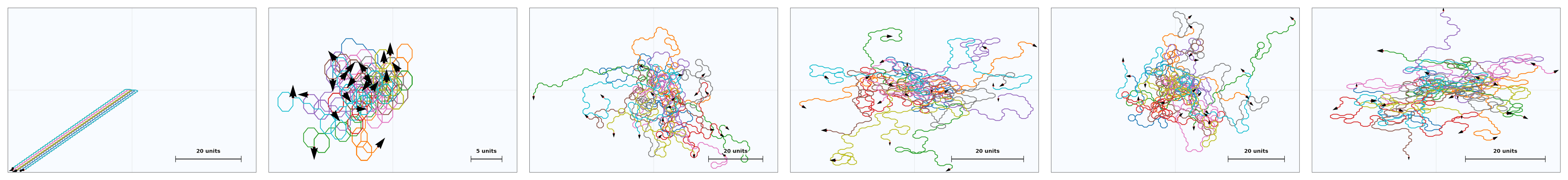}
        \caption{Resulting agent trajectories, with methods ordered as in panel (a).}
        \label{fig:main_multimodal_trajectories}
    \end{subfigure}
    \caption{Action multimodality and behavioral diversity on the Multimodal Agent benchmark. (a) REPPO and DPPO remain unimodal; \dappo{}, \dareppo{}, and \dawpo{} assign probability mass to both reward modes. (b) REPPO produces directed motion, and DPPO produces cyclic trajectories; the DA-MDP methods exhibit more diverse trajectories. Figure~\ref{fig:extended_multimodal} in the appendix provides the full comparison, including TruDi and state-visitation histograms.}
    \label{fig:bothplots}
\end{figure*}

We evaluate REPPO, TruDi, DPPO, \dareppo{}, \dawpo{}, and \dappo{} on this benchmark. Figure~\ref{fig:bothplots} shows the learned action distributions and resulting agent behavior for the three DA-MDP methods, REPPO, and DPPO; Figure~\ref{fig:extended_multimodal} in App.~\ref{app:double_well} provides the full comparison, including TruDi. The REPPO policy collapses to one mode in every state and consequently drives the agent toward the lower-left corner. DPPO also remains unimodal and produces cyclic trajectories. In contrast, all three DA-MDP methods assign probability mass to both reward modes, resulting in substantially more diverse behavior (Fig.~\ref{fig:main_multimodal_trajectories}); TruDi exhibits the same qualitative behavior (Fig.~\ref{fig:multimodal_trajectories}).
We also quantify the tightness of the DPI bound on this benchmark. Writing $D_{\mathrm{aug}}$ and $D_{\mathrm{act}}$ for the augmented-trajectory and executed-action-sequence KLs, respectively, the DPI gap is $D_{\mathrm{aug}}-D_{\mathrm{act}}=\Delta_{\mathrm{state}}+\Delta_{\mathrm{diffusion}}$ (Eq.~\ref{eq:exact_dpi_gap_decomposition}). Lower KL values indicate closer distributional matching, while smaller gaps indicate tighter bounds. \dareppo{} achieves a lower action-marginal KL ($3.742$ versus $927.011$ nats) and a smaller state gap ($0.069$ versus $2.079$ nats) than REPPO, despite an additional diffusion gap of $284.603$ nats. The full \dareppo{} decomposition is $288.415\approx3.742+0.069+284.603$ nats (details in App.~\ref{app:dpi_gap_numerical}).

\begingroup
\par\vspace{\runinheadingspace}\noindent\textbf{Contact-Rich Multimodality on StackCube.}
\label{sec:stackcube_multimodality}
{We train on the symmetric StackCube task of \citet{le2026trust}, where either stacking order is valid (App.~\ref{app:stackcube_multimodality}). We evaluate final checkpoints after extended training, with training success approaching a plateau for most runs; App.~\ref{app:manipulation_learning_curves} reports per-seed learning curves and budgets. For each of three training seeds, we generate repeated rollouts from a shared set of configurations drawn IID from the training distribution. We measure strict success $p$ and success-weighted stacking-order entropy $pH$, where $H$ quantifies the balance between the two successful orders. Statistics are computed within each configuration before averaging over configurations and seeds.}
{DA-MDP methods achieve higher success rates and generally greater success-weighted stacking-order entropy than stochastic Gaussian REPPO (results table in Fig.~\ref{fig:stackcube_metrics}). Under stochastic execution, the unchunked \dareppo{}, \dappo{}, and \dawpo{} policies achieve $pH=0.149$--$0.170$ bits, more than twice Gaussian REPPO's $0.072$ bits. Their advantage therefore extends beyond task completion to realizing alternative successful strategies from the same initial configuration.}
{Figure~\ref{fig:stackcube_xy_heatmaps} shows successful multimodality across varied cube placements. It also highlights widely separated configurations where \dareppo{} realizes both stacking orders while Gaussian REPPO rarely succeeds.}
{The DA-MDP framework also supports \textbf{Action Chuncking}, allowing several actions to be generated jointly and executed per policy call and is a widely used approach in diffusion-based robot control \citep{chi2025diffusion,ren2025diffusion}. With four-action chunks, \dareppo{} achieves the highest success-weighted entropy under every evaluated sampling scheme (table in Fig.~\ref{fig:stackcube_metrics}). Under ODE sampling, it reaches $88.9\%$ success and $pH=0.188$ bits, nearly twice the unchunked variant's $0.096$ bits. This successful diversity accompanies a substantial computational benefit: generating four actions in one diffusion pass reduces amortized ODE inference from $67.07$ to $17.01$\,ms per executed action on an A100, a $3.94\times$ speedup (App.~\ref{app:stackcube_chunking_timing}). Thus, the chunked policy combines high task success and diverse stacking strategies with nearly fourfold faster action generation.}

\par\vspace{\runinheadingspace}\noindent\textbf{Orientation-Conditioned Multimodality on PushT.}
\label{sec:pusht_multimodality}
{On PushT \citep{le2026trust}, we examine whether diffusion policies can maintain successful control while transitioning between distinct solution strategies. After approximately $250$M interactions, \dareppo{} (ODE with sampled prior) achieves $98.3\pm0.4\%$ success across three training seeds, compared with $84.5\pm1.6\%$ for stochastic Gaussian REPPO on IID initial configurations (mean $\pm$ SD; evaluation protocols and training budgets in Apps.~\ref{app:pusht_multimodality} and~\ref{app:manipulation_learning_curves}).}
{Figure~\ref{fig:pusht_motion_and_metrics} reveals the difference near the transition between clockwise and counter-clockwise solutions. Holding the goal and robot poses fixed, we vary the block's initial orientation. \dareppo{} changes its preferred direction between $190^\circ$ and $192^\circ$, sampling successful trajectories in both directions from the same initial configuration while maintaining over $99\%$ success at both angles. Gaussian REPPO solves the outer orientations but struggles to connect the two strategies, achieving only $1.2\%$--$1.4\%$ success in this switching region. At $190^\circ$, success-weighted mode entropy is $0.467$ bits for \dareppo{} versus $0.012$ bits for Gaussian REPPO. These results demonstrate that the diffusion policy supports alternative successful strategies through the switching transition, consistent with the benefit of representing multimodal action distributions.}

\begin{figure*}[!t]
    \normalcolor
    \centering
    \setlength{\parskip}{0pt}
    \captionsetup{skip=3pt}
    \captionsetup[subfigure]{skip=0pt,justification=centering}
    \begin{subfigure}[t]{\textwidth}
        \begin{minipage}[t]{.48\linewidth}
            \centering
            {\small\textbf{StackCube}}\par\vspace{2pt}
            \includegraphics[height=1.02in,viewport=0 0 320 270,clip]{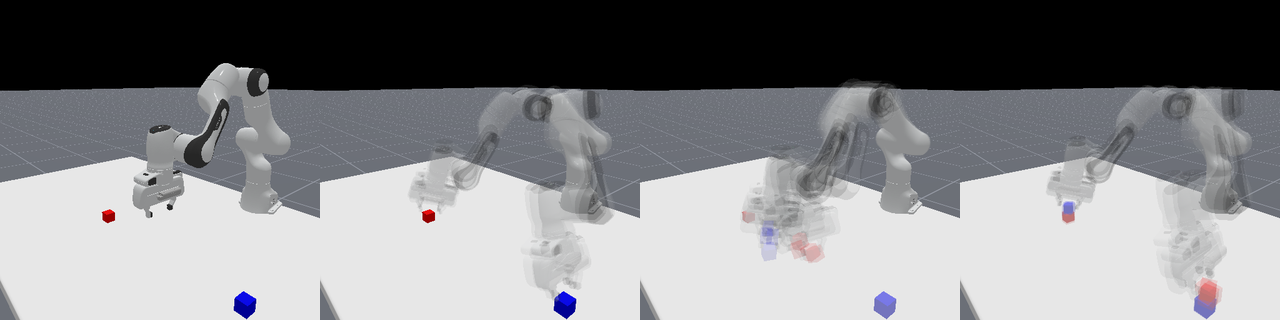}\hspace{2pt}\includegraphics[height=1.02in,viewport=960 0 1280 270,clip]{Figures/Rebuttal/stackcube_g08_successful_ode_overlay_strip.png}
        \end{minipage}
        \hfill
        \begin{minipage}[t]{.50\linewidth}
            \centering
            {\small\textbf{PushT: CW and CCW}}\par\vspace{2pt}
            \includegraphics[height=1.02in,viewport=512 0 2560 1248,clip]{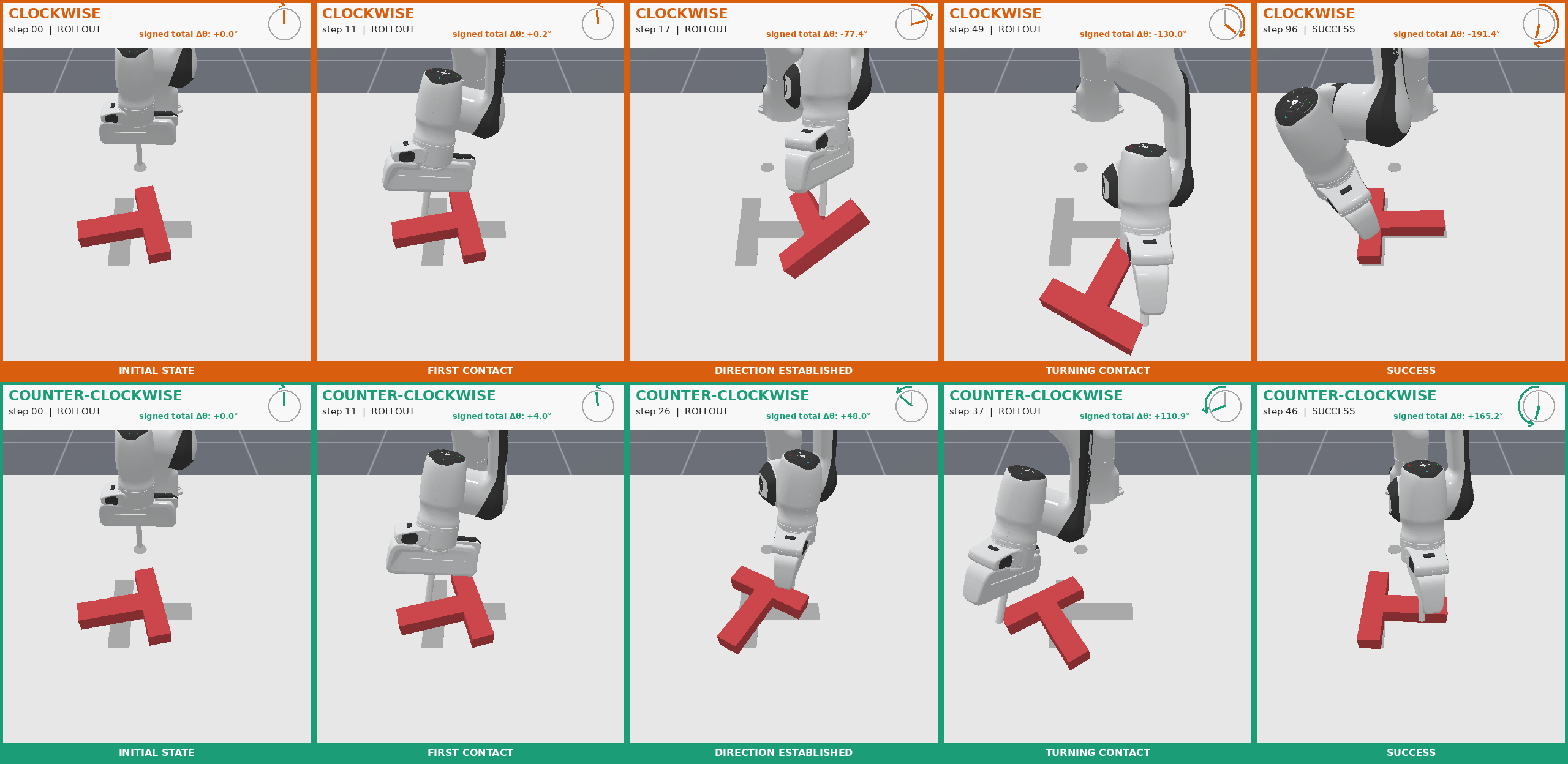}
        \end{minipage}
        \caption{}
        \label{fig:stackcube_motion_overlays}
    \end{subfigure}
    \par\smallskip
    \begin{subfigure}[t]{.48\textwidth}
        \vspace{0pt}
        \centering
        \begingroup\fontsize{6.5}{8}\selectfont
\setlength{\tabcolsep}{1pt}\renewcommand{\arraystretch}{1.08}
{\begin{adjustbox}{max width=\linewidth}
\begin{tabular}{@{}lccc@{\hspace{4pt}}ccc@{}}\toprule
 & \multicolumn{3}{c}{\textbf{Success (\%) $\uparrow$}} & \multicolumn{3}{c}{\shortstack{\textbf{Success-weighted}\\\textbf{$pH$ (bits) $\uparrow$}}}\\
\cmidrule(lr){2-4}\cmidrule(l){5-7}
Method & \shortstack{Means\\+ prior} & \shortstack{ODE\\+ prior} & Stoch. & \shortstack{Means\\+ prior} & \shortstack{ODE\\+ prior} & Stoch.\\\midrule
DA:REPPO $C{=}1$ & \textbf{\boldmath \makecell[t]{$96.5$\\$\pm0.2$}} & \textbf{\boldmath \makecell[t]{$96.4$\\$\pm0.5$}} & \textbf{\boldmath \makecell[t]{$95.0$\\$\pm0.8$}} & \makecell[t]{$.107$\\$\pm.008$} & \makecell[t]{$.096$\\$\pm.008$} & \makecell[t]{$.149$\\$\pm.011$}\\
DA:REPPO $C{=}4$ & \makecell[t]{$92.6$\\$\pm2.3$} & \makecell[t]{$88.9$\\$\pm4.4$} & \makecell[t]{$88.2$\\$\pm3.0$} & \textbf{\boldmath \makecell[t]{$.126$\\$\pm.007$}} & \textbf{\boldmath \makecell[t]{$.188$\\$\pm.015$}} & \textbf{\boldmath \makecell[t]{$.211$\\$\pm.009$}}\\
DA:PPO & \makecell[t]{$93.9$\\$\pm2.0$} & \makecell[t]{$92.6$\\$\pm3.4$} & \makecell[t]{$89.2$\\$\pm2.3$} & \makecell[t]{$.093$\\$\pm.013$} & \makecell[t]{$.069$\\$\pm.007$} & \makecell[t]{$.166$\\$\pm.015$}\\
DA:WPO & \makecell[t]{$96.0$\\$\pm1.6$} & \makecell[t]{$96.1$\\$\pm1.5$} & \makecell[t]{$94.0$\\$\pm1.5$} & \makecell[t]{$.104$\\$\pm.011$} & \makecell[t]{$.104$\\$\pm.016$} & \makecell[t]{$.170$\\$\pm.015$}\\
Gaussian REPPO & --- & --- & \makecell[t]{$82.7$\\$\pm2.1$} & --- & --- & \makecell[t]{$.072$\\$\pm.003$}\\
\bottomrule\end{tabular}\end{adjustbox}}
\par\vspace{1pt}
{\fontsize{5.8}{7}\selectfont{Mean $\pm$ SD across three training seeds; budgets differ.}}
\par{\fontsize{5.8}{7}\selectfont{$\uparrow$: higher is better. Bold: best mean per column.}}
\endgroup

        \caption{}
        \label{fig:stackcube_metrics}
    \end{subfigure}
    \hfill
    \begin{subfigure}[t]{.50\textwidth}
        \vspace{0pt}
        \includegraphics[width=\linewidth]{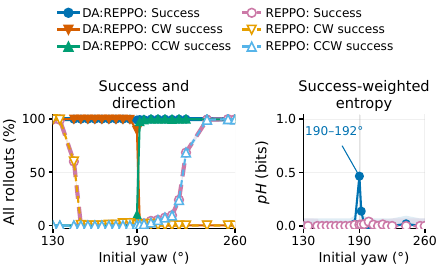}
        \caption{}
        \label{fig:pusht_motion_and_metrics}
    \end{subfigure}
    \caption{{Multimodal manipulation. (a) Illustrative motions. (b) IID StackCube success and success-weighted mode entropy $pH$: mean $\pm$ SD across three training seeds. (c) PushT success, successful CW/CCW fractions of all attempts, and $pH$: one seed, $95\%$ sampling intervals.}}
    \label{fig:pusht_multimodality}
\end{figure*}
\endgroup

\par
\begin{figure*}[t]
    \centering
    \setlength{\parskip}{0pt}
    \captionsetup{skip=3pt}
    \captionsetup[subfigure]{skip=0pt,justification=centering}
    \begin{subfigure}[b]{0.32\textwidth}
        \includegraphics[width=\linewidth,height=0.58\linewidth]{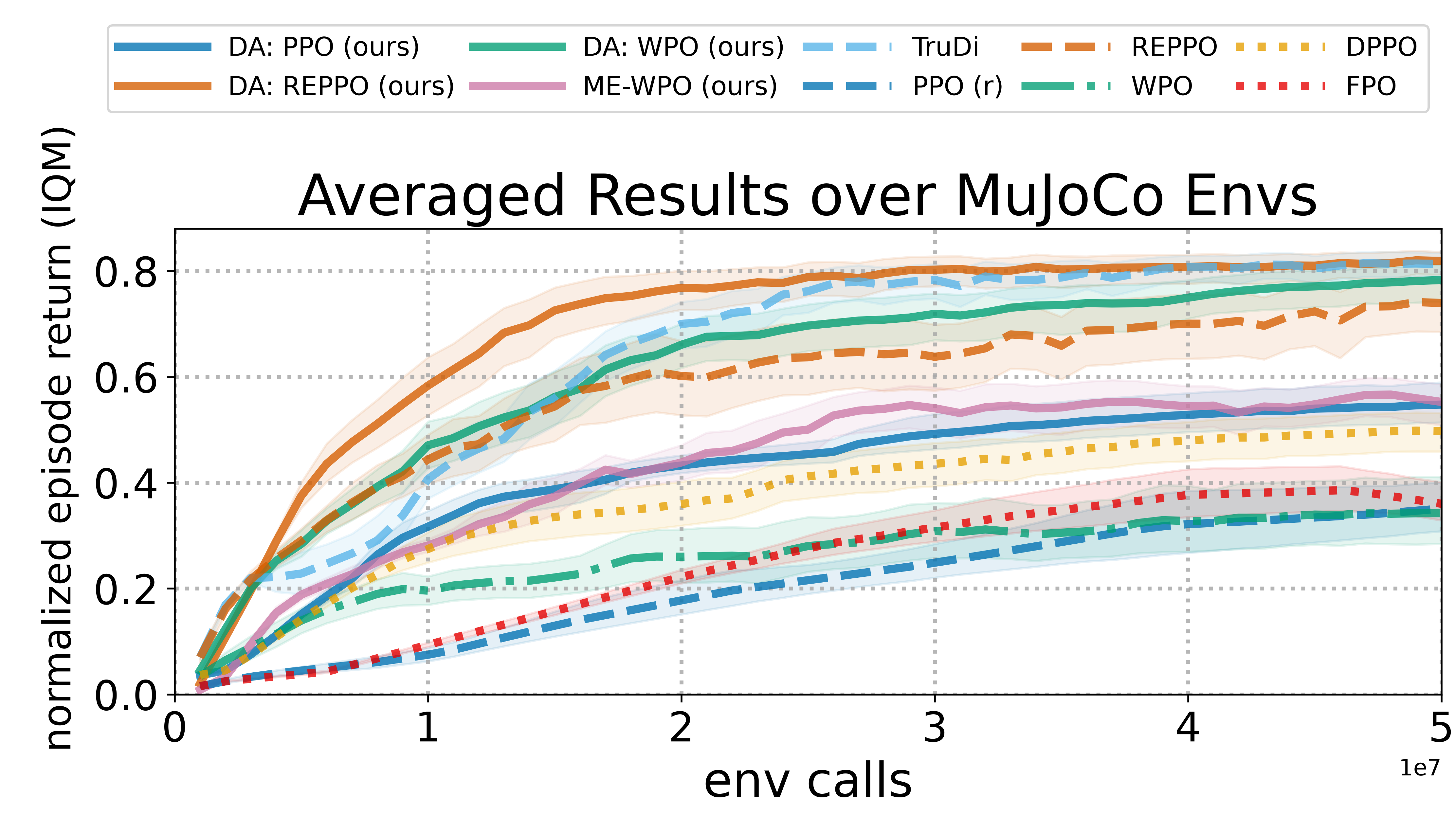}
        \caption{}
        \label{fig:all_envs}
    \end{subfigure}
    \hfill
    \begin{subfigure}[b]{0.32\textwidth}
        \begin{minipage}[c][0.58\linewidth][c]{\linewidth}
            \centering
            \begingroup\fontsize{6.5}{8}\selectfont
\setlength{\tabcolsep}{1.5pt}\renewcommand{\arraystretch}{1.08}
{\begin{adjustbox}{max width=\linewidth}
\begin{tabular}{@{}cccc@{}}\toprule
\multicolumn{4}{c}{\textbf{G1 deterministic return $\uparrow$}}\\
$K$ & \dareppo{} & \dappo{} & \dawpo{}\\
 & $50.07$M & $200.02$M & $69.07$M\\\midrule
$2$ & $21.11\pm0.81$ & $-1.89\pm0.05$ & $-1.60\pm0.02$\\
$4$ & $24.98\pm0.68$ & $20.30\pm2.07$ & $15.89\pm0.55$\\
$8$ & \textbf{\boldmath $28.32\pm0.42$} & \textbf{\boldmath $31.74\pm0.78$} & \textbf{\boldmath $18.04\pm0.40$}\\
\bottomrule\end{tabular}\end{adjustbox}}
\endgroup

        \end{minipage}
        \caption{}
        \label{fig:diffusion_step_ablation}
        \label{fig:g1_ppo_diffusion_step_ablation}
        \label{fig:g1_wpo_diffusion_step_ablation}
    \end{subfigure}
    \hfill
    \begin{subfigure}[b]{0.32\textwidth}
        \includegraphics[width=\linewidth,height=0.58\linewidth]{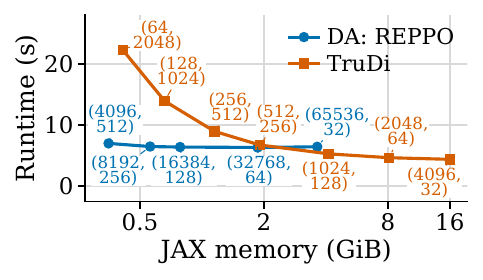}
        \caption{}
        \label{fig:jax_runtime_sweep}
    \end{subfigure}
    \caption{(a) Equal-task-weighted IQM of normalized returns across twelve tasks with $95\%$ stratified-bootstrap confidence intervals. {(b) G1 returns at the reported checkpoints (mean $\pm$ SD over seeds). Headers give method-specific interaction counts; bold marks the best depth per method.} (c) Memory--runtime trade-off between \dareppo{} and TruDi on CheetahRun at $K=16$.}
    \label{fig:experiment_summary}
\end{figure*}

\vspace{\runinheadingspace}\noindent\textbf{Diffusion-Step Ablation.}
{We ablate the number of diffusion steps, $K\in\{2,4,8\}$, for \dareppo{}, \dappo{}, and \dawpo{} on G1JoystickFlatTerrain. Figure~\ref{fig:diffusion_step_ablation} reports deterministic returns at method-specific checkpoints. Increasing $K$ from $2$ to $8$ raises the mean return from $21.11$ to $28.32$ for \dareppo{}, from $-1.89$ to $31.74$ for \dappo{}, and from $-1.60$ to $18.04$ for \dawpo{}. The evaluation learning curves are presented in Fig.~\ref{fig:g1_diffusion_learning_curves}. Larger diffusion depths thus improve the reported return for all three methods. See App.~\ref{app:diffusion_step_details}. for details.}

\par\vspace{\runinheadingspace}\noindent\textbf{Memory--Runtime Trade-off with TruDi.} To test how the local one-step objective trades active memory against training time, we benchmark \dareppo{} against the full-chain method TruDi on CheetahRun using $16$ diffusion steps and vary the minibatch size of each method (Fig.~\ref{fig:jax_runtime_sweep}). Point labels show (samples per minibatch, optimizer updates), with median runtime measured per 131,072 environment transitions. Every TruDi sample requires differentiation through the complete reverse and forward diffusion chains; under a tight memory budget, TruDi must therefore process fewer samples per minibatch and perform more optimizer updates. \dareppo{} instead processes local one-step transitions, allowing substantially larger minibatches at the same memory budget. Across all measured active-memory settings up to approximately $1.9$ GiB, \dareppo{} is faster, requiring $6.3$--$7.0$ s per training block compared with $6.7$--$22.3$ s for TruDi. TruDi becomes faster only when additional memory allows more full-chain samples to be processed in each minibatch. Thus, \dareppo{} is particularly advantageous when larger networks or more diffusion steps make memory the limiting resource. Profiling details are given in App.~\ref{app:runtime}.

\par\vspace{\runinheadingspace}\noindent\textbf{Continuous-control Experiments.}
On MuJoCo Playground \mbox{\citep{zakkamujoco}}, we compare \dappo{}, \dareppo{}, and \dawpo{} with their respective non-diffusion counterparts and with DPPO \citep{ren2025diffusion}, TruDi \citep{le2026trust}, and FPO \citep{mcallister2025flow}. Figure~\ref{fig:all_envs} reports the aggregate IQM \citep{agarwal2021deep}; the task-level learning curves are in Fig.~\ref{fig:main_results} in App.~\ref{app:continuous_control_protocol}. Unless stated otherwise, diffusion policies use $K=8$ steps. App.~\ref{app:continuous_control_protocol} describes the benchmark setup and per-task results. We first evaluate the maximum-entropy extension of WPO. WPO and ME-WPO share the same REPPO implementation and differ only in their policy loss and entropy treatment: ME-WPO automatically tunes the temperature, whereas WPO sets $\Tau=0$. The aggregate curves in Fig.~\ref{fig:all_envs} and task-level curves in Fig.~\ref{fig:main_results} (App.~\ref{app:continuous_control_protocol}) show that ME-WPO is more stable and performs better than WPO, although REPPO remains stronger overall. At matched environment interactions, each DA variant also improves on its non-diffusion counterparts, with the largest gains on HopperHop, HopperStand, FingerSpin, and the sparse swing-up tasks. In particular, \dappo{} learns faster than PPO and reaches a higher final return. Comparing \dappo{} with DPPO isolates the maximum-entropy correction because DPPO uses the same augmented MDP with $\Tau=0$. A fixed positive temperature improves CartpoleSwingupSparse and FishSwim and slightly increases aggregate IQM, but cannot adapt across tasks like the automatically tuned temperatures of \dareppo{}, \dawpo{}, and TruDi, which span several orders of magnitude. Finally, \dareppo{} learns faster initially than TruDi and reaches a comparable final IQM, while FPO is weaker than the other diffusion baselines. See App.~\ref{app:continuous_control_protocol} for protocol and baseline details.

\par\vspace{0.0em}\noindent\textbf{Diffusion-Sampler Ablation.} We compare VP-SDE and bounded Beta diffusion for \dareppo{} (Fig.~\ref{fig:action_transform_entropy_results} in App.~\ref{app:action_transform_entropy_ablation}). Our diffusion policies use VP-SDE in all other experiments. For VP-SDE, we additionally compare latent-entropy and transformed-action entropy formulations.

\section{Conclusion, Limitations and Future Work}

DA-MDPs provide a general framework for adapting maximum-entropy RL algorithms to diffusion policies. Derived from the variational formulation of ME-RL, the construction expresses denoising as a sequence of local decisions with diffusion-augmented soft rewards, value functions, and policy objectives. This allows each update to differentiate through only one diffusion transition. The accompanying return-improvement bound establishes conditions under which improvements in the local surrogate translate into higher augmented return, including when the forward kernels are learned.

In our continuous-control experiments, the resulting variants of PPO, REPPO, and maximum-entropy WPO improve on their non-diffusion counterparts. On StackCube, \dareppo{} achieves higher success and success-weighted mode entropy than Gaussian REPPO, and realizes both stacking orders at some widely separated placements where the baseline fails. DA policies also retain successful diversity under action chunking. On PushT, \dareppo{} realizes both motion directions near the transition, where Gaussian REPPO has near-zero success and little successful diversity.  Our ablations show that additional diffusion steps can improve performance, while the runtime study demonstrates the practical benefit of local training when memory is constrained.

Adapting a Gaussian ME-RL algorithm to a DA-MDP can also require method-specific design choices. In \dappo{}, learning the forward kernels requires an explicit gradient correction to the PPO surrogate. In \dareppo{}, auxiliary critic losses must account for both environment transitions and intermediate denoising transitions. Entropy tuning remains task dependent: greater entropy can encourage exploration and diversity but may favor lower-reward actions. Learned state-dependent priors and extensions to discrete or structured action spaces offer directions for future research. App.~\ref{app:limitations} discusses additional limitations and future work.

\section*{Acknowledgements}

We thank Gerhard Neumann and Denis Blessing for helpful discussions. We also thank Huy Le for sharing the \trudi{} codebase and answering our questions about its implementation.

We acknowledge the EuroHPC Joint Undertaking for awarding us access to Vega at the Institute of Information Science (IZUM), Slovenia; Karolina at IT4Innovations National Supercomputing Center, Czechia; MeluXina at LuxProvide, Luxembourg; Discoverer at Sofia Tech Park, Bulgaria; Deucalion at the Minho Advanced Computing Center (MACC), Portugal; and MareNostrum5 at the Barcelona Supercomputing Center (BSC), Spain. We thank the staff of these centres for their technical support.

\clearpage
\newpage

\bibliographystyle{iclr2027_conference}
\bibliography{neuripsbib}

\newpage
\onecolumn
\appendix
\renewcommand{\thetar}{{\theta_{\mathrm{roll}}}}

\etocdepthtag.toc{appendixmatter}

\begingroup
\etocsettagdepth{mainmatter}{none}
\etocsettagdepth{appendixmatter}{subsection}
\etocobeydepthtags
\etocsetnexttocdepth{subsection}
\etocsettocstyle{\section*{Appendix Contents}}{}
\tableofcontents
\endgroup
\etocignoredepthtags

\clearpage
\section{Visual Overview of DA-MDPs}
\label{app:graphical_overview}

Figure~\ref{fig:da_mdp_overview} shows how a diffusion policy turns one environment-level decision into $K$ local denoising decisions. The same state--action diagram describes both processes: circles denote states, rounded squares denote actions, and blue arrows denote policy samples.

\paragraph{States and actions.}
For readability, write $\tilde s_{t,k}:=\tilde s_{\tilde t(t,k)}=(s_t,a_t^k,k)$, with $\tilde t(t,k)=tK+(K-k)$. The reverse kernel $q_\theta(a_t^{k-1}\mid a_t^k,s_t)$ samples the local action. For $k>1$, the successor is deterministically $(s_t,a_t^{k-1},k-1)$. At $k=1$, the environment executes $a_t^0$, samples $s_{t+1}\sim p(\cdot\mid s_t,a_t^0)$, and starts a new denoising chain with $a_{t+1}^K\sim q_{\mathrm{prior}}$, as formalized in Sec.~\ref{sec:MDP}.

\paragraph{Rewards and discounting.}
Here, $r_t=R_{\mathrm{env}}(s_t,a_t)$, with $a_t=a_t^0$ in panel (b). The ordinary policy's entropy contribution is replaced by a sum of reverse--forward log-ratio contributions. In the local boxes, $q_\theta$ and $\pi_\theta$ denote the reverse and forward conditional densities evaluated at the displayed denoising transition; their full arguments appear in the return below. The environment reward appears only at $k=1$. Denoising leaves the environment clock unchanged: the discount is one for $k>1$ and $\gamma$ at $k=1$. Thus, all $K$ contributions at environment step $t$ share the factor $\gamma^t$. Taking expectations of the displayed returns recovers the objectives in Eqs.~\ref{eq:KL_loss} and~\ref{eq:kl_decomp_fixed}.

\clearpage

\section{Notation}
\label{app:notation}

\begingroup
\begin{table}[H]
\centering
\caption{Notation used throughout the paper.}
\label{tab:notation}
\renewcommand{\arraystretch}{1.12}
\small
\begin{tabularx}{\textwidth}{@{}>{$}l<{$}X@{}}
\toprule
\text{Symbol} & \text{Meaning} \\
\midrule
t\in\{0,1,\ldots\} & Environment-time index in the stationary discounted setting; $T$ is used only for the finite trajectory prefixes that introduce the DPI. \\
k\in\{K,\ldots,1\} & Reverse-diffusion index; $a_t^K$ is prior noise and $a_t^0$ is executed in the environment. \\
\tilde t,\;\tilde s_{\tilde t},\;\tilde a_{\tilde t} & Augmented index, state, and action: $\tilde t=tK+(K-k)$, $\tilde s_{\tilde t}=(s_t,a_t^k,k)$, and $\tilde a_{\tilde t}=a_t^{k-1}$. \\
q_\theta,\;\pi_\theta & Learned reverse policy/kernel and forward diffusion kernel, respectively; $\widetilde\pi$ denotes an unnormalized reward target and $\pi$ its normalized version. \\
\theta & Current actor parameters with respect to which gradients are taken. \\
\thetar & Frozen behavior-policy snapshot that generated the stored rollout states and actions. \\
\thetastop=\operatorname{sg}(\theta) & Stop-gradient copy of the current parameters: numerically equal to $\theta$ but treated as constant. In general, $\thetastop\neq\thetar$ after the first minibatch update. \\
Q^{q_{\thetar}},\;V^{q_{\thetar}} & Discounted critic functions evaluated for the frozen rollout policy; $\gamma$ is fixed and therefore not repeated as an index. \\
\Tau,\;\alpha=1/\Tau & Entropy temperature and inverse temperature. \\
\gamma,\;H & Infinite-horizon discount factor and the corresponding geometric terminal index, with $\Pr(H=h)=(1-\gamma)\gamma^h$. \\
{d^{q_\theta}(s)} & {Normalized discounted state-visitation density for $s_0\sim p(s_0)$.} \\
\bar\gamma(k),\;d_{\mathrm{DA}}^{q_\theta} & Exact phase-dependent DA-MDP discount and normalized discounted augmented-state visitation; $\bar\gamma(k)=1$ for $k>1$ and $\bar\gamma(1)=\gamma$. \\
\mathcal J,\;\mathcal J_{\mathrm{DA}} & Discounted maximum-entropy returns for the original and diffusion-augmented MDPs; both are maximized. \\
{q_\theta^{(k)}(a^k\mid s)} & {Noisy-action marginal at diffusion phase $k$, generated from the shared prior using the reverse kernels.} \\
{\mathcal L_{\mathrm{DA}}(\theta;\thetar)} & {Local denoising loss with frozen rollout critic and visitation; its decrease predicts the return gain bounded in App.~\ref{app:da_monotonic_improvement}.} \\
\Dkl(p(X)\|r(X)) & Ordinary KL divergence; its scope is exactly the random-variable tuple $X$ shown in both arguments. \\
D_{\mathrm{KL}}^{\mathrm u}(q\|\widetilde p) & Unnormalized-target KL (variational free energy), $\int q\log(q/\widetilde p)$; used only for the learned-forward local surrogate. \\
X=a^{(0)}\;\text{vs.}\;X=a^{(0:K)} & Action marginal versus full diffusion path. \\
X=a_{0:T}\;\text{vs.}\;X=(a_{0:T},s_{0:T+1}) & Action-sequence marginal versus joint environment state--action trajectory. \\
X=(a_{0:T}^{0:K},s_{0:T+1}) & Joint environment--diffusion trajectory; conditioning such as $a_t\mid s_t$ denotes a per-state KL. \\
\bottomrule
\end{tabularx}
\end{table}
\endgroup

\section{Details of the Relevant Reinforcement-Learning Algorithms}
\label{app:RL_algos}

\subsection{Proximal Policy Optimization (PPO)}
Proximal Policy Optimization (PPO) can be derived by applying the log-derivative trick to the Maximum Entropy surrogate objective (see Eq.\ref{eq:surroga_loss}) combined with importance sampling. Importance sampling is required because PPO does not learn a state--action value function $Q_\theta$ directly; thus, returns or rewards $R_{\mathrm{env}}(s_t,a_t)$ can only be evaluated at state--action pairs observed under the behavior policy. The resulting surrogate objective can be written as
\begin{align*}
\mathcal{L}_\text{ME}(\theta, s_t)
= \mathbb{E}_{  a_t \sim q_{\thetar}(a_t|s_t) }
\Big[ \frac{q_\theta(a_t|s_t)}{q_{\thetar}(a_t|s_t)}  \Big( \Tau \log q_\theta(a_t|s_t) - Q^{q_{\thetar}}(s_t,a_t) \Big) \Big ] + C
\,
\end{align*}

where $q_{\thetar}$ denotes the behavior policy used to collect trajectories. As the gradient is computed with respect to $\theta$ this equation effectively implements the log-derivative trick as 

\begin{equation}
\begin{aligned}
   \nabla_\theta \frac{q_\theta(a_t|s_t)}{q_{\thetar} (a_t|s_t)} &= \nabla_\theta \exp \big (\log q_\theta(a_t|s_t) - \log q_{\thetar}(a_t|s_t) \big ) \\ &= \exp \big (\log q_\theta(a_t|s_t) - \log q_{\thetar}(a_t|s_t) \big ) \nabla_\theta  \log q_\theta(a_t|s_t)  \\ & = \frac{q_\theta(a_t|s_t)}{q_{\thetar} (a_t|s_t)} \nabla_\theta  \log q_\theta(a_t|s_t)
   \end{aligned}
\end{equation}

The PPO objective, however, additionally uses a clipping strategy, where the objective is then given by:
\begin{align*}
\mathcal{L}_\mathrm{PPO}(\theta, s_t)
= -\mathbb{E}_{  a_t \sim q_{\thetar}(a_t|s_t) }
\Big[
\min \Big( r_t(\theta)\, \hat{A}_t, \, \mathrm{clip}(r_t(\theta), 1-\epsilon, 1+\epsilon)\, \hat{A}_t \Big)
\Big],
\end{align*}
where $r_t(\theta) = \frac{q_\theta(a_t \mid s_t)}{q_{\thetar}(a_t \mid s_t)}$ are importance weights, $\hat{A}_t$ is an estimate of the advantage function, and $\epsilon$ is the clipping parameter. The advantage function $A_t$ quantifies how much better taking action $a_t$ in state $s_t$ is compared to the expected value of the policy from that state and is defined as
$A_t = Q^{\thetar}(s_t,a_t) - V^{\thetar}(s_t) - \Tau \log q_\theta(a_t | s_t)$ where the subtraction of $V^{\thetar}(s_t)$ reduces the variance of the gradient estimates while leaving the average gradient unchanged. {The estimate $\hat A_t$ is treated as constant when differentiating the clipped surrogate.}

PPO evaluates the surrogate on frozen rollout states using importance-weighted actions. At the rollout parameters, its population gradient matches the local actor objective. Clipping discourages large policy updates and provides a first-order alternative to KL-based trust-region control.

\subsection{Relative Entropy Pathwise Policy Optimization (REPPO)}
\label{app:reppo}
Relative Entropy Pathwise Policy Optimization (REPPO) combines the local actor objective in Eq.~\ref{eq:surroga_loss} with an entropy target and a KL trust region. Holding the rollout critic fixed, its constrained policy-improvement problem is
\begin{equation}
\begin{aligned}
\min_\theta\quad &-\mathbb E_{s\sim d^{q_{\thetar}},\,a\sim q_\theta}
[Q^{q_{\thetar}}(s,a)]\\
\text{subject to}\quad
&\mathbb E_{s\sim d^{q_{\thetar}}}
[D_{\mathrm{KL}}(q_{\thetar}(\cdot\mid s)\|q_\theta(\cdot\mid s))]
\leq\delta_{\mathrm{KL}},\\
&\mathbb E_{s\sim d^{q_{\thetar}}}[\mathcal H[q_\theta(\cdot\mid s)]]
\geq\mathcal H_{\mathrm{tar}}.
\end{aligned}
\end{equation}

\paragraph{Lagrangian formulation.}
The temperature $\Tau\geq0$ is the entropy-constraint multiplier, while $\lambda_{\mathrm{KL}}\geq0$ controls the KL constraint. Their Lagrangian is
\begin{equation}
\begin{aligned}
\mathcal L(\theta,\lambda_{\mathrm{KL}},\Tau)
={}&\mathbb E_{s\sim d^{q_{\thetar}},\,a\sim q_\theta}
[\Tau\log q_\theta(a\mid s)-Q^{q_{\thetar}}(s,a)]
+\Tau\mathcal H_{\mathrm{tar}}\\
&+\lambda_{\mathrm{KL}}\left(
\mathbb E_{s\sim d^{q_{\thetar}}}
[D_{\mathrm{KL}}(q_{\thetar}(\cdot\mid s)\|q_\theta(\cdot\mid s))]
-\delta_{\mathrm{KL}}\right).
\end{aligned}
\end{equation}
For fixed $\Tau$, the first expectation equals the temperature-scaled reverse-KL actor objective up to a $\theta$-independent normalizer. Entropy therefore enters once, through $\Tau\log q_\theta$.

\paragraph{Policy update.}
Holding the critic, rollout states, and multipliers fixed, the policy update is
\begin{equation}
\theta_{i+1}=\theta_i-\eta_\theta
\nabla_\theta\mathcal L(\theta_i,\lambda_{\mathrm{KL}}^i,\Tau^i).
\end{equation}
We parameterize $\lambda_{\mathrm{KL}}=\exp(\alpha_{\mathrm{KL}})$ and $\Tau=\exp(\alpha_\Tau)$ to keep both multipliers positive.

\paragraph{Dual updates.}
With policy statistics held fixed, gradient ascent in the log-multipliers gives
\begin{equation}
\alpha_{\mathrm{KL}}^{i+1}
=\alpha_{\mathrm{KL}}^i+\eta_{\mathrm{KL}}\lambda_{\mathrm{KL}}^i
\left(\mathbb E_{s\sim d^{q_{\thetar}}}
[D_{\mathrm{KL}}(q_{\thetar}(\cdot\mid s)\|q_{\theta_i}(\cdot\mid s))]
-\delta_{\mathrm{KL}}\right),
\end{equation}
\begin{equation}
\alpha_\Tau^{i+1}
=\alpha_\Tau^i+\eta_\Tau\Tau^i
\left(\mathcal H_{\mathrm{tar}}
-\mathbb E_{s\sim d^{q_{\thetar}}}[\mathcal H[q_{\theta_i}(\cdot\mid s)]]\right).
\end{equation}
Thus the KL penalty increases when its constraint is exceeded, and the temperature increases when entropy falls below its target.

\paragraph{Q-target estimation via TD($\lambda$).}
REPPO directly learns an action-value function $Q_\phi(s,a)$ and does not employ a separate state-value function. Consequently, action-value targets are computed using multi-step temporal-difference updates with eligibility traces, i.e.\ TD($\lambda$).

\paragraph{Reparameterized policy gradient.}
REPPO uses a learned Q-function together with the \emph{reparameterization trick} to differentiate through freshly sampled actions. Actions are generated as
\begin{align}
a_t = \mu_\theta(s_t) + {\xi_t}\,\sigma_\theta(s_t):=f_\theta(\xi_t;s_t), \quad \xi_t\sim\mathcal N(0,I),
\end{align}
allowing gradients to flow directly through sampled actions. The resulting policy-gradient estimator takes the form
\begin{align}
\nabla_\theta \mathcal{L}_{\mathrm{REPPO}}(\theta, s_t)
=
\mathbb{E}_{\xi_t}
\Big[
\Tau \, \nabla_\theta \log q_\theta(f_\theta(\xi_t; s_t) \mid s_t)
-
\nabla_\theta Q^{q_{\thetar}}\!\left(s_t,f_\theta(\xi_t;s_t)\right)
\Big].
\end{align}

\paragraph{KL-aware clipped surrogate objective.}
To enforce the KL trust region in practice, REPPO employs a \emph{sample-wise clipped loss} that selects between the local reverse-KL actor objective and a pure trust-region KL penalty. For each minibatch state $s_t^{(i)}$, define the behavior-to-current Monte Carlo estimate explicitly as
\[
\widehat{\Dkl}^{(i)}
:=
\widehat{\Dkl}\!\left(
q_{\thetar}(a_t\mid s_t^{(i)})
\,\middle\|\,
q_\theta(a_t\mid s_t^{(i)})
\right).
\]

The final REPPO surrogate loss can then be written equivalently as
\begin{equation}
\mathcal{L}_{\mathrm{REPPO}}^{\mathrm{clip}}(\theta)
=
\mathbb{E}_{s_t^{(i)} \sim q_{\thetar}}
\left[
\begin{cases}
\mathcal{L}_{\mathrm{ME}}^{(i)}(\theta, s_t^{(i)}),
& \text{if } \widehat{\Dkl}^{(i)} \le \delta_{\mathrm{KL}}, \\[6pt]
\lambda_{\mathrm{KL}} \, \widehat{\Dkl}^{(i)},
& \text{if } \widehat{\Dkl}^{(i)} > \delta_{\mathrm{KL}} .
\end{cases}
\right],
\label{eq:reppo_clipping}
\end{equation}

where $\delta_{\mathrm{KL}}$ is a predefined trust-region threshold, $\lambda_{\mathrm{KL}}$ is a KL penalty coefficient, and ${\mathcal{L}_{\mathrm{ME}}^{(i)}(\theta, s_t^{(i)})}$ denotes the reverse-KL surrogate loss evaluated at sample $i$.

The adaptive rule follows the reverse-KL update while the empirical divergence remains within the prescribed trust region and switches to a KL penalty once the boundary is exceeded. It should therefore be interpreted as a practical mechanism for keeping the sampled update within the regime motivated by the {theoretical improvement bound}.

\paragraph{Auxiliary tasks.}
REPPO optionally employs auxiliary objectives to stabilize representation learning in sparse-reward regimes and when using small update batches. In particular, it adopts the latent self-prediction auxiliary task introduced in the REPPO paper, which has been shown to significantly improve critic stability and sample efficiency in low-signal settings (see also \citealp{jaderberg2016reinforcement}). 
Concretely, the critic is augmented with an auxiliary prediction head that encourages the learned latent representation of state--action pairs to be predictive of future representations under the on-policy data distribution. The auxiliary loss is optimized jointly with the critic objective using on-policy rollouts. We refer to the REPPO paper for the precise formulation and architectural details.

\paragraph{Cross-entropy regression for critic targets.}
REPPO further stabilizes critic learning by replacing mean-squared-error regression with a cross-entropy-based regression loss (HL-Gauss; \citealp{farebrother2024stop}), originally inspired by distributional reinforcement learning methods \citep{bellemare2017distributional}. All critic targets, including those used for policy optimization, are regressed using this loss.
This formulation has been shown to yield improved numerical stability and learning dynamics, particularly in deterministic and sparse-reward environments. We refer the reader to the REPPO paper for implementation details and ablation results.

\subsection{Maximum-Entropy Wasserstein Policy Optimization (ME-WPO)}
\label{app:me_wpo_algorithm}
\begingroup
ME-WPO uses the same Gaussian policy, $Q$-critic, TD($\lambda$) targets, trust-region constraint, temperature update, and auxiliary losses as REPPO. It changes only the actor direction: instead of differentiating the reverse-KL objective through a reparameterized action, ME-WPO follows the Wasserstein gradient flow of the same soft policy-improvement objective. For a fresh action $a_t\sim q_\theta(\cdot\mid s_t)$ that is held fixed during differentiation, this direction is obtained from the score-matching loss
\begin{equation}
\widehat{\mathcal L}_{\mathrm{ME\text{-}WPO}}(\theta)
=
\frac{\Tau}{2}
\mathbb E_{\substack{s_t\sim q_{\thetar}\\
a_t\sim q_{\thetastop}(\cdot\mid s_t)}}
\left[
\left\|
\nabla_{a_t}\!\left(
\log q_\theta(a_t\mid s_t)
-\frac{1}{\Tau}Q^{q_{\thetar}}(s_t,a_t)
\right)
\right\|^2
\right].
\label{eq:me_wpo_algorithm_loss}
\end{equation}
App.~\ref{app:WPO_deriv} derives Eq.~\ref{eq:me_wpo_algorithm_loss} from the Wasserstein gradient flow of the soft policy-improvement objective and its projection onto the parametric policy family. Let $\widehat{\mathcal L}_{\mathrm{ME\text{-}WPO}}^{(i)}(\theta)$ denote the statewise contribution to Eq.~\ref{eq:me_wpo_algorithm_loss}, including the expectation over fresh actions at $s_t^{(i)}$. Using the behavior-to-current KL estimate $\widehat{\Dkl}^{(i)}$ defined above, the complete conditional preconditioned actor gradient is
\begin{equation}
\widetilde{\nabla}_\theta
\mathcal L_{\mathrm{ME\text{-}WPO}}^{\mathrm{clip}}(\theta)
=
\mathbb E_{s_t^{(i)}\sim q_{\thetar}}
\left[
\begin{cases}
\mathcal F_{\theta\theta}^{-1}
\nabla_\theta\widehat{\mathcal L}_{\mathrm{ME\text{-}WPO}}^{(i)}(\theta),
& \text{if }\widehat{\Dkl}^{(i)}\leq\delta_{\mathrm{KL}},\\[6pt]
\lambda_{\mathrm{KL}}\nabla_\theta\widehat{\Dkl}^{(i)},
& \text{if }\widehat{\Dkl}^{(i)}>\delta_{\mathrm{KL}}.
\end{cases}
\right]
\label{eq:me_wpo_clipped_gradient}
\end{equation}
Here, $\mathcal F_{\theta\theta}$ is the policy Fisher matrix defined in Eq.~\ref{eq:me_wpo_fisher}, and the branch condition and $\lambda_{\mathrm{KL}}$ are held fixed during the actor update, exactly as in REPPO. Thus, states inside the trust region follow the inverse-Fisher-preconditioned score gradient, whereas states outside it contribute only the ordinary KL-penalty gradient. Within the score-matching branch, the policy score $\nabla_{a_t}\log q_\theta(a_t\mid s_t)$ is matched to the soft target score $\nabla_{a_t}Q^{q_{\thetar}}(s_t,a_t)/\Tau$ without evaluating the target normalizer. In mean--standard-deviation coordinates, the Gaussian preconditioner of \citet{pfau2025wasserstein} implements the inverse-Fisher factor through the element-wise transformations
\begin{equation}
\widetilde{\nabla}_{\mu_\theta}\widehat{\mathcal L}_{\mathrm{ME\text{-}WPO}}
=
\sigma_\theta^2\odot
\nabla_{\mu_\theta}\widehat{\mathcal L}_{\mathrm{ME\text{-}WPO}},
\qquad
\widetilde{\nabla}_{\sigma_\theta}\widehat{\mathcal L}_{\mathrm{ME\text{-}WPO}}
=
\frac{1}{2}\sigma_\theta^2\odot
\nabla_{\sigma_\theta}\widehat{\mathcal L}_{\mathrm{ME\text{-}WPO}}.
\label{eq:me_wpo_gaussian_preconditioner}
\end{equation}
Apart from this score-based actor gradient and inverse-Fisher scaling, ME-WPO follows the REPPO implementation described above. Standard WPO is recovered by removing the entropy contribution, whereas ME-WPO retains REPPO's automatically tuned positive temperature.
\endgroup

\section{DA-MDP Algorithms}
\label{app:DA_MDP_algos}
\label{app:implem_details}
\label{app:hyperparameters}
\begingroup

\begingroup
\subsection{From the DA-MDP to the Implementation}
\label{sec:implemented_algorithms}

The implementation follows the construction in Sec.~\ref{sec:MDP} directly. It replaces each environment transition by $K$ local denoising transitions and stores them in an ordinary on-policy rollout buffer. The buffer is flattened over environment time and diffusion time, with leading dimensions $[TK,N_{\mathrm{env}}]$; the explicit chain dimension is restored only when a loss must be summed over all $K$ diffusion steps. Consequently, actor and critic minibatches contain local transitions and never require backpropagation through a complete diffusion chain.

\paragraph{Rollout collection.}
Let $\thetar$ denote the frozen behavior parameters used during collection. At environment state $s_t$, the implementation samples $a_t^K\sim q_{\mathrm{prior}}$, applies the reverse kernels successively for $k=K,\ldots,1$, and stores each transition $(s_t,a_t^k,k,a_t^{k-1})$. It additionally stores the reverse and forward log densities, critic predictions, termination flags, and
\begin{equation}
r_{\tilde t}=\mathbf 1_{\{k=1\}}R_{\mathrm{env}}(s_t,a_t^0),
\qquad
\ell_{\tilde t}=\log\frac{q_{\thetar}(a_t^{k-1}\mid a_t^k,s_t)}
{\pi_{\thetar}(a_t^k\mid a_t^{k-1},s_t)}.
\label{eq:implementation_stored_terms}
\end{equation}
For $k>1$, the successor is formed in memory without calling the environment. For Gaussian policies, the buffer stores the terminal latent $x_t^0$, while the environment executes $a_t^0=\tanh(x_t^0)$; its reward and episode boundary are attached to the final row of the chain. A new prior sample initializes the next chain. For the transformed-action entropy formulation, the terminal densities in the DA-MDP objective are evaluated in executed-action coordinates:
\[
q_\theta(a_t^0\mid a_t^1,s_t)
=q_\theta^X(x_t^0\mid a_t^1,s_t)e^{-j_{\tanh}(x_t^0)},
\qquad
\pi_\theta(a_t^1\mid a_t^0,s_t)
=\pi_\theta^X(a_t^1\mid x_t^0,s_t),
\]
where $q_\theta^X,\pi_\theta^X$ are latent-coordinate densities and
$j_{\tanh}(x^0)=\sum_i\log(1-\tanh^2(x_i^0))$.
The transformed-action formulation therefore stores the latent terminal log ratio minus $j_{\tanh}(x_t^0)$. The latent-entropy formulation instead stores the latent terminal log ratio without this correction. The selected convention is used consistently in the actor loss, critic soft rewards, and temperature statistic (App.~\ref{app:action_transform_entropy_ablation}).

\paragraph{Environment-time discounting.}
\label{app:discount_rescaling}
Two conventions are possible. The exact, phase-dependent convention defines $\bar\gamma(k)=1$ for every internal denoising transition $k>1$ and $\bar\gamma(1)=\gamma$ for the transition that executes the action and advances environment time. It assigns every reward and log-ratio term associated with environment time $t$ the intended weight $\gamma^t$, and therefore exactly preserves the discounted DA-MDP objective. An alternative constant convention sets $\bar\gamma(k)=\gamma^{1/K}$ for every phase. It gives one factor of $\gamma$ between consecutive environment rewards, but it additionally phase-weights the intermediate log-ratio terms and therefore corresponds to a different augmented objective. We found that the exact phase-dependent convention, $\bar\gamma(k)=1$ for $k>1$ and $\bar\gamma(1)=\gamma$, performs better; we use it throughout the paper and in all reported experiments. The trace parameter $\lambda$ is tuned separately; empirically, we find that it should be chosen slightly larger than for Gaussian policies, for example $\lambda=0.98$.

\subsection{Critic Targets}
\label{app:critic_training_details}

\dareppo{} and \dawpo{} learn an action-value function, whereas \dappo{} learns a state-value function. Both critics are trained from trajectories recorded under the rollout policy and targets computed over the flattened augmented trajectory. We describe the two cases separately below.
The log ratios $\ell_{\tilde t}$ follow the entropy convention selected for the actor: the transformed-action formulation includes the terminal soft-reward contribution $+\Tau j_{\tanh}(x_t^0)$, whereas the latent-entropy formulation omits it (App.~\ref{app:action_transform_entropy_ablation}).

\subsubsection{Q-Function Training}
\label{app:q_function_training}

For \dareppo{} and \dawpo{}, the critic estimates $Q_{\omega}(\tilde s_{\tilde t},\tilde a_{\tilde t})$ for the sampled local reverse-diffusion action. The soft $Q$-function excludes the entropy cost of its own action. Consequently, its Bellman target contains the reverse--forward log-ratio of the \emph{successor} action. Let $\bar\omega$ denote the critic parameters used during rollout collection, let $z_{\tilde t}$ indicate a true environment termination, and define the stored bootstrap value
\begin{equation}
B_{\tilde t}^{Q}
:=Q_{\bar\omega}(\tilde s_{\tilde t+1},\tilde a_{\tilde t+1}),
\qquad
\bar r_{\tilde t}^{Q}
:=r_{\tilde t}
-\gamma_{\tilde t}(1-z_{\tilde t})\Tau\ell_{\tilde t+1}.
\label{eq:q_critic_bootstrap_reward}
\end{equation}
Within a diffusion chain, $(\tilde s_{\tilde t+1},\tilde a_{\tilde t+1})$ is simply the next stored local transition. At the last diffusion step, the implementation samples the first local transition of the next chain and evaluates both $B_{\tilde t}^{Q}$ and $\ell_{\tilde t+1}$ using that same sample. Thus, for $\lambda=0$, the target reduces exactly to
\begin{equation}
y_{\tilde t}^{Q,(1)}
=r_{\tilde t}
+\gamma_{\tilde t}(1-z_{\tilde t})
\left[B_{\tilde t}^{Q}-\Tau\ell_{\tilde t+1}\right].
\label{eq:q_critic_one_step_target}
\end{equation}
For the multi-step target used in the experiments, let $b_{\tilde t}$ indicate any episode boundary, including either a true termination or a time-limit truncation. The buffer computes TD-$\lambda$ backwards as
{
\begin{equation}
  \begin{aligned}
  \delta_{\tilde t}^{Q}
&=\bar r_{\tilde t}^{Q}
+\gamma_{\tilde t}(1-z_{\tilde t})B_{\tilde t}^{Q}
-Q_{\bar\omega}(\tilde s_{\tilde t},\tilde a_{\tilde t}),\\
  A_{\tilde t}^{Q,\lambda}
&=\delta_{\tilde t}^{Q}
+\gamma_{\tilde t}\lambda_{\tilde t}(1-b_{\tilde t})
A_{\tilde t+1}^{Q,\lambda}, \qquad y_{\tilde t}^{Q,\lambda}
=Q_{\bar\omega}(\tilde s_{\tilde t},\tilde a_{\tilde t})
+A_{\tilde t}^{Q,\lambda}.
  \end{aligned}
  \label{eq:q_critic_td_lambda_target}
\end{equation}
}
A true termination sets $z_{\tilde t}=b_{\tilde t}=1$, eliminating both the bootstrap and continuation of the trace. A time-limit truncation sets $z_{\tilde t}=0$ and $b_{\tilde t}=1$: it ends the trace but still bootstraps from the terminal observation recorded before the environment is reset. This prevents returns from leaking across episode boundaries without incorrectly treating a time limit as a terminal state.

The Q-network is regressed onto $y_{\tilde t}^{Q,\lambda}$ while the actor is frozen. In the reported experiments, an HL-Gauss head encodes each scalar target as a Gaussian-smoothed categorical distribution over a fixed support, and the critic minimizes the corresponding cross-entropy loss \citep{farebrother2024stop}. The Q update precedes the actor update in every minibatch, so the \dareppo{} and \dawpo{} objectives use the most recent critic while keeping its parameters fixed during actor differentiation.

\subsubsection{Value-Function Training}
\label{app:value_function_training}

\dappo{} instead estimates $V_{\omega}(\tilde s_{\tilde t})$. Since a value function averages over the current local action, the current reverse--forward log-ratio belongs to its immediate soft reward,
\begin{equation}
\bar r_{\tilde t}^{V}:=r_{\tilde t}-\Tau\ell_{\tilde t}.
\label{eq:value_critic_soft_reward}
\end{equation}
During rollout collection, the implementation records both $V_{\bar\omega}(\tilde s_{\tilde t})$ and the successor bootstrap $B_{\tilde t}^{V}:=V_{\bar\omega}(\tilde s_{\tilde t+1})$. It then computes generalized advantage estimates and value targets using the same termination and boundary masks as above:
{
\begin{equation}
  \begin{aligned}
  \delta_{\tilde t}^{V}
&=\bar r_{\tilde t}^{V}
+\gamma_{\tilde t}(1-z_{\tilde t})B_{\tilde t}^{V}
-V_{\bar\omega}(\tilde s_{\tilde t}),\\
  A_{\tilde t}^{V,\lambda}
&=\delta_{\tilde t}^{V}
+\gamma_{\tilde t}\lambda_{\tilde t}(1-b_{\tilde t})
A_{\tilde t+1}^{V,\lambda}, \qquad y_{\tilde t}^{V,\lambda}
=V_{\bar\omega}(\tilde s_{\tilde t})
+A_{\tilde t}^{V,\lambda}.
  \end{aligned}
  \label{eq:value_critic_gae_target}
\end{equation}
}
We discount only when the environment advances: $\gamma_{\tilde t}=1$ during internal denoising steps ($k>1$), and $\gamma_{\tilde t}=\gamma$ when the final step ($k=1$) executes the action (App.~\ref{app:discount_rescaling}). Thus, denoising introduces no additional discount. The trace parameter $\lambda_{\tilde t}$ controls how strongly later temporal-difference errors contribute to the current advantage. We train the value head on $y_{\tilde t}^{V,\lambda}$ using HL-Gauss cross-entropy and use $A_{\tilde t}^{V,\lambda}$ to construct the \dappo{} actor advantage. As in PPO, we update the actor using the frozen rollout critic, then fit the value function with the actor frozen.

\endgroup
\subsection{\dareppo{}}
\label{app:da_reppo}

\dareppo{} applies REPPO to the local transitions stored in the augmented rollout buffer. The construction does not require a particular reverse or forward diffusion kernel. For any reparameterizable reverse kernel, write a fresh local action as
\begin{equation}
\hat a_\theta^{k-1}:=g_\theta(\epsilon;\tilde s),
\qquad
\epsilon\sim\rho(\epsilon),
\qquad
\tilde s=(s_t,a_t^k,k),
\label{eq:da_reppo_general_reparameterization}
\end{equation}
where the base distribution $\rho$ is independent of $\theta$. The sample-wise maximum-entropy policy loss is
\begin{equation}
f_\theta(\tilde s,a^{k-1})
:=\Tau\log\frac{
q_\theta(a^{k-1}\mid a_t^k,s_t)}{
\pi_\theta(a_t^k\mid a^{k-1},s_t)}
-Q_{\mathrm{DA}}^{q_{\thetar}}(\tilde s,a^{k-1}).
\label{eq:da_reppo_local_loss}
\end{equation}
The reverse-kernel term encourages entropy in each denoising decision, the forward-kernel term enforces consistency with the prescribed noising process, and the critic favors local actions with high predicted return. The unconstrained actor objective is therefore
\begin{equation}
\mathcal L_{\mathrm{DA:REPPO}}^{\mathrm{uncon}}(\theta)
=\mathbb E_{\substack{
\tilde s\sim d_{\mathrm{DA}}^{q_{\thetar}}\\
\epsilon\sim\rho}}
\!\left[
f_\theta\!\left(\tilde s,g_\theta(\epsilon;\tilde s)\right)
\right].
\label{eq:diff_reppo_loss}
\end{equation}
A minibatch average of Eq.~\ref{eq:diff_reppo_loss} estimates the unnormalized local reverse-KL objective in Eq.~\ref{eq:diffusion_surrogate}; no state-dependent partition function is evaluated. The augmented-state distribution, rollout parameters $\thetar$, and critic parameters are held fixed, while gradients pass through the reparameterized action, the reverse and forward log densities, and the critic's action input. Consequently, learned forward-kernel parameters receive the correction implied by their log density.

\dareppo{} measures the change in each reverse kernel using the rollout-to-current KL,
\begin{equation}
d_\theta(\tilde s)
:=D_{\mathrm{KL}}\!\left(
q_{\thetar}(\cdot\mid\tilde s)\,\middle\|\,
q_\theta(\cdot\mid\tilde s)
\right),
\qquad
\widehat D_\theta^{\mathrm{chain}}(\tilde s):=K d_\theta(\tilde s).
\label{eq:da_reppo_chain_kl}
\end{equation}
Each local KL can be computed analytically or estimated by sampling, depending on the kernel family. For uniformly sampled diffusion steps, the minibatch average of $K d_\theta$ estimates the expected sum of KLs over all $K$ steps. Actor clipping uses the scaled KL of each sample separately. With bound $\delta_{\mathrm{KL}}$ and multiplier $\lambda_{\mathrm{KL}}\geq0$, the actor loss is
\begin{equation}
\mathcal L_{\mathrm{actor}}^{\mathrm{REPPO}}(\theta)
=\mathbb E_{\substack{
\tilde s\sim d_{\mathrm{DA}}^{q_{\thetar}}\\
\epsilon\sim\rho}}
\!\left[
\begin{cases}
f_\theta\!\left(\tilde s,g_\theta(\epsilon;\tilde s)\right),
&\widehat D_\theta^{\mathrm{chain}}\leq\delta_{\mathrm{KL}},\\
\lambda_{\mathrm{KL}}\widehat D_\theta^{\mathrm{chain}},
&\widehat D_\theta^{\mathrm{chain}}>\delta_{\mathrm{KL}}.
\end{cases}
\right].
\label{eq:da_reppo_clipped_actor}
\end{equation}
The branch indicator and $\lambda_{\mathrm{KL}}$ are stopped during the policy update. An in-bound sample therefore follows the maximum-entropy policy gradient, whereas an out-of-bound sample contributes only the trust-region gradient. The KL multiplier and temperature are updated separately by the dual objectives in Sec.~\ref{app:temp_tuning}.

\paragraph{Gaussian trust region.}
For the Gaussian reverse kernels of the VP-SDE (App.~\ref{app:vp_sde}), the rollout-to-current KL has a closed form. At each diffusion step, let $\mu_{\thetar},\sigma_{\thetar}^2$ denote the rollout mean and diagonal variance, and $\mu_\theta,\sigma_\theta^2$ their current values. Then
\begin{equation}
d_\theta(\tilde s)
=\frac12\sum_{i=1}^{N}\left[
\log\frac{\sigma_{\theta,i}^2}{\sigma_{\thetar,i}^2}
+\frac{\sigma_{\thetar,i}^2
+(\mu_{\thetar,i}-\mu_{\theta,i})^2}
{\sigma_{\theta,i}^2}
-1
\right].
\label{eq:da_reverse_kernel_closed_form_kl}
\end{equation}
Both \dareppo{} and \dawpo{} evaluate the rollout-to-current KL at sampled diffusion steps and scale it by $K$ (Eq.~\ref{eq:da_reppo_chain_kl}). The KL multiplier is updated using the minibatch average, while actor clipping is applied separately to each sample.

\subsection{\dawpo{}}
\label{app:da_wpo}

\dawpo{} uses the same augmented rollout, Q critic, KL trust region, and dual variables as \dareppo{}. It replaces only the in-bound maximum-entropy policy direction by a Wasserstein score-matching direction. In the following equations, $\tilde s=(s_t,a_t^k,k)\sim d_{\mathrm{DA}}^{q_{\thetar}}$ and $a^{k-1}\sim q_{\thetastop}(\cdot\mid a_t^k,s_t)$, with the sampled action held fixed during differentiation. For any differentiable reverse and forward kernels, the score-matching loss is
{\eqsizetiny
\begin{equation}
\mathcal L_{\mathrm{DA:WPO}}(\theta)
:=\frac{\Tau}{2}\mathbb E_{\tilde s,a^{k-1}}\!\left[
\left\|\nabla_{a^{k-1}}\!\left(
\log\frac{q_\theta(a^{k-1}\mid\tilde s)}{\pi_\theta(a_t^k\mid a^{k-1},s_t)}
-\alpha Q_{\mathrm{DA}}^{q_{\thetar}}(\tilde s,a^{k-1})
\right)\right\|_2^2\right].
\label{eq:diffmaxent_wpo_main_full}
\end{equation}
}
{For Gaussian policies, the implementation takes score derivatives in unsquashed coordinates $x^{k-1}$. At $k=1$, the differentiated terminal log ratio includes $-j_{\tanh}(x^0)$ under transformed-action entropy and omits it under latent entropy. The critic uses the same entropy convention, and its terminal action dependence is differentiated with respect to $x^0$. The implemented score norm thus uses latent-coordinate geometry.}

This loss directly compares the reverse-kernel score with the forward-kernel score plus the temperature-scaled critic gradient. It requires only that the two kernel densities and their action derivatives can be evaluated locally; it does not depend on a Gaussian or SDE parameterization. Because the critic is nonlinear in $a^{k-1}$, \dawpo{} generally samples the local action even when the kernel scores themselves are analytic. With the sampled action and critic held fixed, the parameter gradient is
{\fontsize{6.5pt}{7.5pt}\selectfont
\begin{equation}
\nabla_\theta\mathcal L_{\mathrm{DA:WPO}}(\theta)
=\mathbb E_{\tilde s,a^{k-1}}\!\left[
\left(\nabla_\theta\nabla_{a^{k-1}}\log\frac{q_\theta(a^{k-1}\mid\tilde s)}{\pi_\theta(a_t^k\mid a^{k-1},s_t)}\right)
\nabla_{a^{k-1}}\!\left(\Tau\log\frac{q_\theta(a^{k-1}\mid\tilde s)}{\pi_\theta(a_t^k\mid a^{k-1},s_t)}-Q_{\mathrm{DA}}^{q_{\thetar}}(\tilde s,a^{k-1})\right)
\right].
\label{eq:da_wpo_parameter_gradient}
\end{equation}
}
The implementation obtains Eq.~\ref{eq:da_wpo_parameter_gradient} from the explicit first-order surrogate
{\eqsizetiny
\begin{equation}
\widehat f_\theta^{\mathrm{DA:WPO}}
:=\operatorname{sg}\!\left[\nabla_{a^{k-1}}\!\left(
\Tau\log\frac{q_\theta(a^{k-1}\mid\tilde s)}{\pi_\theta(a_t^k\mid a^{k-1},s_t)}
-Q_{\mathrm{DA}}^{q_{\thetar}}(\tilde s,a^{k-1})\right)\right]^{\!\top}
\nabla_{a^{k-1}}\log\frac{q_\theta(a^{k-1}\mid\tilde s)}{\pi_\theta(a_t^k\mid a^{k-1},s_t)}.
\label{eq:da_wpo_stopped_residual}
\end{equation}
}
{\eqsizetiny
\begin{equation}
\widehat{\mathcal L}_{\mathrm{DA:WPO}}(\theta)
=\mathbb E_{\tilde s,a^{k-1}}\!\left[\widehat f_\theta^{\mathrm{DA:WPO}}\right].
\label{eq:da_wpo_stopped_objective}
\end{equation}
}
The stop-gradient operator avoids derivatives through the critic gradient while retaining derivatives through the reverse and forward kernel scores.

Combining the score loss with the clipped reverse-kernel trust region gives the final \dawpo{} actor loss
\begin{equation}
\mathcal L_{\mathrm{actor}}^{\mathrm{DA:WPO}}(\theta)
=\mathbb E\!\left[
\begin{cases}
\widehat f_\theta^{\mathrm{DA:WPO}},
&\widehat D_\theta^{\mathrm{chain}}\leq\delta_{\mathrm{KL}},\\
\lambda_{\mathrm{KL}}\widehat D_\theta^{\mathrm{chain}},
&\widehat D_\theta^{\mathrm{chain}}>\delta_{\mathrm{KL}}.
\end{cases}
\right].
\label{eq:da_wpo_clipped_actor}
\end{equation}
The branch condition and $\lambda_{\mathrm{KL}}$ are stopped. Hence, an in-bound sample follows the (optionally preconditioned) score-matching gradient, whereas an out-of-bound sample contributes only the KL-penalty gradient. When a separate forward-kernel trust region is enabled, its gate and KL penalty are applied independently.

\paragraph{Inverse-Fisher approximation.}
Computing and inverting the Fisher matrix over all network parameters is intractable for our policies. Following \citet{pfau2025wasserstein}, we instead use the inverse Fisher of the Gaussian distribution parameters as a local preconditioner. For the VP reverse kernel in Eq.~\ref{eq:vp_discrete_kernels}, this requires accounting for the coupling between the control $u_{\Psi,i}$ and diffusion coefficient $\delta_{k,i}$: both affect the mean, and $\delta_{k,i}$ also determines the variance. Expressing the Gaussian Fisher in these coordinates gives an exact $2\times2$ block for each action coordinate and diffusion step. Writing $v_{k,i}=\nu^2\delta_{k,i}$ and $h_{k,i}=\tfrac12a_{t,i}^k+\nu^2u_{\Psi,i}(s_t,a_t^k,k)$, this block is
\begin{equation}
F_{k,i}^{\mathrm{VP}}
=\begin{pmatrix}
v_{k,i} & h_{k,i}\\
h_{k,i} & h_{k,i}^2/v_{k,i}+1/(2\delta_{k,i}^2)
\end{pmatrix}.
\label{eq:da_wpo_vp_fisher_block}
\end{equation}
We stabilize each block by diagonal normalization, correlation clipping, and relative damping before inversion, then rescale its inverse to cap the largest eigenvalue. The resulting matrix $P_{k,i}$ preconditions the score-loss gradient with its entries held fixed:
\begin{equation}
\widetilde g_{k,i}=\operatorname{sg}(P_{k,i})g_{k,i},
\qquad
g_{k,i}:=\nabla_{(u_{\Psi,i},\delta_{k,i})}\widehat f_\theta^{\mathrm{DA:WPO}}.
\label{eq:da_wpo_adjusted_vp_gradient}
\end{equation}

\subsection{Trust-Region and Temperature Dual Updates}
\label{app:temp_tuning}

\dareppo{} and \dawpo{} optimize the KL multiplier in log space. Let
$\lambda_{\mathrm{KL}}=\exp(\xi_{\mathrm{KL}})$ and let
$\widehat D_{\mathrm{KL}}$ be the minibatch mean of {the estimated KL sum over diffusion steps} in Eq.~\ref{eq:da_reppo_chain_kl}. Although each local Gaussian KL is available in closed form, its sum over diffusion steps can be estimated by Monte Carlo sampling. If $\kappa$ indices $k_j$ are drawn independently from probabilities $\rho_k>0$, then
\begin{equation}
\widehat D_{\theta}^{\mathrm{chain}}
:=\frac{1}{\kappa}\sum_{j=1}^{\kappa}
\frac{d_{\theta,k_j}}{\rho_{k_j}},
\qquad
\mathbb E_{k_{1:\kappa}}\!\left[
\widehat D_{\theta}^{\mathrm{chain}}
\right]
=\sum_{k=1}^{K}d_{\theta,k}.
\label{eq:da_kl_diffusion_step_mc}
\end{equation}
For uniform sampling, $\rho_k=1/K$, so every sampled local KL receives weight $K$. The KL dual loss is
\begin{equation}
\mathcal L_{\mathrm{KL}}^{\mathrm{dual}}(\xi_{\mathrm{KL}})
:=\lambda_{\mathrm{KL}}
\left(\delta_{\mathrm{KL}}-\operatorname{sg}(\widehat D_{\mathrm{KL}})\right).
\label{eq:da_kl_dual_loss}
\end{equation}
Minimizing this loss gives the update direction
\begin{equation}
\xi_{\mathrm{KL}}
\leftarrow
\xi_{\mathrm{KL}}
+\eta_{\mathrm{KL}}\lambda_{\mathrm{KL}}
\left(\widehat D_{\mathrm{KL}}-\delta_{\mathrm{KL}}\right),
\label{eq:da_kl_dual_update}
\end{equation}
so the penalty increases precisely when {the estimated KL sum exceeds its bound}. During this dual update, the KL estimate is stopped; during the actor update, the multiplier is stopped.

For Gaussian policies, temperature tuning uses the same entropy convention as the actor and critic. The latent-entropy formulation targets a path-space lower bound on terminal latent entropy; the transformed-action formulation adds the expected terminal $\tanh$ log-Jacobian to this bound (App.~\ref{app:action_transform_entropy_ablation}). Here $x^{0:K}$ denotes the unsquashed path, and all displayed kernels are latent-coordinate densities. {At a fixed environment state $s$, the latent-space path-entropy lower bound is}
\begin{equation}
\mathcal H_{\mathrm{lower}}[q_\theta]{(s)}
={\mathcal H[q_{\mathrm{prior}}]}
+\mathbb E_{{x^{0:K}\sim q_\theta(\cdot\mid s)}}\!\left[
\sum_{k=1}^{K}\log
\frac{\pi_\theta(x^k\mid x^{k-1},s)}
{q_\theta(x^{k-1}\mid x^k,s)}
\right].
\label{eq:implementation_entropy_lower_bound}
\end{equation}
{Here, $q_\theta(x^{0:K}\mid s)
=q_{\mathrm{prior}}(x^K)\prod_{k=1}^{K}q_\theta(x^{k-1}\mid x^k,s)$
is the complete current-policy reverse path at fixed $s$.
The state-averaged bound is
$\mathcal H_{\mathrm{lower}}[q_\theta]
:=\mathbb E_{s\sim d^{q_{\theta_{\mathrm{roll}}}}}
[\mathcal H_{\mathrm{lower}}[q_\theta](s)]$.
An unbiased Monte Carlo estimator draws $s_b\sim d^{q_{\theta_{\mathrm{roll}}}}$,
then $x_b^{0:K}\sim q_\theta(\cdot\mid s_b)$, and independently
$\kappa$ indices $k_{b,j}\sim\rho$ for each path $b$:}
\begin{equation}
\begin{aligned}
\widehat{\mathcal H}_{\mathrm{lower}}
&:={\mathcal H[q_{\mathrm{prior}}]}
+\frac{1}{B\kappa}\sum_{b=1}^{B}\sum_{j=1}^{\kappa}
\frac{1}{\rho_{k_{b,j}}}
\log\frac{
\pi_\theta(x_b^{k_{b,j}}\mid x_b^{k_{b,j}-1},s_b)}
{q_\theta(x_b^{k_{b,j}-1}\mid x_b^{k_{b,j}},s_b)},\\
\mathbb E\!\left[\widehat{\mathcal H}_{\mathrm{lower}}\right]
&=\mathcal H_{\mathrm{lower}}[q_\theta].
\end{aligned}
\label{eq:da_entropy_diffusion_step_mc}
\end{equation}
{The expectation in the second line averages over the
sampled states, complete reverse paths, and diffusion indices.
In implementation, states and noisy actions $x^k$ are taken from
the rollout buffer, while local actions are sampled freshly from
$q_\theta(\cdot\mid x^k,s)$. Thus, the state average is empirical,
and reusing rollout noisy actions approximates current-policy path
sampling when $\theta\neq\theta_{\mathrm{roll}}$.}
{Only the sampled local log ratios need to be evaluated.}
Under uniform sampling, the inverse-probability factor is again $K$; for nonuniform sampling, $1/\rho_k$ preserves unbiasedness {with respect to diffusion-index sampling}. Let $\mathcal H_{\mathrm{tar}}$ be the target, and parameterize $\Tau=\exp(\xi_\Tau)$. {In the dual update below, $\widehat{\mathcal H}_{\mathrm{lower}}$ denotes the entropy estimate for the selected formulation: Eq.~\ref{eq:da_entropy_diffusion_step_mc} for latent entropy, and that estimate plus the sampled terminal log-Jacobian for transformed-action entropy (App.~\ref{app:action_transform_entropy_ablation}).} The temperature dual loss and its update direction are
\begin{align}
\mathcal L_{\Tau}^{\mathrm{dual}}(\xi_\Tau)
&:=\Tau\left(
\operatorname{sg}(\widehat{\mathcal H}_{\mathrm{lower}})
-\mathcal H_{\mathrm{tar}}
\right),
\label{eq:da_temperature_dual_loss}\\
\xi_\Tau
&\leftarrow
\xi_\Tau+\eta_\Tau\Tau
\left(\mathcal H_{\mathrm{tar}}
-\widehat{\mathcal H}_{\mathrm{lower}}\right).
\label{eq:da_temperature_dual_update}
\end{align}
Thus, the temperature increases when the estimated entropy is below target and decreases when it is above target. The exponential parameterizations keep both dual variables positive. Equations~\ref{eq:da_kl_dual_update} and~\ref{eq:da_temperature_dual_update} display the gradient-descent directions; the implementation applies them with separate Adam parameter groups and learning rates.

\subsection{\dappo{}}
\label{app:diffppo}

\dappo{} applies PPO to stored local denoising transitions. {We describe the clipped-forward variant used in the diffusion-step ablation (Mode~2 in the implementation). It combines the current soft advantage with a learned-forward-kernel correction restricted to PPO's active set.} For the augmented transition indexed by $(t,k)$, Eq.~\ref{eq:value_critic_gae_target} gives the rollout-policy advantage as
\begin{equation}
A_{t,k}^{\mathrm{roll}}
:=y_{t,k}^{V,\lambda}-V_{\bar\omega}(\tilde s_{t,k})
=A_{t,k}^{V,\lambda}.
\label{eq:da_ppo_advantage_from_target}
\end{equation}
Thus, $A_{t,k}^{\mathrm{roll}}$ is the GAE estimate obtained from the rollout soft rewards
$\bar r_{t,k}^{\thetar}=r_{t,k}-\Tau\ell_{t,k}^{\thetar}$ and the phase-dependent discounts in Sec.~\ref{app:discount_rescaling}.

The soft reward depends on the policy and must therefore be updated during the actor optimization. {For the stored rollout action $a_t^{k-1}$, define the current path log-ratio and soft reward}
\begin{equation}
\ell_{t,k}^{\theta}
:=\log q_\theta({a_t^{k-1}}\mid a_t^k,s_t)
-\log\pi_\theta(a_t^k\mid{a_t^{k-1}},s_t),
\qquad
\bar r_{t,k}^{\theta}:=r_{t,k}-\Tau\ell_{t,k}^{\theta},
\label{eq:da_ppo_current_soft_reward}
\end{equation}
Replacing the current rollout reward in the GAE estimate gives
\begin{equation}
A_{t,k}^{\theta}
=A_{t,k}^{\mathrm{roll}}
-\bar r_{t,k}^{\thetar}
+\bar r_{t,k}^{\theta}.
\label{eq:da_ppo_current_policy_advantage}
\end{equation}
{The implementation inserts this current soft reward into the stopped PPO advantage:}
\begin{equation}
{A_{t,k}^{\mathrm{cur}}
:=A_{t,k}^{\mathrm{roll}}-\bar r_{t,k}^{\thetar}+r_{t,k}-\Tau\ell_{t,k}^{\theta},}
\qquad
\widehat A_{t,k}
:=\operatorname{sg}\!\left(\frac{{A_{t,k}^{\mathrm{cur}}}-\mu_A}{\sigma_A}\right).
\label{eq:da_ppo_base_advantage}
\end{equation}
{Here, $\mu_A$ and $\sigma_A$ are the minibatch advantage mean and scale, with $\sigma_A=1$ when variance normalization is disabled. The current transition's soft reward is recomputed under the current reverse and forward diffusion kernels and then stopped inside the PPO surrogate, whereas successor soft rewards and critic targets stored at rollout time remain fixed.}

For the stored action $a_t^{k-1}$, the current-to-behavior likelihood ratio and its clipped version are
\begin{equation}
\rho_{t,k}(\theta)
:=
\frac{q_\theta(a_t^{k-1}\mid a_t^k,s_t)}
     {q_{\thetar}(a_t^{k-1}\mid a_t^k,s_t)},
\qquad
\bar\rho_{t,k}:=\operatorname{clip}(\rho_{t,k},1-\epsilon,1+\epsilon).
\label{eq:da_ppo_ratio}
\end{equation}
The standard PPO contribution is
\begin{equation}
\mathcal L_{\mathrm{clip}}(\theta)
=-\mathbb E\!\left[
\min\!\left(
\rho_{t,k}\widehat A_{t,k},
\bar\rho_{t,k}\widehat A_{t,k}
\right)
\right].
\label{eq:da_ppo_clipped_loss}
\end{equation}
{Stopping the advantage leaves the direct derivative through a learned forward kernel to be supplied separately. We restrict this correction to the same active set as the reverse-policy PPO update. The implementation uses the mask}
\begin{equation}
{
m_{t,k}
:=\mathbf 1\!\left\{
\rho_{t,k}\widehat A_{t,k}<\bar\rho_{t,k}\widehat A_{t,k}
\quad\text{or}\quad
1-\epsilon\leq\rho_{t,k}\leq1+\epsilon
\right\}.
}
\label{eq:da_ppo_active_mask}
\end{equation}
{The forward correction is evaluated on stored rollout actions. Up to terms with stopped gradients, it is}
\begin{equation}
{
\mathcal L_{\mathrm{fwd}}^{\mathrm{clip}}(\theta)
=-\frac{\Tau}{\sigma_A}\,
\mathbb E_{\mathrm{roll}}\!\left[
\operatorname{sg}(m_{t,k}\rho_{t,k})
\log\pi_\theta(a_t^k\mid a_t^{k-1},s_t)
\right].
}
\label{eq:da_ppo_forward_loss}
\end{equation}
{Here, the expectation uses stored local transitions, and the actions, mask, importance ratio, temperature, and advantage scale are held fixed during differentiation. The ratio corrects for sampling actions from the rollout policy; stopping it avoids an additional reverse-policy score term. Samples outside the active set contribute no direct forward-kernel gradient.}
The full actor loss is
\begin{equation}
\mathcal L_{\mathrm{actor}}^{\mathrm{PPO}}
=\mathcal L_{\mathrm{clip}}
{+\mathcal L_{\mathrm{fwd}}^{\mathrm{clip}}}.
\label{eq:da_ppo_complete_actor}
\end{equation}
{An alternative is the unmasked fresh-sample correction (Mode~1). At each stored augmented state, sample $\hat a_t^{k-1}\sim q_\theta(\cdot\mid a_t^k,s_t)$ without reparameterization and replace $\mathcal L_{\mathrm{fwd}}^{\mathrm{clip}}$ by}
\begin{equation}
{
\mathcal L_{\mathrm{fwd}}^{\mathrm{fresh}}(\theta)
=-\frac{\Tau}{\sigma_A}\,
\mathbb E\!\left[
\log\pi_\theta\!\left(a_t^k\mid
\operatorname{sg}(\hat a_t^{k-1}),s_t\right)
\right].
}
\label{eq:da_ppo_fresh_forward_loss}
\end{equation}
{This alternative needs neither an importance ratio nor a PPO mask. At the rollout parameters, $\rho_{t,k}=m_{t,k}=1$, so both variants recover the same population actor gradient, up to fixed advantage scaling. They differ away from the rollout policy: Mode~2 gates the forward correction with PPO clipping, whereas Mode~1 retains it on fresh samples. If the forward kernel is fixed, neither correction contributes a parameter gradient. Empirical clipping and KL control do not guarantee monotonic improvement with finite samples.} Setting $\Tau=0$ removes the path correction and yields a DPPO-style local PPO objective \citep{ren2025diffusion}.
\endgroup

\section{Extended Paper Sections}

\subsection{Related Work}
\label{app:related_work}

\subsubsection{Sampling from Unnormalized Target Distributions}
Sampling from unnormalized target distributions is a vibrant and active research area. Classical methods, such as Markov Chain Monte Carlo (MCMC) \citep{metropolis1953equation}, have long been foundational, but recent advances increasingly leverage neural networks to approximate target distributions \citep{BoltzmannGen, wu_solving_2018}. These approaches find broad applications, including molecule configuration prediction \citep{BoltzmannGen}, statistical physics \citep{wu_solving_2018}, Monte Carlo integration \citep{NeuralImportanceSampling}, and combinatorial optimization \citep{hibat-allah_variational_2021}.
Diffusion samplers offer an expressive solution by transporting noise samples toward high-density regions of the target distribution $\pi$ via a learned reverse diffusion process. These methods have been successfully applied in both continuous \citep{PathIntegralSampler, berner2022optimal, DDS, vargas2024transport, richter2023Imporved} and discrete domains \citep{sanokowski2024diffusion, Sanokowski2025scalable}, demonstrating robust performance on challenging sampling problems. Adjoint Matching \citep{domingo-enrich2025adjoint} also employs a score-matching-based loss similar to our \dawpo{} loss, though it is derived from the adjoint method and, in contrast to our method, does not allow for learnable diffusion coefficients.
Physics-informed neural network (PINN) \citep{raissi2019physics} losses, derived from the continuity equation of continuous normalizing flows, have been explored in \citep{tian2024liouville, PINN2023learning, fan2024path}.
Reinforcement learning (RL) naturally aligns with the unnormalized-sampling paradigm. A substantial body of work has established that RL can be formulated as probabilistic inference \citep{todorov2008general, ziebart2010modeling, kappen2012optimal, levine2018reinforcement, abdolmaleki2018maximum}. 

\subsection{Diffusion Models in Reinforcement Learning}
\label{subsec:diffusion_rl}

Diffusion models have recently gained traction in RL for representing complex action distributions. \citet{dong2025maximum} optimize a forward-KL objective, whereas \citet{ma2025reinforcement} consider both forward- and reverse-KL updates toward a policy-mirror-descent target. Their target-weighted diffusion training differs from our local maximum-entropy trajectory construction.
DIME \citep{celik2025dime} integrates diffusion models with Soft Actor-Critic (SAC), but unlike our approach, it does not evaluate a Q-function at each diffusion step. Instead, its objective is based on applying the Data Processing Inequality directly to the surrogate loss in Eq.~\ref{eq:surroga_loss}. This method requires backpropagation through the entire diffusion chain for every environment action, which becomes memory-intensive for large numbers of diffusion steps {$K$}.
The work most closely related to ours is \citet{Sanokowski2025scalable}, who use RL to train diffusion models scalably for unnormalized sampling. We extend their local policy-gradient construction to RL itself: the diffusion model becomes the policy acting in an environment, and its local transition rewards are derived from the ME-RL trajectory objective. 
At $\Tau=0$, \dappo{} yields a DPPO-style local PPO objective \citep{ren2025diffusion}; the original DPPO algorithm additionally specifies denoising discounting and fine-tuning choices. Our formulation adds the maximum-entropy path correction and supports learned forward kernels.
Recent works have also explored flow models in RL \citep{liu2025flow, park2025flow, zhang2025reinflow, mcallister2025flow}, further expanding the intersection of generative models and RL.

\subsection{Additional Experimental Details}
\subsubsection{Continuous-Control Experiments}
\label{app:continuous_control_protocol}

\begin{figure*}[htbp]
  \centering
  \includegraphics[width=.98\textwidth]{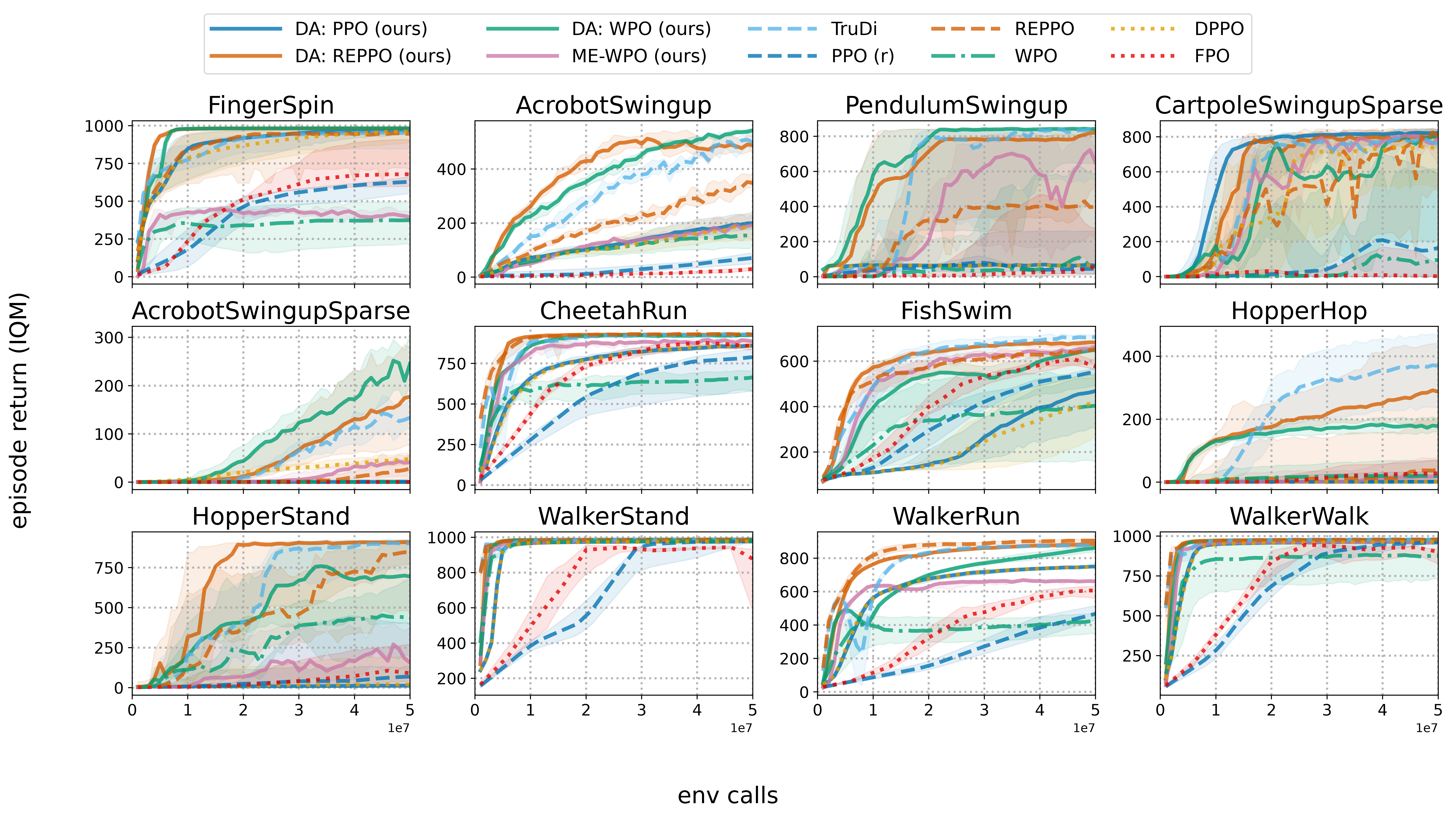}
  \vspace{-0.6em}
  \caption{Continuous-control performance across twelve tasks. Curves show IQM returns; shaded regions indicate $95\%$ bootstrap confidence intervals.}
  \label{fig:main_results}
\end{figure*}

We compare the three DA-MDP methods with PPO, REPPO, WPO, ME-WPO, TruDi, DPPO, and FPO on twelve DeepMind Control tasks in MuJoCo Playground \citep{zakkamujoco}. Our experiments use one configuration per method across tasks and $50$M environment interactions. We aggregate eight independent training seeds per method--task pair. Diffusion policies use $K=8$ steps; DPPO is implemented as \dappo{} with $\Tau=0$. PPO~(r) denotes the externally reported results of \citet{voelcker2025reppo}; all other curves are from our experiments. The NaNs originated \emph{within the environment simulations themselves}, caused by bugs in the Humanoid Run, Walk, and Stand environments. These simulation failures occurred more frequently for diffusion-based variants. After four consecutive NaN failures despite restarting with new random seeds, we excluded these environments from the experiments. Common settings are given in App.~\ref{app:common_hyperparameters}, with exact experiment configurations provided alongside the code.

{The twelve-task comparison evaluates diffusion policies using ODE sampling (Eq.~\ref{eq:vp_evaluation_modes}).} We report the interquartile mean (IQM) \citep{agarwal2021deep}, both per task (Fig.~\ref{fig:main_results}) and across tasks (Fig.~\ref{fig:all_envs}). For aggregate results, we divide returns by $1000$ and compute an equally task-weighted IQM over seed-level scores. We obtain $95\%$ confidence intervals from $10{,}000$ bootstrap replicates, resampling seeds independently within each task.

All three DA-MDP variants improve on their corresponding base algorithms in aggregate. \dareppo{} learns fastest overall and reaches final performance comparable to TruDi. \dawpo{} substantially improves on WPO and ME-WPO, while \dappo{} outperforms PPO and DPPO. The task-level results show particularly large gains for the Q-based DA methods on PendulumSwingup and AcrobotSwingupSparse, although TruDi remains stronger on HopperHop. These results show that local denoising updates can retain the performance benefits of diffusion policies across diverse control tasks.

\subsubsection{Action-Transform and Entropy-Measure Ablation}
\label{app:action_transform_entropy_ablation}

Continuous-control environments require bounded actions, whereas the Gaussian reverse process used by \dareppo{} produces a terminal latent $x^0\in\mathbb R^d$. The usual solution is to execute $a=\tanh(x^0)$. {The Gaussian ablation compares two entropy conventions throughout training. The latent-entropy formulation omits the terminal $\tanh$ log-Jacobian from the actor objective, critic soft rewards, and temperature statistic. The transformed-action entropy formulation includes this correction in all three components. Both formulations execute the same action transform.} We ablate these two choices and compare them with a diffusion process that is bounded by construction.

\paragraph{Latent-space and transformed-space entropy.}
\begingroup
We distinguish two entropy measures: \mbox{\textbf{Tanh latent}}, which constrains entropy in latent coordinates $x^0\in\mathbb R^d$, and \mbox{\textbf{Tanh $+J$}}, which constrains entropy in transformed-action coordinates $a=\tanh(x^0)$. In the latent formulation, the environment applies the action transform and entropy is measured before squashing. In the transformed-action formulation, the transform belongs to the policy and its log-Jacobian contributes to the actor objective, critic soft rewards, and temperature statistic. The latent-action environment is defined by
\[
R_X(s,x^0)=R_{\mathrm{env}}(s,\tanh x^0),
\qquad
P_X(s'\mid s,x^0)=P_{\mathrm{env}}(s'\mid s,\tanh x^0).
\]
The latent density $q_X$ and transformed density $q_A$ are related by change of variables:
\[
\begin{aligned}
\log q_A(a\mid s)&=\log q_X(x^0\mid s)-j_{\tanh}(x^0),\\
\mathcal H[q_A(\cdot\mid s)]&=\mathcal H[q_X(\cdot\mid s)]
+\mathbb E_{q_X(\cdot\mid s)}[j_{\tanh}(x^0)],\\
j_{\tanh}(x^0)&:=\sum_{i=1}^{d}\log\!\left(1-\tanh^2(x_i^0)\right).
\end{aligned}
\]
The same correction relates the corresponding path-space entropy bounds.

For a minibatch of $B$ stored augmented states $\tilde s_b=(s_b,x_b^{k_b},k_b)$ with uniformly sampled phases $k_b\in\{1,\ldots,K\}$, draw fresh local actions $\hat x_b^{k_b-1}\sim q_\theta(\cdot\mid x_b^{k_b},s_b)$. The sampled entropy bounds are
\begin{equation}
\begin{aligned}
\widetilde{\mathcal H}_X
&=\mathcal H[q_{\mathrm{prior}}]
+\frac{K}{B}\sum_{b=1}^{B}
\log\frac{\pi_\theta(x_b^{k_b}\mid\hat x_b^{k_b-1},s_b)}
{q_\theta(\hat x_b^{k_b-1}\mid x_b^{k_b},s_b)},\\
\widetilde{\mathcal H}_A
&=\widetilde{\mathcal H}_X+\widehat j_{\tanh},
\qquad
\widehat j_{\tanh}:=\frac{1}{B}\sum_{b=1}^{B}j_{\tanh}(\widetilde x_b^0).
\end{aligned}
\label{eq:latent_transformed_entropy_estimates}
\end{equation}
The factor $K$ accounts for sampling diffusion phases. For the Jacobian estimate, each row supplies the stored terminal-phase pair $(s_b,x_b^1)$ from the same rollout chain, and we draw $\widetilde x_b^0\sim q_\theta(\cdot\mid x_b^1,s_b)$ independently of the selected phase $k_b$. Thus, the terminal correction is estimated for every row and has no additional factor $K$. Reusing rollout conditioning states gives the approximation described in App.~\ref{app:temp_tuning}.

Let $\mathcal J_{\mathrm{env}}(\theta)$ denote expected environment return, with the same state/time weighting understood for the entropy terms. The two sampled Lagrangians are
\begin{equation}
\begin{aligned}
\mathcal L_X(\theta,\Tau)
&=\mathcal J_{\mathrm{env}}(\theta)
+\Tau\bigl(\widetilde{\mathcal H}_X-\mathcal H_{\mathrm{tar}}^X\bigr),\\
\mathcal L_A(\theta,\Tau)
&=\mathcal J_{\mathrm{env}}(\theta)
+\Tau\bigl(\widetilde{\mathcal H}_A-\mathcal H_{\mathrm{tar}}^A\bigr).
\end{aligned}
\label{eq:latent_transformed_lagrangians}
\end{equation}
Actors maximize the corresponding Lagrangian: the latent formulation omits the Jacobian, whereas the transformed formulation includes it. At any diffusion phase $k$, their local actor losses, before trust-region modifications, have the form
\begin{equation}
\begin{aligned}
\ell_k^m(\theta;\Tau)
&=\mathbb E_{x^{k-1}\sim q_\theta(\cdot\mid x^k,s)}
\Bigg[-\bar Q_{\mathrm{DA}}^m(\tilde s_k,x^{k-1})
+\Tau\log\frac{q_\theta(x^{k-1}\mid x^k,s)}
{\pi_\theta(x^k\mid x^{k-1},s)}\\
&\hspace{4em}
-\Tau\,\mathbf 1_{\{m=A\}}\mathbf 1_{\{k=1\}}j_{\tanh}(x^{k-1})
\Bigg],
\qquad m\in\{X,A\}.
\end{aligned}
\label{eq:entropy_local_actor_loss}
\end{equation}
Here $\tilde s_k=(s,x^k,k)$, and $\bar Q_{\mathrm{DA}}^m$ is the augmented critic for formulation $m$, excluding the current local entropy terms and including future soft rewards (App.~\ref{app:q_function_training}). Its parameters are held fixed during actor differentiation. Every phase uses its own local decision $x^{k-1}$. At $k=1$, both formulations execute the environment action $\tanh(x^0)$, but only the transformed-action formulation ($m=A$) includes the explicit Jacobian term. The critic's future soft rewards use the same entropy convention.

For either formulation, temperature tuning minimizes
\begin{equation}
\mathcal L_\Tau^m(\xi_\Tau)
=\exp(\xi_\Tau)\,\operatorname{sg}\!\left(
\widetilde{\mathcal H}_m-\mathcal H_{\mathrm{tar}}^m\right),
\qquad \Tau=\exp(\xi_\Tau),\quad m\in\{X,A\}.
\label{eq:latent_transformed_temperature_losses}
\end{equation}
The entropy statistic, including the Jacobian sample for $m=A$, is detached in this loss. Gradient descent therefore increases the temperature when entropy is below its target and decreases it when entropy is above target.

The latent-space return can be optimized directly under the cost and gradient integrability conditions in App.~\ref{app:infinite_horizon_da_mdp}. To additionally recover the finite-horizon normalized-target KL interpretation of Secs.~\ref{sec:maxentRL} and~\ref{sec:da_mdp_derivation}, we assume, for each fixed $\Tau>0$,
\[
M_\Tau:=\sup_s\int_{\mathbb R^d}
\exp\!\left(R_X(s,x)/\Tau\right)\,dx<\infty.
\]
A sufficient condition is $R_X(s,x)\le C-c\|x\|^2$, uniformly in $s$, with $c>0$. Normalized forward kernels preserve the finite partition function. For geometric stopping, the sufficient condition in Eq.~\ref{eq:geometric_normalizability_condition} becomes $\gamma M_\Tau<1$, which a constant reward shift can enforce as in Eq.~\ref{eq:geometric_reward_shift}. This additional normalization assumption concerns the KL interpretation and is not required to define the return itself.
\endgroup

\paragraph{Variance-preserving Gaussian diffusion.}
\label{app:vp_sde}
For the Gaussian variants, let $x^k\in\mathbb R^d$ denote the latent at diffusion step $k$ and let $q_{\mathrm{prior}}=\mathcal N(\mathbf 0,\nu^2I_d)$. The forward process is the variance-preserving SDE
\begin{equation}
d x=-\frac12\boldsymbol\beta(t)x\,dt
+\nu\boldsymbol\beta(t)^{1/2}dW_t,
\label{eq:vp_forward_sde}
\end{equation}
whose stationary distribution is $\mathcal N(\mathbf 0,\nu^2I_d)$. To parameterize the reverse process, we introduce a learned control $u_\Psi(s,x^k,k)\in\mathbb R^d$, predicted by a neural network with parameters $\Psi$ from the environment state $s$, noisy latent $x^k$, and diffusion step $k$. This control provides the learned correction to the reverse drift. We discretize each coordinate independently. Writing $\boldsymbol\delta_k=\operatorname{diag}(\delta_{k,1},\ldots,\delta_{k,d})$, with $(\delta_{k,1},\ldots,\delta_{k,d})^\top=\boldsymbol\beta_k\Delta_k$, gives the paired transition kernels
\begin{equation}
\begin{aligned}
\pi_\theta(x^k\mid x^{k-1},s)
&=\mathcal N\!\left(
(I-\tfrac12\boldsymbol\delta_{k-1})x^{k-1},
\nu^2\boldsymbol\delta_{k-1}\right),\\
q_\theta(x^{k-1}\mid x^k,s)
&=\mathcal N\!\left(
(I+\tfrac12\boldsymbol\delta_k)x^k
+\nu^2\boldsymbol\delta_k u_\Psi(s,x^k,k),
\nu^2\boldsymbol\delta_k\right).
\end{aligned}
\label{eq:vp_discrete_kernels}
\end{equation}
The complete parameter vector is $\theta=(\Psi,\boldsymbol\beta_K,\ldots,\boldsymbol\beta_0)$. At reverse step $k$, $q_\theta$ uses $\boldsymbol\beta_k$ while its paired forward kernel uses $\boldsymbol\beta_{k-1}$. The pair therefore shares no learnable diffusion coefficient, although the global coefficient sequence is shared across neighboring transition pairs. App.~\ref{app:da_monotonic_improvement} discusses the resulting forward-kernel update and trust-region term. Starting from $x^K\sim q_{\mathrm{prior}}$, the policy applies the reverse kernels down to $x^0$ and executes $a=\tanh(x^0)$ in the bounded environment. The evaluation rules below distinguish stochastic transitions, ODE integration, and conditional-mean inference.

\paragraph{SDE, ODE, and deterministic evaluation.}
\label{app:diffusion_evaluation_modes}
All three evaluation modes start from a sampled prior latent $x^K\sim q_{\mathrm{prior}}$. With independent $\epsilon_k\sim\mathcal N(0,I_d)$, their updates are
\begin{equation}
\begin{aligned}
\mathrm{SDE}:\quad x^{k-1}
&=(I+\tfrac12\boldsymbol\delta_k)x^k
+\nu^2\boldsymbol\delta_k u_\Psi(s,x^k,k)
+\nu\boldsymbol\delta_k^{1/2}\epsilon_k,\\
\mathrm{ODE}:\quad x^{k-1}
&=(I+\tfrac12\boldsymbol\delta_k)x^k
+\tfrac12\nu^2\boldsymbol\delta_k u_\Psi(s,x^k,k),\\
\mathrm{Deterministic}:\quad x^{k-1}
&=(I+\tfrac12\boldsymbol\delta_k)x^k
+\nu^2\boldsymbol\delta_k u_\Psi(s,x^k,k).
\end{aligned}
\label{eq:vp_evaluation_modes}
\end{equation}
With an exact score in continuous time, the probability-flow ODE samples from the same marginal distributions as the reverse SDE \citep{SongDiff}. Deterministic sampling instead selects the mean, which is also the mode, of each Gaussian reverse kernel \citep[Sec.~4.4]{le2026trust}. Following these locally most probable transitions reduces exploration and can improve evaluation return, analogous to executing the mean of a Gaussian policy. The sampled prior still permits diverse actions at a fixed state. The G1 ablation additionally fixes the prior to its mean (App.~\ref{app:diffusion_step_details}).

\paragraph{Learning Gaussian diffusion coefficients.}
\label{app:vp_coefficient_learning}
We ensure positive diffusion coefficients by setting $\beta_{k,i}=\operatorname{softplus}(b_{k,i})$ and $\delta_{k,i}=\Delta_k\beta_{k,i}$, with fixed integration increments $\Delta_k$. Starting from the prescribed initial coefficients, we optimize $b_{k,i}$ jointly with the control network, including gradients through both reverse and forward kernels (App.~\ref{app:learned_forward_kernel_correction}).

In the Gaussian ablation, we fix the cosine integration schedule and instead learn positive friction $f_i$ per action coordinate, shared across diffusion steps. This adapts the effective diffusion coefficients through $\delta_{k,i}=2\Delta_k/(\nu^2 f_i)$.

\begin{figure}[!htbp]
    \centering
    \includegraphics[width=\textwidth]{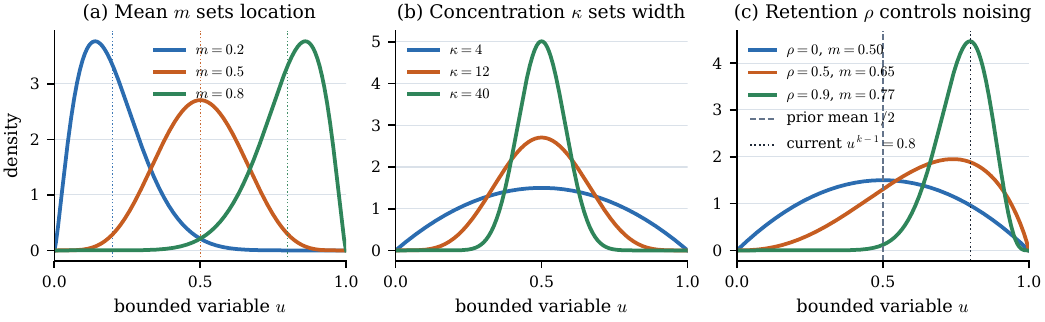}
    \caption{Product-Beta forward transitions in one dimension. (a) Varying $m$ shifts the mean at fixed concentration. (b) Increasing $\kappa$ narrows the distribution at fixed mean. (c) With $u^{k-1}=0.8$, $\rho$ controls the influence of the current value on the next latent. At $\rho=0$, the transition follows the $\operatorname{Beta}(2,2)$ reference distribution; increasing $\rho$ moves the conditional mean toward $u^{k-1}$ and reduces the variance.}
    \label{fig:product_beta_parameterization}
\end{figure}

\paragraph{Bounded Product-Beta diffusion.}
\label{app:product_beta}
To keep the entire diffusion process within the action bounds, we use Product-Beta transitions inspired by Dirichlet diffusion with Jacobi processes \citep{avdeyev2023dirichlet}. Let $u^k\in[0,1]^d$ and map the terminal sample to the environment action as $a=2u^0-1$. To describe the transition kernels, recall that a Beta distribution with parameters $(\alpha,\beta)$ can equivalently be parameterized by its mean $m$ and total concentration $\kappa$:
\[
m=\frac{\alpha}{\alpha+\beta},
\qquad
\kappa=\alpha+\beta,
\qquad
\alpha=m\kappa,
\quad
\beta=(1-m)\kappa.
\]
This reparameterization introduces no additional degrees of freedom. It merely separates the two roles of $(\alpha,\beta)$: $m$ locates the density, whereas $\kappa$ controls its width, as illustrated in Fig.~\ref{fig:product_beta_parameterization}(a--b).

We use the factorized reference distribution $\operatorname{Beta}(2,2)^{\otimes d}$, whose mean is $1/2$ and whose total concentration is $4$. At diffusion step $k$, the forward transition is defined by
\[
\begin{aligned}
\pi_k(u^k\mid u^{k-1})
&=\prod_{i=1}^{d}\operatorname{Beta}
\!\left(u_i^k;m_{k,i}\kappa_k,(1-m_{k,i})\kappa_k\right),\\
m_{k,i}&=(1-\rho_k)\frac{1}{2}+\rho_k u_i^{k-1},
&\kappa_k&=\frac{4}{1-\rho_k^2}.
\end{aligned}
\]
The coefficient $\rho_k\in[0,1)$ controls how strongly each diffusion step depends on its input. If $\rho_k=0$, then $m_{k,i}=1/2$ and $\kappa_k=4$, so the transition is exactly $\operatorname{Beta}(2,2)$ and forgets its input. As $\rho_k$ approaches one, $m_{k,i}$ approaches $u_i^{k-1}$ and $\kappa_k$ increases, yielding an increasingly narrow transition around the input; Fig.~\ref{fig:product_beta_parameterization}(c) visualizes this effect. The particular relation $\kappa_k=4/(1-\rho_k^2)$ makes the first two moments of the reference distribution stationary under every forward step.

It remains to choose how quickly information is removed over the complete chain. {We initialize the schedule by defining the cumulative signal $c_k^{(0)}$ and setting $\rho_k^{(0)}=c_k^{(0)}/c_{k-1}^{(0)}$:}
\[
{c_k^{(0)}=\cos\!\left(\frac{\pi k}{2K}\right),
\qquad
\rho_k^{(0)}=\frac{c_k^{(0)}}{c_{k-1}^{(0)}},}
\qquad k=1,\ldots,K.
\]
 {At initialization, $\prod_{j=1}^k\rho_j^{(0)}=c_k^{(0)}$, and therefore}
\[
\mathbb E\!\left[u_i^k-\frac12\,\middle|\,u_i^0\right]
={c_k^{(0)}}\left(u_i^0-\frac12\right).
\]
 {The initialization therefore interpolates from $c_0^{(0)}=1$ to $c_K^{(0)}=0$. We use $K=8$, keep $\rho_K=0$ so that the final forward step reaches the $\operatorname{Beta}(2,2)$ reference distribution exactly, and learn $\rho_1,\ldots,\rho_{K-1}$ jointly with the reverse policy. The cosine construction is thus an initialization rather than a fixed schedule.}

The reverse policy is likewise factorized,
\[
q_\theta(u^{k-1}\mid u^k,s)
=\prod_{i=1}^{d}\operatorname{Beta}
\!\left(u_i^{k-1};\alpha_{\theta,k,i},\beta_{\theta,k,i}\right),
\]
The score network predicts corrections to the forward-kernel mean and log concentration. This construction is a discrete-time analogue of a Jacobi diffusion whose complete path remains inside the action domain. The affine map from $u^0$ to $a$ contributes the constant log-Jacobian $d\log2$; unlike the Gaussian alternatives, it requires no sample-dependent terminal correction. Both the transition entropy and the KL between old and updated reverse kernels are available in closed form.

\paragraph{Product-Beta trust region.}
For Product-Beta kernels, we likewise constrain the sum of rollout-to-current KLs over diffusion steps. At a fixed augmented state, suppress $k$ and write $\kappa_{\theta,i}=\alpha_{\theta,i}+\beta_{\theta,i}$. By independence, the local KL is the sum of the closed-form Beta divergences \citep[Table~3]{gil2013renyi}:
\begin{equation}
\begin{aligned}
d_\theta(\tilde s)
&=\sum_{i=1}^{d}\Bigg[
\log\frac{B(\alpha_{\theta,i},\beta_{\theta,i})}{B(\alpha_{\thetar,i},\beta_{\thetar,i})}
+(\alpha_{\thetar,i}-\alpha_{\theta,i})\psi(\alpha_{\thetar,i})\\
&\hspace{4em}
+(\beta_{\thetar,i}-\beta_{\theta,i})\psi(\beta_{\thetar,i})
+(\kappa_{\theta,i}-\kappa_{\thetar,i})\psi(\kappa_{\thetar,i})
\Bigg],
\end{aligned}
\label{eq:beta_reverse_kernel_kl}
\end{equation}
where $B$ is the Beta function and $\psi$ the digamma function. We sum these local KLs over diffusion steps, or use $K d_\theta$ when sampling a phase uniformly (Eq.~\ref{eq:da_reppo_chain_kl}); no action sampling is needed to evaluate each local KL.

{Learning $\rho_{1:K-1}$ also changes the forward kernels that define the DA objective. We therefore constrain the forward chain divergence from the rollout kernels to the updated kernels, $\sum_{k=1}^{K}D_{\mathrm{KL}}(\pi_{\thetar,k}\,\|\,\pi_{\theta,k})$, to $0.08$ and optimize its Lagrange multiplier jointly with the policy. This forward trust region complements the reverse-policy chain-KL constraint above.}

\paragraph{Product-Beta evaluation.}
The discrete Beta kernels support stochastic sampling and conditional-mean evaluation:
\begin{equation}
\begin{aligned}
\mathrm{Stochastic}:\quad
&u^K\sim\operatorname{Beta}(2,2)^{\otimes d},
&u_i^{k-1}&\sim\operatorname{Beta}(\alpha_{\theta,k,i},\beta_{\theta,k,i}),\\
\mathrm{Deterministic}:\quad
&u^K\sim\operatorname{Beta}(2,2)^{\otimes d},
&u_i^{k-1}&=\frac{\alpha_{\theta,k,i}}{\alpha_{\theta,k,i}+\beta_{\theta,k,i}}.
\end{aligned}
\label{eq:beta_evaluation_modes}
\end{equation}
Both modes sample the Beta prior; deterministic sampling then replaces each reverse transition by its conditional mean. The reported Beta ablation uses the stochastic rule.

\paragraph{Learning Product-Beta kernel parameters.}
\label{app:beta_coefficient_learning}
The reverse network learns both the location and noise level of each denoising transition. Let $h_{\Psi,k,i}(s,u^k)$ and $g_{\Psi,k,i}(s,u^k)$ denote its mean and log-concentration corrections, and define $\bar m_{k,i}=(1-\rho_k)/2+\rho_k u_i^k$ and $\omega_k=\sqrt{1-\rho_k^2}$. Before numerical stabilization, the reverse parameters are
\[
\begin{aligned}
\widehat m_{\theta,k,i}
&=\operatorname{sigmoid}\!\left(\operatorname{logit}(\bar m_{k,i})+\omega_k h_{\Psi,k,i}(s,u^k)\right),\\
\widehat\kappa_{\theta,k,i}
&=\kappa_k\exp\!\left(\operatorname{clip}\!\left(\omega_k g_{\Psi,k,i}(s,u^k),-5,5\right)\right),\\
\alpha_{\theta,k,i}
&=\widehat m_{\theta,k,i}\widehat\kappa_{\theta,k,i},
\qquad
\beta_{\theta,k,i}=(1-\widehat m_{\theta,k,i})\widehat\kappa_{\theta,k,i}.
\end{aligned}
\]
The sigmoid and exponential enforce a valid mean and positive concentration. Actor updates train both network outputs. To learn the forward schedule, we parameterize $\rho_k=\varepsilon+(1-2\varepsilon)\operatorname{sigmoid}(r_k)$ for $k<K$, with $\varepsilon=10^{-6}$, and fix $\rho_K=0$ to preserve the exact prior endpoint. We initialize the unconstrained parameters $r_k$ to reproduce the cosine schedule and optimize them jointly with the reverse network. The concentrations follow $\kappa_k=4/(1-\rho_k^2)$, and gradients account for the coefficients' effects on both forward and reverse kernels.

\paragraph{Experimental setup.}
We compare Gaussian diffusion under latent-entropy and transformed-action entropy formulations with Product-Beta diffusion. For each family, we test three target-entropy factors in $\mathcal H_{\mathrm{tar}}(c)=\mathcal H[q_{\mathrm{prior}}]+cd$, where $d$ is the action dimension. The latent-space Gaussian runs use $c\in\{-4,-5,-6\}$, the transformed-space Gaussian runs use $c\in\{-5,-6,-7\}$, and the Product-Beta runs use $c\in\{-4,-5,-6\}$. Different ranges are necessary because the two Gaussian estimators differ by the negative $\tanh$ Jacobian and because the Beta prior has a different entropy. The Gaussian prior has standard deviation $2.5$. {Both Gaussian variants set the environment-action clipping bound to $1.0$, with no additional interior clamp.} {The Product-Beta runs use a $\operatorname{Beta}(2,2)$ prior in each coordinate, initialize the forward coefficients from the cosine schedule, and learn them under a forward chain-KL bound of $0.08$.}

\begin{table}[H]
    \centering
    \small
    \setlength{\tabcolsep}{5pt}
    \caption{Policy families in the action-transform ablation. The Gaussian variants differ in the entropy terms used by the actor, critic, and temperature estimator. The transformed-action formulation includes the terminal $\tanh$ log-Jacobian in all three components, whereas the latent formulation omits it.}
    \label{tab:action_transform_entropy_settings}
    \begin{tabularx}{\textwidth}{lXXXX}
        \toprule
        Policy family & Executed action & Terminal log-Jacobian in actor/critic & Entropy used for temperature tuning & Target factors $c$ \\
        \midrule
        $\tanh$, latent entropy
        & $a=\tanh(x^0)$ & Omitted
        & $\widehat{\mathcal H}_{X}$
        & $-4,-5,-6$ \\
        $\tanh$, transformed entropy
        & $a=\tanh(x^0)$ & Included at $k=1$
        & $\widehat{\mathcal H}_{X}+\widehat{\mathbb E}[j_{\tanh}(x^0)]$
        & $-5,-6,-7$ \\
        Product-Beta / Jacobi
        & $a=2u^0-1$ & Constant $d\log 2$
        & Bounded path-space entropy
        & $-4,-5,-6$ \\
        \bottomrule
    \end{tabularx}
\end{table}

For every policy family and target factor, we train four independent seeds on PendulumSwingup, AcrobotSwingup, HopperHop, FishSwim, and AcrobotSwingupSparse. This gives nine settings per environment and $180$ runs in total. All settings use \dareppo{}, eight diffusion steps, $50$ million environment interactions, no action chunking, and a chain-KL bound of $0.08$. {For Product-Beta, the additional forward-kernel chain-KL bound is also $0.08$.} The score network has three hidden layers of width $256$, and the critic has two hidden layers of width $512$. All remaining optimization settings are shared.

We evaluate each policy every $10\%$ of training using $128$ independent evaluation episodes. Both Gaussian variants use sampled priors and ODE transitions (Eq.~\ref{eq:vp_evaluation_modes}); Product-Beta uses the stochastic rule in Eq.~\ref{eq:beta_evaluation_modes}. We report mean returns over four training seeds with $95\%$ bootstrap confidence intervals.

\paragraph{Results.}
\begin{figure}[!ht]
    \centering
    \includegraphics[width=\textwidth]{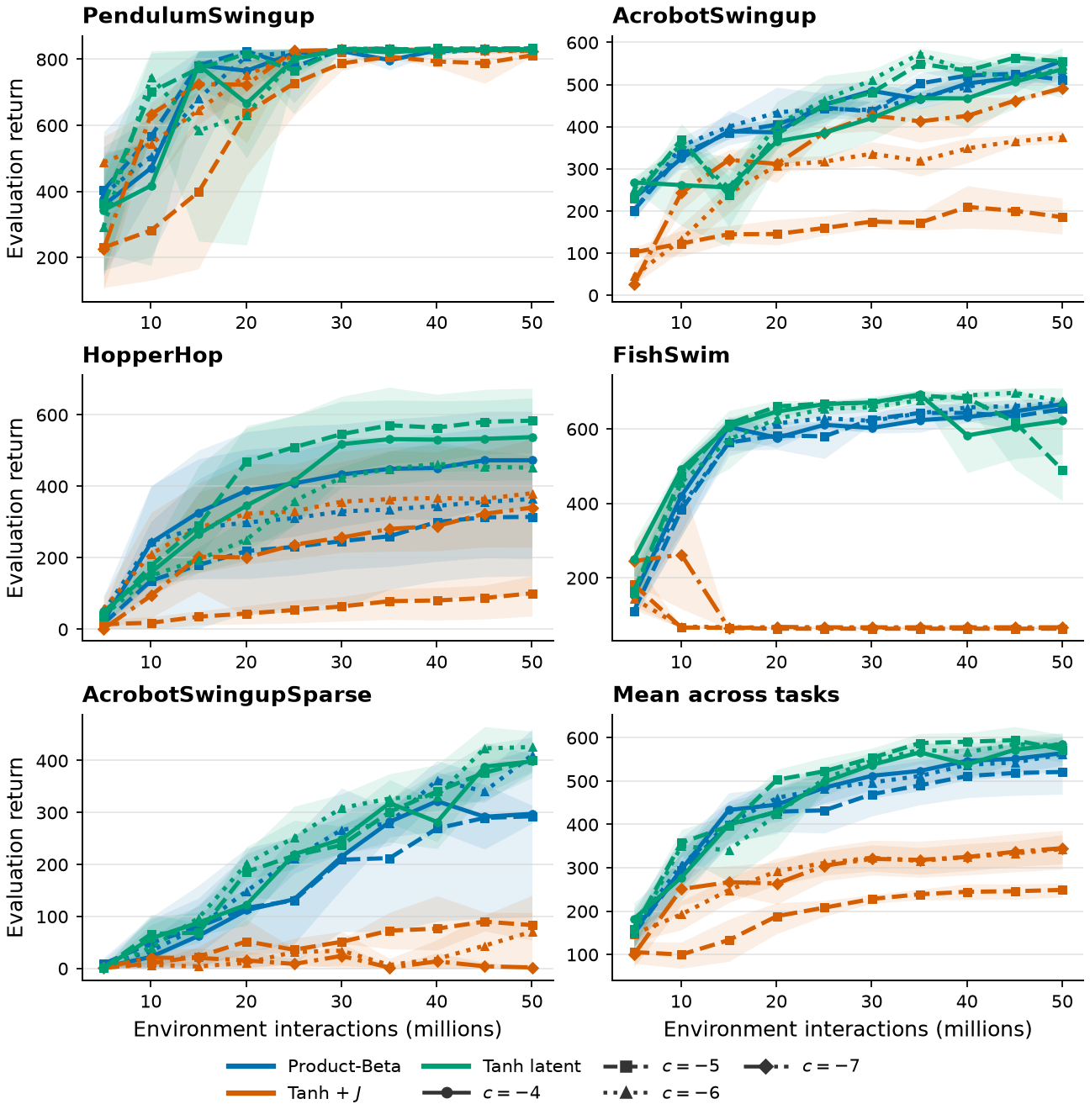}
    \vspace{-0.5em}
    \caption{Action-transform and entropy-measure ablation. Curves show mean evaluation return over four training seeds, with $95\%$ bootstrap confidence intervals. The aggregate weights the five tasks equally and bootstraps seeds within each task. Gaussian policies use ODE sampling; {Product-Beta uses stochastic reverse transitions and a learned forward schedule initialized from the cosine coefficients and constrained by a forward chain-KL bound of $0.08$.}}
    \label{fig:action_transform_entropy_results}
\end{figure}

{Figure~\ref{fig:action_transform_entropy_results} compares learned Product-Beta diffusion with Gaussian diffusion under latent-entropy and transformed-action entropy formulations. After $50$M interactions, the learned Product-Beta policy reaches aggregate returns of $564.2$ for $c=-4$ and $561.4$ for $c=-6$, compared with $584.9$ and $583.7$ for the corresponding latent-entropy Gaussian settings. PendulumSwingup is largely saturated, while HopperHop remains variable across seeds. Over the tested target-entropy ranges, the transformed-action entropy formulation is substantially less robust than the latent-entropy formulation, particularly on FishSwim and AcrobotSwingupSparse. This comparison changes the actor, critic, and temperature update together and does not isolate the effect of temperature tuning. Overall, learned Product-Beta diffusion achieves aggregate performance close to the latent-entropy Gaussian formulation without requiring a $\tanh$ transform.}

\clearpage
\begingroup

\subsubsection{StackCube: Task, Training, and Matched-State Evaluation}
\label{app:stackcube_multimodality}
\small
\setlength{\intextsep}{4pt}
\setlength{\parskip}{3pt}
\captionsetup{font=small,skip=3pt}

We compare final checkpoints after extended, method-specific training budgets, with success rates approaching a plateau for most runs. Gaussian REPPO reaches its success plateau substantially earlier than the DA variants, as shown in Fig.~\ref{fig:manipulation_training_success}.

\paragraph{Task and randomized training states.}
We adapt ManiSkill3 StackCube \citep{tao2025maniskill} to the symmetric task of \citet{le2026trust}. A Panda controls seven joint increments and its gripper to stack two $4$\,cm cubes. Centers are sampled over $x\in[-0.2,0.2]$, $y\in[-0.4,0.4]$\,m, rejecting overlaps, with randomized yaws and robot joint-reset noise of $0.02$\,rad standard deviation. Either order is valid; no target order is sampled or observed. Training resets on success or at $50$ physical steps, stopping traces but bootstrapping both boundaries from the final observation.

\paragraph{Observations and permutation invariance.}
Training and evaluation share $85$-dimensional states: robot state, both cube poses, tool-to-cube and directed relative/stack-goal vectors, headings, velocities, grasp flags and symmetric pair features. Permutation-invariant actor and critic encoders compute $z(s)=[r,c,\tfrac12\{\phi([r,o_A])+\phi([r,o_B])\}]$, where $r\in\mathbb R^{25}$ is robot state, $c\in\mathbb R^6$ is pair center/absolute displacement, and $o_i\in\mathbb R^{27}$ collects object features. The shared MLP $\phi$ is $52\to64\to54$ (LayerNorm, GeLU); normalization is tied across object slots. Label swapping leaves $z$ invariant, without enforcing equally likely spatial strategies.

\paragraph{Reward and strict success.}
Let $p_i$ denote cube centers, $p_{\mathrm{TCP}}$ the tool position, $g_i=p_j+(0,0,0.04)$ the target above the other cube, and $f(d)=1-\tanh(5d)$. The implemented raw dense reward is
\[
r_{\rm raw}=
\begin{cases}
8,&\text{strict success},\\
6+(u+s)/2,&\text{cubes geometrically stacked},\\
4+\max_i f(\|g_i-p_i\|),&\text{either cube grasped},\\
2f(\min_i\|p_{\mathrm{TCP}}-p_i\|),&\text{otherwise}.
\end{cases}
\]
Cases have the displayed priority. Training uses $r_{\rm raw}/8$; $u$ is gripper opening (one if the upper cube is released), and $s=1-\tanh(10\|v_{\rm top}\|+\|\omega_{\rm top}\|)$. Geometric stacking requires horizontal distance $\leq3.33$\,cm and height difference $4\pm0.5$\,cm. Strict success also requires release, upper-cube linear speed $<1$\,cm/s and angular speed $<0.5$\,rad/s: lifting alone is insufficient.

\paragraph{Training configurations.}
{Each method uses three training seeds and the final checkpoint of each run.} All use $1024$ environments, randomized initial episode counters, $\gamma=0.8$, $\lambda=0.98$, Adam, gradient clipping $0.5$, and $301$-bin HL-Gauss critics on $[-10,10]$. DA actors have three residual hidden layers (LayerNorm/GeLU), critics two; $K=8$, cosine time grid, prior standard deviation $2.5$. Gaussian REPPO has a three-layer state-dependent tanh-Gaussian actor (SiLU/RMSNorm) and a $512$-wide critic encoder/head. DA critic auxiliary weight is $0.05$; Gaussian REPPO disables it and uses exact forward policy KL.
\begin{center}
\footnotesize
\setlength{\tabcolsep}{3pt}
\begin{tabular}{@{}lccccc@{}}
\toprule
 & Gaussian & DA:REPPO & DA:REPPO & DA:PPO & DA:WPO\\
\midrule
Chunk size $C$ & 1 & 1 & 4 & 1 & 1\\
{Interactions/seed (M)} & {50.07} & {100.14} & {250.09} & {300.15} & {250.09}\\
Hidden width & 512 & 512 & 768 & 512 & 512\\
Policy steps/rollout & 128 & 128 & 32 & 128 & 128\\
Epochs/minibatches & 8/64 & 8/64 & 16/16 & 8/64 & 8/64\\
Network LR ($10^{-4}$) & 3 & 3 & 3 & 3 & 6\\
KL bound / PPO clip & .08 & .08 & .16 & .2 (clip) & .04\\
Entropy factor & $-.5$ & $-4$ & $-2$ & $-3$ & $-10$\\
\bottomrule
\end{tabular}
\end{center}
\noindent DA entropy targets are $8C[\frac12\log(2\pi e\,\sigma_{\rm ref}^2)+c]$, with $\sigma_{\rm ref}=2.5$ for PPO and $3$ otherwise; Gaussian REPPO targets $-0.5$ nats/action coordinate. DA:REPPO dual LRs are $2.12\cdot10^{-4}$, PPO temperature LR $3\cdot10^{-4}$, WPO dual LRs $6\cdot10^{-4}$. DA:REPPO/WPO use local forward-score/current-$Q$ guidance.
\label{app:stackcube_chunking}
For $C=4$, KL scales as $0.08\sqrt C$, entropy factor as $-4/\sqrt C$; temperature is learned. Chunk rewards, $\gamma^m$ and $\lambda^m$ use actual physical substeps $m$, with deferred resets.

\paragraph{Action-chunk inference timing.}
\label{app:stackcube_chunking_timing}
{We benchmark the unchunked ($C=1$, width $512$) and chunked ($C=4$, width $768$) \dareppo{} architectures on a single NVIDIA A100-SXM4-40GB with $K=8$. Using freshly initialized policies, we time single-state ODE inference in five blocks of $20$ CUDA-synchronized calls after $10$ warmup calls. Mean call latency is $67.07\pm0.18$\,ms for $C=1$ and $68.03\pm0.16$\,ms for $C=4$ (standard deviations across blocks). Dividing by the number of predicted actions gives $67.07$ versus $17.01$\,ms per action, a $3.94\times$ amortized inference speedup when all four actions are executed. A control with width $768$ for both policies yields a comparable $4.01\times$ speedup. These measurements concern policy computation, not physical motion duration or training time to a target success rate.}

\paragraph{Matched-state evaluation and metrics.}
{For every training seed and sampler, we evaluate $128$ rollouts from each of the same $128$ independently sampled training-reset configurations. Every rollout lasts $100$ physical steps; success is checked at every substep, including within action chunks, without resetting or freezing after success. DA reverse means and probability-flow ODE resample the prior at each decision; SDE additionally samples reverse-step noise. Gaussian REPPO samples its action distribution.}

For each state, count orders $n_{ib},n_{ir}$ at \emph{first strict success}, with $n_i=n_{ib}+n_{ir}$. Report success $p_i=n_i/128$, entropy $H_i=-\sum_{m\in\{b,r\}}q_{im}\log_2q_{im}$ for $q_{im}=n_{im}/n_i$, and success-weighted entropy $J_i=p_iH_i$.
{We first average $p_i$ and $J_i$ across initial states separately for each trained policy, then report the equal-weight mean and sample standard deviation across the three training seeds (Fig.~\ref{fig:stackcube_metrics}). These error bars describe between-seed variability, not $95\%$ confidence intervals. In general, $\overline J\ne\overline p\,\overline H$. If $n_i=0$, $H_i$ is undefined and $J_i=0$; the state remains in the averages of success and $J$. Modes are never pooled across initial states or training seeds. Full-horizon returns include failed rollouts and retain both raw and normalized units.}

\paragraph{Spatial robustness and successful diversity.}
Figure~\ref{fig:stackcube_xy_heatmaps} uses for all five methods and $512$ rollouts at each of $540$ valid blue-cube positions on a uniform $2.5$\,cm grid. Red remains at $(0,-35)$\,cm; the robot pose and cube orientations are fixed, with initial end effector $(0,0,17.0)$\,cm. Each rollout lasts $100$ physical steps. These are controlled configuration slices, not IID resets, and no spatial interpolation is applied. At the original $70$\,cm separation, unchunked \dareppo{} succeeds in $492/512$ attempts ($pH=0.424$ bits), whereas Gaussian REPPO succeeds in $18/512$ ($pH=0.030$ bits). The Gaussian policy's conditional entropy is nevertheless $H=0.852$ bits, illustrating why success must accompany a diversity measure. Distance alone does not explain the spatial pattern; the mixed-order band describes initial configurations, not a motion path.

\noindent{TruDi's multimodality evaluation always executes Gaussian mean actions, while diffusion ODE sampling retains prior randomness \citep[p.~8]{le2026trust}. This is not a matched comparison of behavioral diversity. Fixing the diffusion prior to its mean and taking the mean at every reverse step would likewise produce a single trajectory from a fixed initial state under deterministic dynamics. Zero entropy under such execution does not establish collapse of the learned stochastic policy. We therefore evaluate REPPO using stochastic samples from its Gaussian policy. Stochastic Gaussian REPPO can still realize both orders: action noise can break symmetry and feedback sustain different trajectories. This caveat concerns conditional diversity, not return quality.}
\label{app:stackcube_sampler_caveat}

\clearpage
\paragraph{Spatial comparison across methods.}
{The XY maps reveal configuration-dependent successful diversity. Unchunked \dareppo{} and \dawpo{} exhibit extended mixed-order regions; \dappo{} solves most placements but usually favors one order. Chunked \dareppo{} also samples both orders, but its success deteriorates at large positive blue-cube $y$. The $pH$ maps distinguish reliably successful diversity from balanced orders supported by only a few successful attempts.}

\begin{figure}[H]
    \normalcolor
    \centering
    \includegraphics[width=\linewidth]{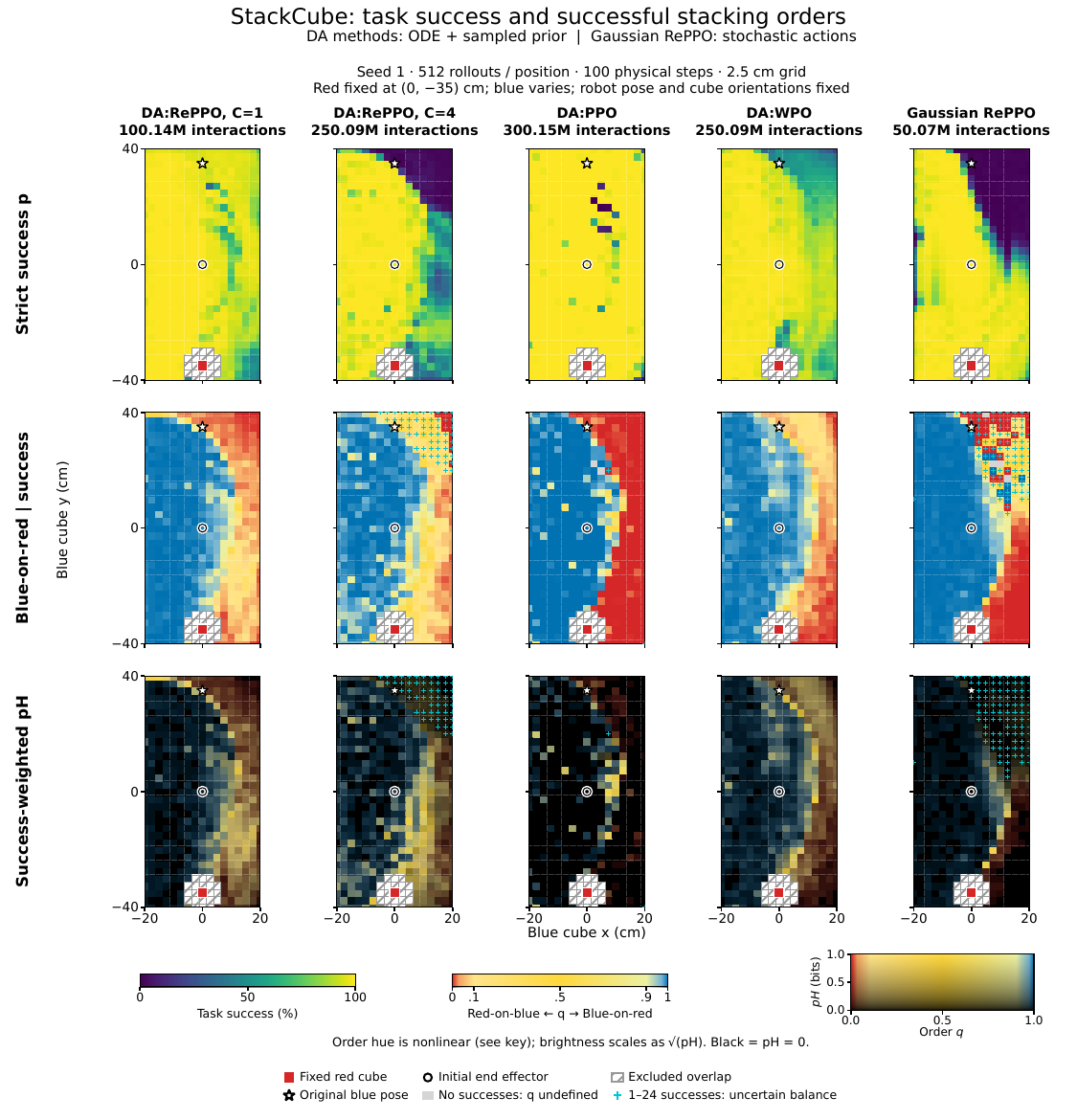}
    \caption{{StackCube spatial evaluation: $540$ positions, $512$ rollouts per position. Rows show strict success $p$, blue-on-red fraction $q$ among first strict successes, and success-weighted entropy $pH$. Bottom-row hue encodes $q$ and brightness scales as $\sqrt{pH}$; black means $pH=0$. Gray denotes no successes; cyan crosses flag fewer than $25$ successes. Hatching excludes overlaps. Circles, stars, and squares mark initial end-effector, blue-cube, and red-cube positions. Budgets differ across methods.}}
    \label{fig:stackcube_xy_heatmaps}
    \label{fig:stackcube_additional_xy_heatmaps}
\end{figure}

\label{app:stackcube_end}
\endgroup
\clearpage

\subsubsection{PushT Orientation-Sweep Multimodality Experiment}
\label{app:pusht_multimodality}

\paragraph{Setup and training.}
{We follow the PushT multimodality setup of \citet{le2026trust}. A Panda robot pushes a T-shaped block until it overlaps at least $90\%$ of a T-shaped goal. Block and goal positions and orientations are randomized during training, and state observations include their poses and the robot state. Training episodes last at most $100$ steps. The dense reward combines orientation and position alignment with an end-effector proximity bonus; success receives raw reward $3$, and training uses rewards divided by $3$. Success resets the episode while retaining the final observation for bootstrapping. {We train three seeds per method and evaluate their final checkpoints: unchunked \dareppo{} at $250.35$M interactions and Gaussian REPPO at $250.09$M.} Both use $1024$ environments, $128$ rollout steps, $\gamma=0.9$, $\lambda=0.98$, and eight optimization epochs. The DA policy uses eight diffusion steps (App.~\ref{app:common_maniskill_hyperparameters}); the Gaussian baseline uses $512$-wide actor and critic networks, configured entropy target $-1$, and KL bound $0.1$. Budgets and evaluation conditions are matched approximately; this is not an otherwise identical hyperparameter ablation.}

\paragraph{IID training-distribution evaluation.}
{For each of six trained policies, we generate $128$ rollouts from each of $128$ shared configurations drawn IID from native training resets, randomizing block and goal poses as during training. Each rollout restores the initial state and lasts $200$ physical steps without resetting or freezing after success. \dareppo{} uses ODE sampling with a fresh prior per action; Gaussian REPPO samples stochastic actions. Using the first-success CW/CCW classifier defined below, we compute success $p_i$ and successful-direction entropy $H_i$ within each configuration. We average $p_i$ and $p_iH_i$ over configurations separately for each policy, then report the equal-weight mean and sample SD across three training seeds. Modes are never pooled across configurations. With no successes, $H_i$ is undefined and $p_iH_i=0$. The evaluation comprises $98{,}304$ rollouts.}

\paragraph{Orientation-sweep evaluation.}
we fix the block and goal centers at $(-0.1,-0.05)$\,m, the goal yaw at zero, and the robot pose at its test value. We evaluate $512$ rollouts from each of $32$ initial block yaws between $30^\circ$ and $330^\circ$; Fig.~\ref{fig:pusht_motion_and_metrics} displays $130^\circ$--$260^\circ$. \dareppo{} uses ODE sampling with a fresh Gaussian prior per action; REPPO samples its Gaussian action distribution. Each rollout lasts $200$ physical steps without resetting or freezing after success. We record whether task success occurs and classify direction by the sign of accumulated unwrapped yaw at the \emph{first success}. Failed trajectories receive no mode label.

For $N$ attempts with $n_{\rm CW}$ and $n_{\rm CCW}$ successful directional rollouts, $p=(n_{\rm CW}+n_{\rm CCW})/N$ and $q=n_{\rm CW}/(n_{\rm CW}+n_{\rm CCW})$. We plot $n_{\rm CW}/N$, $n_{\rm CCW}/N$, and $pH$, where $H=-q\log_2 q-(1-q)\log_2(1-q)$. Thus the two directional frequencies sum to success probability, and $pH=0$ with no successes. Probability bands are $95\%$ Wilson intervals. For $pH$, we map separate $97.5\%$ Clopper--Pearson intervals for $p$ and $q$ through $pH(q)$, giving a conservative $95\%$ interval by a Bonferroni bound.

\paragraph{Results.}
At $190^\circ$, \dareppo{} records $457$ CW and $51$ CCW successes ($508/512$, $pH=0.467$ bits). At $192^\circ$, the dominant direction reverses: $10$ CW and $500$ CCW successes ($510/512$, $pH=0.139$ bits). The grid therefore brackets the switching region but does not resolve the exact equal-probability angle or the entropy maximum between these samples. Gaussian REPPO records only $6/512$ successes at $190^\circ$ and $7/512$ at $192^\circ$, despite high success at outer angles ($511/512$ at both $130^\circ$ and $260^\circ$). It has no observed successes at $155^\circ$, $160^\circ$, or $165^\circ$. Thus the failure is concentrated in a range of orientations rather than explained by uniformly poor task performance.

\paragraph{{Earlier deterministic Gaussian REPPO control.}}
{As a separate diagnostic, the earlier $200.02$M REPPO checkpoint was evaluated using mean actions, $a_t=\tanh(\mu_\theta(s_t))$, from the same fixed poses for $200$ steps. It succeeded in $0/512$ executions at every tested yaw from $145^\circ$ through $225^\circ$, but in $512/512$ at $135^\circ$, $240^\circ$, and $255^\circ$. These are repeated deterministic executions, not independent policy samples; we assign no policy-sampling confidence intervals. This historical control does not evaluate the final $250.09$M checkpoint used in Fig.~\ref{fig:pusht_motion_and_metrics}.}

\clearpage
\subsubsection{{Manipulation Training Curves and Evaluation Checkpoints}}
\label{app:manipulation_learning_curves}
\begin{figure}[H]
    \centering
    \includegraphics[width=\linewidth]{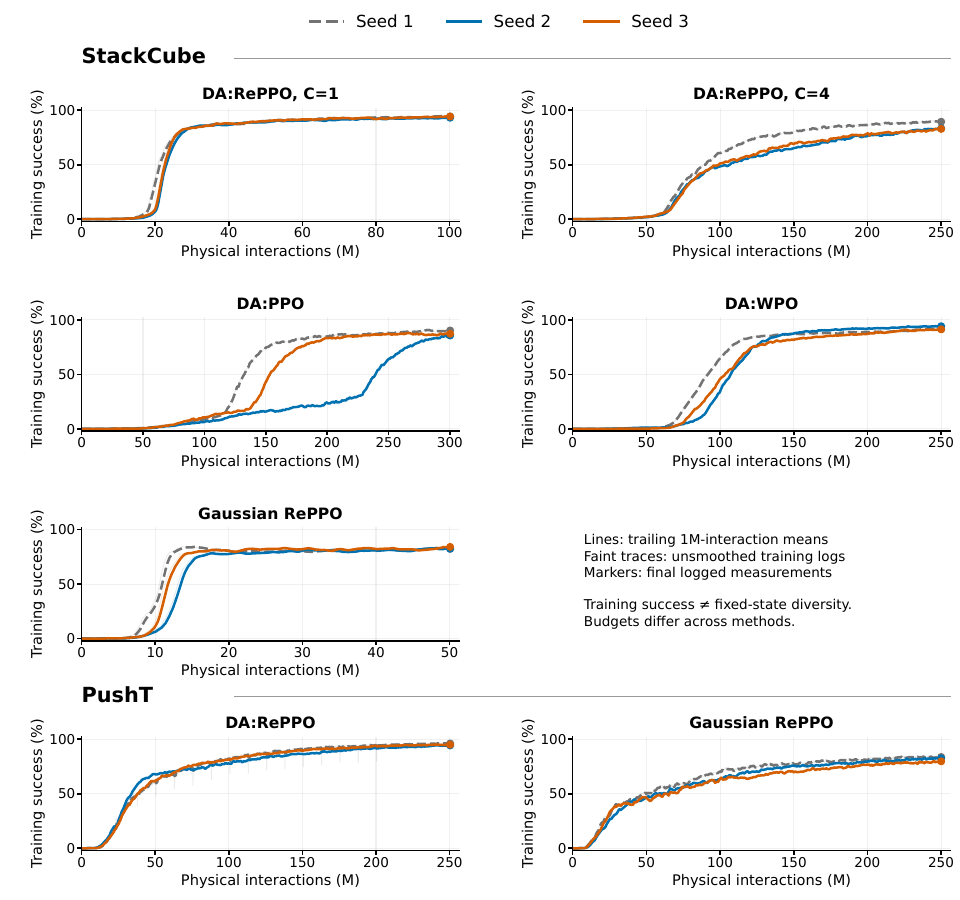}
    \caption{{Training success for all manipulation runs, grouped by task: StackCube above and PushT below. Lines show trailing $1$M-interaction means; faint traces show raw logs and markers identify final measurements. Original and resumed stages share a physical-interaction axis. Curves use randomized training starts and are not fixed-state multimodality evaluations. All three seeds are shown separately; panel budgets differ.}}
    \label{fig:manipulation_training_success}
\end{figure}

{Figure~\ref{fig:manipulation_training_success} reports all three training seeds for each manipulation configuration. We evaluate the final checkpoint at the stated budget, not a checkpoint selected by held-out success or mode entropy. StackCube budgets are $50.07$M interactions for Gaussian REPPO, $100.14$M for unchunked \dareppo{}, $250.09$M for chunked \dareppo{} and \dawpo{}, and $300.15$M for \dappo{}; both PushT methods receive approximately $250$M. These method-specific budgets assess performance after extended training, rather than equal-budget sample efficiency.}

{Most success curves approach a plateau, but training ends at a prescribed budget rather than a convergence-triggered stopping rule. Comparing the final two non-overlapping $10$M-interaction windows, PushT gains range from $0.18$ to $0.50$ percentage points for \dareppo{} and $0.12$ to $0.72$ for Gaussian REPPO. Some StackCube seeds continue to improve: \dappo{} seed 2 gains $2.12$ points and chunked \dareppo{} seed 3 gains $1.46$ points. These diagnostics concern success on randomized training episodes; they establish neither convergence of mode entropy nor uniform reliability on fixed-state slices.}

\clearpage

\subsubsection{Diffusion-Step Ablation Details}
\label{app:diffusion_step_details}

\paragraph{Training protocol.}
We vary the diffusion depth $K\in\{2,4,8\}$ on G1JoystickFlatTerrain. \dareppo{} uses seven training seeds and $50.07$M environment interactions per run; {\dappo{} uses eight seeds and $200.02$M interactions.} Within each method, architectures, hyperparameters, and optimizer-update counts are fixed across depths, with all diffusion transitions included. Both use $1024$ environments, $\gamma=0.98$ and a Gaussian prior standard deviation of $2.5$; domain randomization and external pushes are disabled. {The environment-action clipping bound is $1.0$, with no additional interior clamp.} {The \dappo{} ablation uses the clipped-forward PPO variant, which restricts the forward correction to the PPO active set. Budgets and optimization settings differ between methods, so these are within-method depth comparisons.} {\dawpo{} uses six seeds and $128$-step rollouts, with $\lambda=0.98$ and the same environment count, discount, prior, and randomization settings.}

\begin{figure}[htbp]
    \centering
    \includegraphics[width=\linewidth]{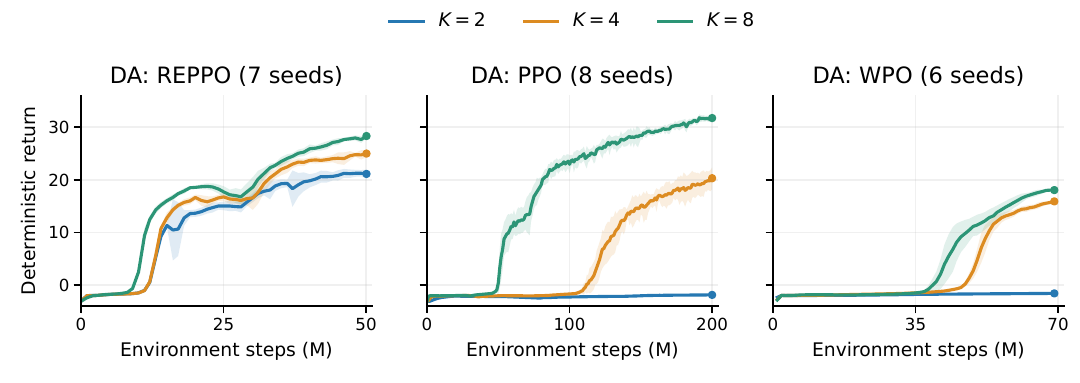}
    \caption{{G1 diffusion-step ablation: deterministic evaluation learning curves for \dareppo{}, \dappo{}, and \dawpo{} (left to right), using seven, eight, and six training seeds. Colors indicate $K$; bands show $\pm1$ sample standard deviation, without temporal smoothing. Panels share a return scale and use method-specific interaction ranges. Endpoint markers correspond to the table in Fig.~\ref{fig:diffusion_step_ablation}: $50.07$M, $200.02$M, and $69.07$M interactions. WPO is displayed through $70$M. Evaluation follows prior and reverse-kernel means and excludes entropy bonuses.}}
    \label{fig:g1_diffusion_learning_curves}
    \label{fig:g1_wpo_evaluation}
\end{figure}

\paragraph{Deterministic checkpoint evaluation.}
We initialize at the Gaussian prior mean and apply the conditional mean of each reverse Gaussian kernel. Evaluations occur approximately every $1$M interactions and record the first episode from each of $128$ environments, ending on failure or at $1000$ steps. Returns sum environment rewards without entropy bonuses. {Figure~\ref{fig:g1_diffusion_learning_curves} shows the evaluation learning curves separately for each method, with a shared return scale and method-specific interaction ranges. For \dareppo{} and \dappo{}, the curves include the initial policy, periodic evaluations, and dedicated final-checkpoint evaluations. The \dawpo{} curves use periodic evaluations within the $0$--$70$M display window, whose last common logged checkpoint is $69.07$M. All six WPO seeds contribute equally, with evaluation times aligned by linear interpolation within their observed ranges. Figure~\ref{fig:diffusion_step_ablation} summarizes the returns at these reported checkpoints.} Results report the mean and sample standard deviation across all training seeds, without temporal smoothing.

\subsubsection{Multimodal Agent Environment Details}
\label{app:double_well}

We provide the full specification of the environment used in Sec.~\ref{sec:toy_example}.

\paragraph{State and Action Space.}
The state is an orientation on the circle, represented by $\mathcal S=\{0^\circ,45^\circ,90^\circ,\ldots,315^\circ\}$, with angles identified modulo $360^\circ$. The action $a_t \in [-1,1]$ represents a relative movement direction, mapped to angles in $[-90^\circ, 90^\circ]$.

\paragraph{Reward Function.}
{For $a\in[-1,1]$, the state-independent reward is
\[
R_{\mathrm{env}}(s,a)=
\begin{cases}
1-0.15(1-4a^2)^2,& |a|\leq\tfrac12,\\
1-\left(\dfrac{4a^2-1}{3}\right)^2,& \tfrac12<|a|\leq1.
\end{cases}
\]
It has two symmetric maxima at $a=\pm\tfrac12$, corresponding to $\pm45^\circ$. The positive-temperature maximum-entropy optimum is bimodal; reward maximization alone also admits policies that select only one mode.}

\paragraph{Transition Dynamics.}
After an action is taken, the environment maps it to the nearest mode of the reward landscape, while the reward is evaluated at the original (continuous) action:
{\small
\begin{equation*}
    {s_{t+1}} =
    \begin{cases}
        (s_t + 45^\circ)\bmod 360^\circ & \text{if } a_t > 0 \\
        (s_t - 45^\circ)\bmod 360^\circ & \text{if } a_t \leq 0
    \end{cases}
\end{equation*}
}

\paragraph{Discrete State Space.}
The modulo convention closes the dynamics on these eight states; for example, $315^\circ+45^\circ\equiv0^\circ$ and $0^\circ-45^\circ\equiv315^\circ$ modulo $360^\circ$. This finite circular state space allows explicit visualization of the action distributions conditioned on each state.

\begin{figure*}[htbp]
    \centering
    \begin{subfigure}[b]{1.\textwidth}
        \includegraphics[width=\textwidth]{Figures/Ablations/multi_agent_hist_extended.png}
        \caption{Action distributions and bimodal reward landscape.}
    \end{subfigure}
    \\
    \begin{subfigure}[b]{1.\textwidth}
        \includegraphics[width=\textwidth]{Figures/Ablations/multi_agent_extended.png}
        \caption{Resulting agent trajectories.}
        \label{fig:multimodal_trajectories}
    \end{subfigure}
    \\
    \begin{subfigure}[b]{1.\textwidth}
        \includegraphics[width=\textwidth]{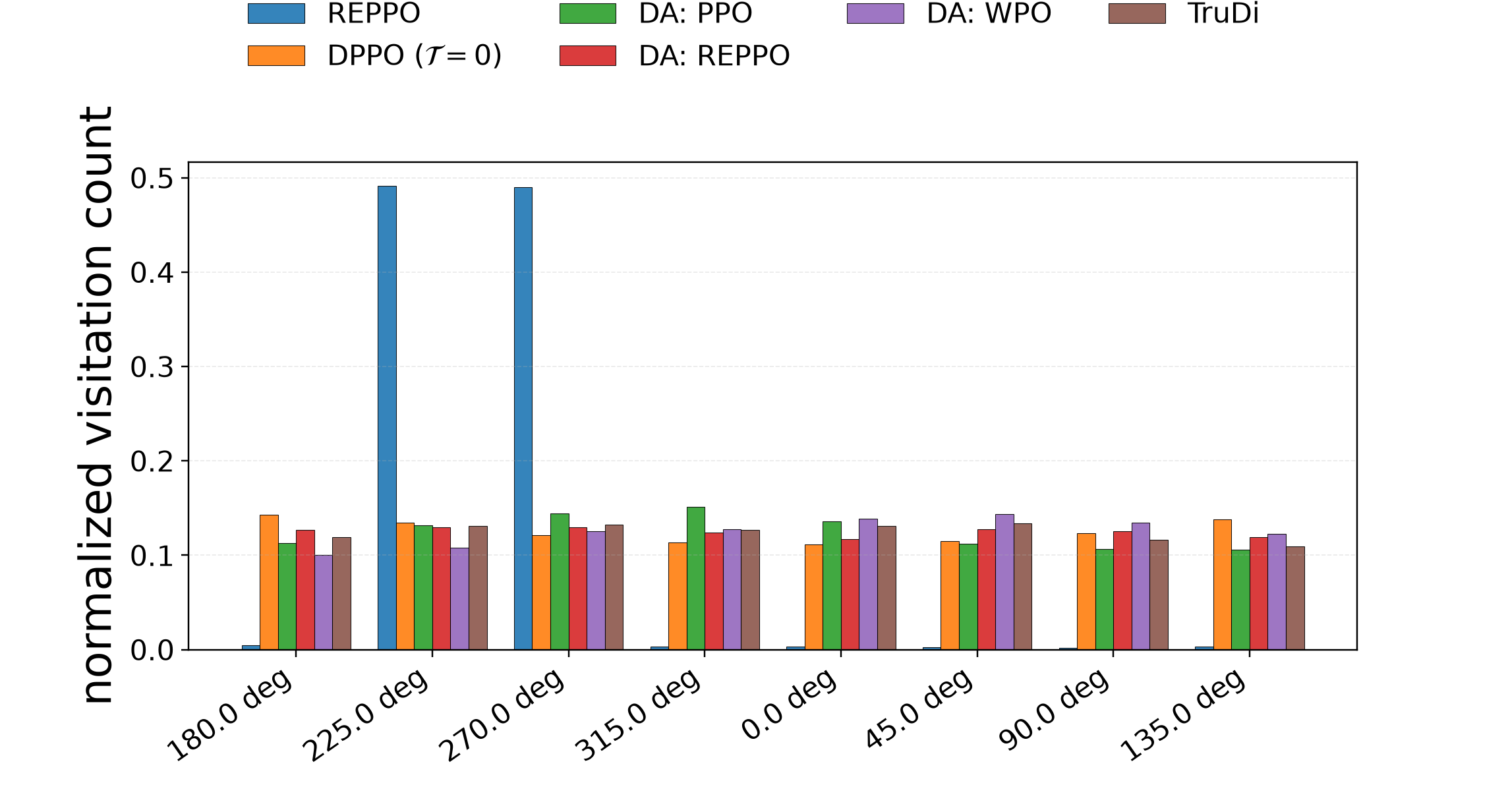}
        \caption{State visitation histogram of each method on the Multimodal Agent Benchmark.}
        \label{fig:state_entropy}
    \end{subfigure}
    \caption{Extended Results on the Multimodal Agent benchmark, which show that \dawpo{} and \replaceReppoDime{} are also able to cover the multimodal action distribution. Panel (c) complements the action and trajectory visualizations with normalized state-visitation counts; it illustrates why high state entropy alone does not imply multimodal action coverage.}
    \label{fig:extended_multimodal}
\end{figure*}

\paragraph{{Evaluating Multimodality.}}
{This benchmark illustrates why reward and state entropy are insufficient for evaluating multimodal policies. Selecting only one of two equally good actions can achieve the optimal reward, and DPPO attains high state entropy by cycling through the discrete states while remaining unimodal in its action distribution (Fig.~\ref{fig:extended_multimodal}). The entropy of state--action trajectories would provide a more informative measure of multimodality, although it is difficult to estimate reliably in high-dimensional control problems.}

\subsubsection{DPI Gap for the KL Divergence}
\label{app:dpi_gap_quantification}

\begingroup
App.~\ref{app:dpi_gap_derivation} proves that the joint KL over $(a_{0:T}^{0:K},s_{0:T+1})$ decomposes exactly into the marginal KL over the executed actions $a_{0:T}$ and two nonnegative conditional KL terms. We restate the decomposition here to define the quantities used in the numerical evaluation.

The resulting identity is
\begin{align}
&D_{\mathrm{KL}}\!\left(
q_\theta(a_{0:T}^{0:K},s_{0:T+1})
\middle\|
\pi_\theta(a_{0:T}^{0:K},s_{0:T+1})
\right)
\nonumber\\
&\quad=D_{\mathrm{KL}}\!\left(q_\theta(a_{0:T})\middle\|\pi(a_{0:T})\right)
+\Delta_{\mathrm{state}}+\Delta_{\mathrm{diffusion}}.
\label{eq:exact_dpi_gap_decomposition}
\end{align}
The two terms are
\begin{align*}
\Delta_{\mathrm{state}}
&:=\mathbb E_{q_\theta(a_{0:T})}\!\left[
D_{\mathrm{KL}}\!\left(
q_\theta(s_{0:T+1}\mid a_{0:T})
\middle\|
\pi(s_{0:T+1}\mid a_{0:T})
\right)
\right],\\
\Delta_{\mathrm{diffusion}}
&:=\mathbb E_{q_\theta(a_{0:T},s_{0:T+1})}\!\left[
D_{\mathrm{KL}}\!\left(
q_\theta(a_{0:T}^{1:K}\mid a_{0:T},s_{0:T+1})
\middle\|
\pi_\theta(a_{0:T}^{1:K}\mid a_{0:T},s_{0:T+1})
\right)
\right].
\end{align*}
Thus, $\Delta_{\mathrm{state}}$ measures the discrepancy between the conditional state trajectories after fixing $a_{0:T}$, while $\Delta_{\mathrm{diffusion}}$ measures the remaining discrepancy between the conditional intermediate diffusion variables after additionally fixing $s_{0:T+1}$. Their sum is exactly the DPI gap, and both terms are nonnegative. The joint bound is tight if and only if, almost surely under the corresponding $q_\theta$ marginals,
\begin{equation}
q_\theta(s_{0:T+1}\mid a_{0:T})
=\pi(s_{0:T+1}\mid a_{0:T}),
\qquad
q_\theta(a_{0:T}^{1:K}\mid a_{0:T},s_{0:T+1})
=\pi_\theta(a_{0:T}^{1:K}\mid a_{0:T},s_{0:T+1}).
\label{eq:dpi_gap_equality_conditions}
\end{equation}
Consequently, two policies with the same executed-action marginal can have different joint KLs because their conditional state or intermediate diffusion distributions differ. If $q_\theta(a_{0:T}^{0:K},s_{0:T+1})=\pi_\theta(a_{0:T}^{0:K},s_{0:T+1})$, the joint KL is zero. Marginalizing over the states and intermediate diffusion variables then yields $q_\theta(a_{0:T})=\pi(a_{0:T})$. The finite-horizon DPI bound and the stationary discounted local-gradient identity are
\begin{align}
D_{\mathrm{KL}}\!\left(q_\theta(a_{0:T})\middle\|\pi(a_{0:T})\right)
&\leq D_{\mathrm{KL}}\!\left(
q_\theta(a_{0:T}^{0:K},s_{0:T+1})
\middle\|
\pi_\theta(a_{0:T}^{0:K},s_{0:T+1})
\right),
\label{eq:dpi_gap_bound_appendix}\\
\left.\nabla_\theta{\mathcal L_{\mathrm{DA}}(\theta;\thetar)}\right|_{\theta=\thetar}
&=
-\left.\nabla_\theta\mathcal J_{\mathrm{DA}}(\theta)\right|_{\theta=\thetar}.
\label{eq:dpi_gap_local_gradient_appendix}
\end{align}

\subsubsection{Numerical Quantification on the Multimodal Agent Benchmark}
\label{app:dpi_gap_numerical}
We quantify every term of Eq.~\ref{eq:exact_dpi_gap_decomposition} on the Multimodal Agent benchmark using one trained policy per method, horizon $H=100$, $K=8$, and $\Tau=0.1$. We use numerical integration to compute the target partition function and the per-state executed-action density $q_\theta(a_t^0\mid s_t)$, marginalizing over all eight continuous intermediate variables. We estimate the state gap directly from the posterior over the eight initial headings using $262{,}144$ sampled action sequences. We then obtain the action-sequence KL by subtracting this gap from the numerically integrated state--action KL. The joint KL over $(a_{0:T}^{0:K},s_{0:T+1})$ is evaluated from the tractable local log-density ratios; for REPPO, which has no intermediate diffusion variables, the analogous quantity is its state--action trajectory KL. All entries below are unscaled KLs in nats, and $\pm$ denotes Monte Carlo standard error for the fixed policy, conditional on the numerical density estimates. REPPO's Monte Carlo error is negligible at the reported precision; numerical integration error is not included.

\begin{table}[H]
\centering
\small
\caption{Numerical estimates of the DPI-gap decomposition on the Multimodal Agent benchmark. The trajectory KL in the final column is method-specific because only DA-MDP contains auxiliary diffusion variables. {These stochastic checkpoints achieve mean per-step environment rewards of $0.9981$ (REPPO) and $0.9572$ (\dareppo{}).}}
\label{tab:dpi_gap_quantification}
\begin{tabular}{@{}lrrrr@{}}
\toprule
Method & \makecell{Marginal action-\\sequence KL $\downarrow$} & \makecell{State gap\\$\downarrow$} & \makecell{Diffusion gap\\$\downarrow$} & \makecell{Method-specific\\trajectory KL $\downarrow$} \\
\midrule
REPPO & $927.01125$ & $2.07944$ & N/A & $929.091$ \\
\dareppo{} & $3.74249\pm0.00012$ & $0.06882\pm0.00012$ & $284.603\pm0.104$ & $288.415\pm0.104$ \\
\bottomrule
\end{tabular}
\end{table}

For \dareppo{}, the numerical identity is therefore
$
288.415 \approx 3.742 + 0.069 + 284.603
$
up to rounding. The action marginal closely matches the target even though the auxiliary diffusion conditional makes the augmented bound loose. This is the standard extended-objective behavior: tightness of the path-space variational bound alone does not determine the quality of the terminal marginal, and a tighter path bound need not produce better marginal samples. In particular, the final column of Table~\ref{tab:dpi_gap_quantification} does not compare the same divergence across the two methods.

Numerical integration gives $\log Z_H=985.152$. \dareppo{}'s action-sequence and augmented-trajectory ELBOs are therefore $981.410$ and $696.737$, respectively. Directly estimating this partition function from $131{,}072$ augmented-trajectory importance samples yields an effective sample size of only $1.24$, illustrating the severe weight degeneracy even in this one-dimensional task.

In a general continuous-control task, evaluating the executed-action marginal requires
\begin{equation}
q_\theta(a_{0:T})=
\int q_\theta(a_{0:T}^{0:K},s_{0:T+1})\,
d a_{0:T}^{1:K}\,d s_{0:T+1}.
\label{eq:intractable_action_marginal}
\end{equation}
This integral contains the complete continuous state trajectory and $HK$ auxiliary diffusion variables; its dimension grows with horizon, state and action dimension, and diffusion depth, while the target normalizer is also unknown. The exact marginal action-sequence KL and DPI gap are therefore computationally intractable on our high-dimensional, long-horizon control benchmarks. The augmented-trajectory KL remains tractable precisely because its log density decomposes into local Markov transition terms.
\endgroup

\subsubsection{Numerical Verification of the Surrogate Loss}
\label{app:numerical_surrogate_verification}

\begingroup
We numerically verify that, at $\theta=\thetar$, the gradients of the local surrogate and the augmented DA-MDP objective are equal in expectation, including when the forward diffusion kernels are learned.
Specifically, we test
\begin{equation}
\left.\nabla_\theta{\mathcal L_{\mathrm{DA}}(\theta;\thetar)}\right|_{\theta=\thetar}
=
-\left.\nabla_\theta\mathcal J_{\mathrm{DA}}(\theta)\right|_{\theta=\thetar}.
\label{eq:surrogate_gradient_identity}
\end{equation}

\paragraph{Exact enumeration setup.}
We verify Eq.~\ref{eq:surrogate_gradient_identity} on an exactly enumerable finite DA-MDP. The task has three environment decisions ($T=2$), three reverse-diffusion steps ($K=3$), two environment states, five latent-action grid points, action-dependent transitions, a tilted double-well reward, and temperature $\Tau=0.7$. Both a small reverse model and one diffusion coefficient per step are learned. We use the VP-SDE forward process, discretized on the latent-action grid (App.~\ref{app:vp_sde}).
We can therefore compute the rollout occupancies, soft Q-values, and expected gradients by enumerating all trajectories in 64-bit arithmetic.

At $\theta=\thetar$, we compare the gradient of the augmented-trajectory objective with (i) the local surrogate in Eq.~\ref{eq:diffusion_surrogate} and (ii) an intentionally incorrect loss that omits the explicit forward-kernel gradient.

\begin{table}[H]
\centering
\small
\caption{Comparison of expected gradients at $\theta=\thetar$. Relative error is $\lVert g-g_{\mathrm{traj}}\rVert_2/\lVert g_{\mathrm{traj}}\rVert_2$.}
\label{tab:surrogate_gradient_verification}
\begin{tabular}{@{}llrr@{}}
\toprule
Comparison & Parameter block & Cosine & Relative error \\
\midrule
Local surrogate & Joint & $1.0000$ & $1.16\times10^{-16}$ \\
Local surrogate & Diffusion coefficients & $1.0000$ & $2.21\times10^{-16}$ \\
Forward gradient omitted & Diffusion coefficients & $0.2065$ & $1.054$ \\
\bottomrule
\end{tabular}
\end{table}

The correct surrogate and trajectory gradients thus agree to machine precision, whereas omitting the explicit $-\log\pi_{\beta_k}$ gradient produces a relative error greater than one. Independent Monte Carlo gradient estimates converge to the values obtained by exact enumeration: with $16{,}384$ samples, the median cosine with the exact gradient is $0.999244$ for the trajectory estimator and $0.999775$ for the local surrogate. Finally, at $\lVert\theta-\thetar\rVert_2=0.4$, the median relative error of the stale surrogate is $9.74\%$, while recomputing the occupancy and Q-function at the updated parameters and resetting $\thetar=\theta$ restores agreement to $1.75\times10^{-16}$. Thus, the local surrogate is exact when $\theta=\thetar$, while reusing a stale rollout makes it only a local approximation.

\paragraph{Rationale for the Unnormalized Local Target.}
For completeness, with $u=a_t^{k-1}$, the learned-forward local target and the corresponding unnormalized KL are
\begin{equation}
\begin{aligned}
\widetilde p_{\theta;\thetar}(u\mid\tilde s)
&:=\pi_\theta(a_t^k\mid u,s_t)
\exp\!\left(\alpha Q_{\mathrm{DA}}^{q_{\thetar}}(\tilde s,u)\right),\\
D_{\mathrm{KL}}^{\mathrm u}(q\|\widetilde p)
&:=\int q(u)\log\frac{q(u)}{\widetilde p(u)}\,du.
\end{aligned}
\label{eq:unnormalized_kl_definition}
\end{equation}
Normalizing $\widetilde p_{\theta;\thetar}$ would introduce a parameter-dependent local partition function whose gradient is absent from the augmented-trajectory objective. In the exact enumeration above, differentiating through this normalizer changes the diffusion-coefficient gradient by $9.18\%$, whereas Eq.~\ref{eq:unnormalized_kl_definition} agrees with the trajectory gradient to machine precision. We therefore leave the local target unnormalized.
\endgroup

\subsection{Runtime Comparison}
\label{app:runtime}

\subsubsection{Controlled Active-Memory Comparison}

We compare \dareppo{} and TruDi under a controlled CheetahRun configuration on a single NVIDIA H100 GPU. Both methods use FP32, matched $4\times512$ policy networks, width-1,024 critics, $K=16$, and four samples for the trust-region KL estimate. Each timing covers a complete training block of 131,072 environment transitions. We exclude compilation, discard one warm-up block, and report the median runtime over the next three blocks. Figure~\ref{fig:jax_runtime_sweep} plots this runtime against the maximum active memory reported by JAX.

{TruDi backpropagates through complete diffusion paths, storing intermediate values from every diffusion step. \dareppo{} computes gradients for individual diffusion transitions, requiring less memory per training sample. For either method, increasing the minibatch size reduces the number of sequential optimizer updates needed to process the training data. Lower memory use does not automatically imply faster training. For a fixed number of environment interactions, runtime depends on training throughput, which is influenced by the computation per sample, minibatch size, and GPU utilization. Available memory constrains which minibatch sizes can be used.}

At approximately $1.9$ GiB of active memory, \dareppo{} requires $6.331$ s per block and TruDi requires $6.730$ s. As the memory budget increases, TruDi can process more full-chain samples per minibatch and therefore perform fewer optimizer updates: its runtime falls to $5.289$ s at $4.123$ GiB and $4.388$ s at $16.063$ GiB. {\dareppo{} trains faster under the tighter memory budgets tested here, while TruDi becomes faster when additional memory permits larger minibatches.}

\subsubsection{Controlled NVML Process-Memory Comparison}
\label{app:nvml_memory_runtime}

Section~\ref{sec:memory_runtime} reports active JAX memory to expose the working-set scaling of the two algorithms. Here we additionally report the operational GPU-process footprint observed by NVML. We use the same controlled single-H100 protocol: FP32, four trust-region KL samples, 131,072 environment transitions per block, one warm-up block, and the median runtime of three complete measured blocks, with compilation excluded. JAX's automatic GPU-memory reservation is disabled. The reported NVML memory is the maximum process allocation observed across the three measured blocks and includes JAX allocator pools, the CUDA context, compiled code, and other process-level GPU allocations.

\begin{figure}[!htbp]
    \centering
    \includegraphics[width=0.98\linewidth]{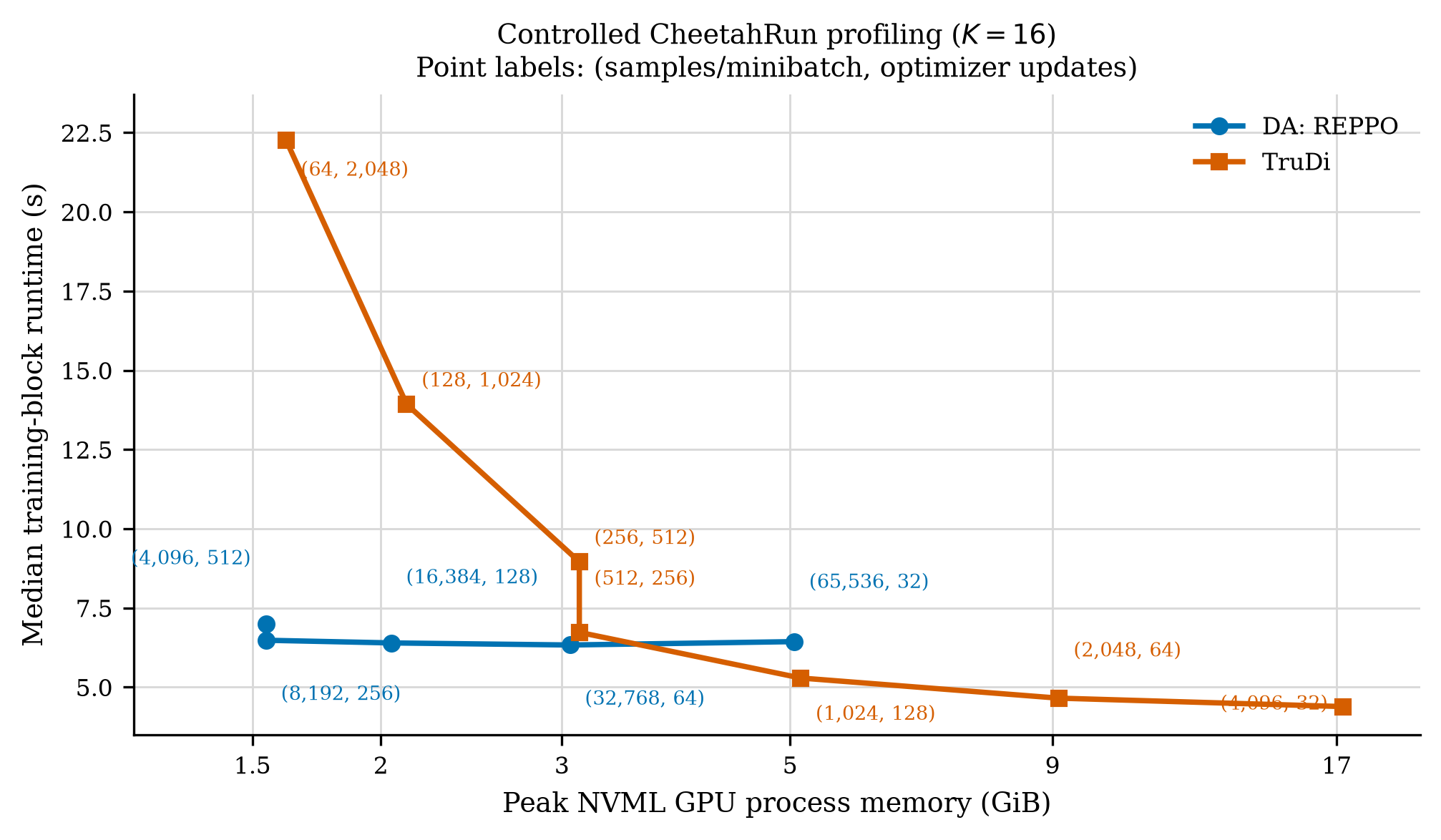}
    \caption{Controlled NVML process-memory--runtime configuration sweep for the complete $K=16$, $4\times512$-policy sweep. {Point labels denote (samples per minibatch, optimizer updates).} Every measured configuration is shown with a filled marker, and lines connect configurations in order of increasing samples per minibatch.}
    \label{fig:nvml_runtime_sweep}
\end{figure}

\begin{table}[H]
    \centering
    \caption{Controlled NVML process-memory--runtime comparison of \dareppo{} and TruDi. DA-MDP minibatches contain diffusion-MDP samples, whereas TruDi minibatches contain environment samples. ``Updates'' denotes the number of minibatch optimizer steps required to process one epoch corresponding to 131,072 environment transitions.}
    \label{tab:nvml_memory_runtime_tradeoff}
    \begin{minipage}{0.99\textwidth}
    \centering
    \textbf{(a) $K=8$, matched $4\times256$ policies}\par\smallskip
    \resizebox{\textwidth}{!}{\begin{tabular}{lrrrrl}
        \toprule
        Method & Minibatches & Samples/minibatch & NVML peak memory & Runtime & Relative result \\
        \midrule
        \dareppo{} & 32 & 32,768 & 3.047 GiB & 3.362 s & $5.74\times$ less memory \\
        TruDi & 16 & 8,192 & 17.502 GiB & 1.646 s & $2.04\times$ faster \\
        \bottomrule
    \end{tabular}}

    \medskip
    \textbf{(b) $K=16$, matched $4\times512$ policies and approximately matched NVML memory}\par\smallskip
    \resizebox{\textwidth}{!}{\begin{tabular}{@{}rrrrrrrrl@{}}
        \toprule
        \makecell{DA samples/\\minibatch} & \makecell{DA\\updates} & \makecell{DA NVML\\memory} & \makecell{DA\\runtime} & \makecell{TruDi samples/\\minibatch} & \makecell{TruDi\\updates} & \makecell{TruDi NVML\\memory} & \makecell{TruDi\\runtime} & Result \\
        \midrule
        8,192  & 256 & 1.547 GiB & 6.481 s & 64    & 2,048 & 1.617 GiB & 22.271 s & DA $3.44\times$ faster \\
        16,384 & 128 & 2.047 GiB & 6.397 s & 128   & 1,024 & 2.119 GiB & 13.928 s & DA $2.18\times$ faster \\
        32,768 & 64  & 3.055 GiB & 6.331 s & 512   & 256   & 3.119 GiB & 6.730 s  & DA $1.06\times$ faster \\
        65,536 & 32  & 5.047 GiB & 6.437 s & 1,024 & 128   & 5.121 GiB & 5.289 s  & TruDi $1.22\times$ faster \\
        \bottomrule
    \end{tabular}}
    \end{minipage}
\end{table}

The NVML results support the same bounded conclusion as the active-memory comparison. At $K=8$, \dareppo{} uses $5.74\times$ less NVML peak process memory, while TruDi is $2.04\times$ faster. At $K=16$ and approximately matched NVML budgets near 1.6, 2.1, and 3.1 GiB, DA-MDP is respectively $3.44\times$, $2.18\times$, and $1.06\times$ faster; near 5.1 GiB, TruDi is $1.22\times$ faster. {In the matched $K=16$ comparison, \dareppo{} achieves shorter runtimes at lower memory budgets, whereas TruDi is faster when more memory is available.}

NVML and JAX memory need not vary smoothly together because NVML includes retained allocator pools. For example, the TruDi configurations $(256,512)$ and $(512,256)$ both occupy $3.119$ GiB according to NVML, although their active JAX peaks are $1.152$ and $1.913$ GiB, respectively. The smaller-minibatch configuration performs twice as many updates and is slower ($8.958$ versus $6.730$ s). Consequently, the connected TruDi sweep contains a vertical segment at $3.119$ GiB in Fig.~\ref{fig:nvml_runtime_sweep}, whereas Fig.~\ref{fig:jax_runtime_sweep} separates the two active working sets. Reporting both metrics distinguishes the algorithmic working set from the whole-process GPU footprint.

\subsubsection{Further Directions for Future Work}
\label{app:limitations}
Another key consideration is optimization stability, particularly when using the HL-Gauss loss \citep{farebrother2024stop} in \dareppo{} or \dawpo{} with automatic temperature tuning. If the temperature overshoots, the log-ratios, which are scaled by the temperature, can grow excessively large and unbounded. This can push $Q$-targets outside the fixed binning range of the HL loss, resulting in numerical instability. While reducing the learning rate for the Lagrange multipliers can partially mitigate this issue, we see this as an important area for future research.
We also found that the minibatch size significantly influences the learned temperature and KL regularization schedule and can make learning unstable, primarily due to noisy updates in the Lagrange multipliers. Smoothing techniques could potentially enhance robustness in this context.
Furthermore, in Eq.~\ref{eq:diffusion_surrogate}, the diffusion time step is currently sampled uniformly. An alternative approach could involve sampling later diffusion steps more frequently using importance sampling, a technique commonly employed in data-based diffusion models \citep{song2021maximum}, which may further improve learning efficiency.

Two architectural extensions could reduce the current computational overhead. First, separate state and noisy-action encoders would allow the state representation to be computed once and reused throughout a diffusion chain. Second, a learned state-dependent prior could initialize the chain closer to useful actions and thereby reduce the required transport. More broadly, the DA-MDP construction is not specific to continuous control: analogous augmented processes could be defined for discrete or structured action spaces, including combinatorial optimization \citep{karalias_erdos_2020} and RL fine-tuning of diffusion language models \citep{ouyang2022training, nie2025large}.

\section{Derivations}
This appendix derives the theoretical identities used by the objectives and algorithms in the main paper. It proceeds from the standard maximum-entropy trajectory KL to its DA-MDP counterpart, establishes the monotonic-improvement result for the DA-MDP surrogate loss, and derives the ME-WPO and \dawpo{} updates.
\subsection{Standard ME-RL Trajectory KL}
\label{app:policygradient}

\subsubsection{Finite-Horizon Identity}
\label{app:reverse_KL}
\label{app:maxent_PGT}

{We first show in the finite-horizon setting that minimizing the KL divergence between the policy and target distributions over joint state--action trajectories is equivalent to maximizing the corresponding maximum-entropy RL objective.} We then derive the local surrogate loss whose gradient matches the objective gradient at the rollout policy.

For a finite horizon $T$, recall the policy trajectory distribution and the unnormalized reward target
\begin{align}
q_\theta(a_{0:T},s_{0:T+1})
&=p(s_0)\prod_{t=0}^{T}
q_\theta(a_t\mid s_t)p(s_{t+1}\mid s_t,a_t),
\label{eq:finite_standard_policy_trajectory}\\
\widetilde\pi(a_{0:T},s_{0:T+1})
&=p(s_0)\prod_{t=0}^{T}
p(s_{t+1}\mid s_t,a_t)
\exp\!\left(\frac{R_{\mathrm{env}}(s_t,a_t)}{\Tau}\right),
\qquad
\pi:=\frac{\widetilde\pi}{\mathcal Z}.
\label{eq:finite_standard_target_trajectory}
\end{align}
The initial-state distribution and environment dynamics are identical in the two trajectories and therefore cancel inside the log ratio. Hence
\begin{equation}
\log\frac{q_\theta(a_{0:T},s_{0:T+1})}
{\pi(a_{0:T},s_{0:T+1})}
=
\sum_{t=0}^{T}\left[
\log q_\theta(a_t\mid s_t)
-\frac{R_{\mathrm{env}}(s_t,a_t)}{\Tau}
\right]
+\log\mathcal Z.
\label{eq:finite_standard_log_ratio}
\end{equation}
Taking the expectation under $q_\theta$ gives
\begin{align}
\Tau D_{\mathrm{KL}}\!\left(
q_\theta(a_{0:T},s_{0:T+1})
\,\middle\|\,
\pi(a_{0:T},s_{0:T+1})
\right)
&=-\mathcal J_T(\theta)+\Tau\log\mathcal Z,
\label{eq:finite_standard_kl_return}\\
\mathcal J_T(\theta)
&:=\mathbb E_{q_\theta}\!\left[
\sum_{t=0}^{T}
\left(R_{\mathrm{env}}(s_t,a_t)
-\Tau\log q_\theta(a_t\mid s_t)\right)
\right].
\label{eq:finite_standard_maxent_return}
\end{align}
Thus, minimizing the joint trajectory KL is exactly equivalent to maximizing the finite-horizon maximum-entropy return, apart from the constant $\Tau\log\mathcal Z$.

For completeness, define the finite-horizon soft value functions by
\begin{align}
Q^{q_\theta}(s_t,a_t)
&=R_{\mathrm{env}}(s_t,a_t)
+\mathbb E_{s_{t+1}\sim p(\cdot\mid s_t,a_t)}
\!\left[V^{q_\theta}(s_{t+1})\right],
\label{eq:finite_standard_soft_q}\\
V^{q_\theta}(s_t)
&=\mathbb E_{a_t\sim q_\theta(\cdot\mid s_t)}\!\left[
Q^{q_\theta}(s_t,a_t)
-\Tau\log q_\theta(a_t\mid s_t)
\right],
\qquad V^{q_\theta}(s_{T+1})=0.
\label{eq:finite_standard_soft_v}
\end{align}
Although the entropy term depends explicitly on $\theta$, its direct derivative vanishes conditionally:
\begin{equation}
\mathbb E_{a_t\sim q_\theta(\cdot\mid s_t)}\!\left[
\nabla_\theta\log q_\theta(a_t\mid s_t)
\right]
=\nabla_\theta\!\int q_\theta(a_t\mid s_t)\,d a_t=0.
\label{eq:finite_standard_score_identity}
\end{equation}
This cancellation removes only the explicit derivative of the entropy term. The policy still changes the distribution of future states, and the causal likelihood-ratio expansion collects that dependence in $Q^{q_\theta}$. Applying the policy-gradient theorem therefore yields
\begin{equation}
\begin{aligned}
&\nabla_\theta\!\left[
\Tau D_{\mathrm{KL}}\!\left(
q_\theta(a_{0:T},s_{0:T+1})
\,\middle\|\,
\pi(a_{0:T},s_{0:T+1})
\right)
\right]
\\[-0.25em]
&\quad=
\sum_{t=0}^{T}
\mathbb E_{q_\theta}\!\left[
\left(
\Tau\log q_\theta(a_t\mid s_t)
-Q^{q_\theta}(s_t,a_t)
\right)
\nabla_\theta\log q_\theta(a_t\mid s_t)
\right].
\end{aligned}
\label{eq:finite_standard_soft_policy_gradient}
\end{equation}

\subsubsection{Reverse Log-Derivative Lemma and Actor Surrogate}
\label{app:reverse_log_derivative}

The policy-gradient integrand in Eq.~\ref{eq:finite_standard_soft_policy_gradient} is itself the gradient of a per-state reverse KL. This reverse application of the usual log-derivative trick produces the actor loss optimized in practice.

\begin{lemma}[Reverse log-derivative trick]
\label{lem:reverse_log_derivative}
Fix a time $t$ and state $s$, freeze the rollout critic $Q^{q_{\thetar}}(s,a)$, and assume that
\begin{equation}
\pi_{Q,t}^{q_{\thetar}}(a\mid s)
:=\frac{\exp\!\left(Q^{q_{\thetar}}(s,a)/\Tau\right)}
{Z_{Q,t}^{q_{\thetar}}(s)},
\qquad
\nabla_\theta Q^{q_{\thetar}}(s,a)=0,
\label{eq:reverse_log_derivative_target}
\end{equation}
is normalizable. Then
{
\begin{equation}
\begin{aligned}
&
\nabla_\theta\!\left[
\Tau D_{\mathrm{KL}}\!\left(
q_\theta(\cdot\mid s)\,\middle\|\,
\pi_{Q,t}^{q_{\thetar}}(\cdot\mid s)
\right)
\right]\\
&\quad=\mathbb E_{a\sim q_\theta(\cdot\mid s)}\!\left[
\left(\Tau\log q_\theta(a\mid s)-Q^{q_{\thetar}}(s,a)\right)
\nabla_\theta\log q_\theta(a\mid s)
\right].
\end{aligned}
\label{eq:reverse_log_derivative_identity}
\end{equation}
}
\end{lemma}
\begin{proof}
Expanding the KL gives the expectation of
$\Tau\log q_\theta(a\mid s)-Q^{q_{\thetar}}(s,a)$ plus the parameter-independent term $\Tau\log Z_{Q,t}^{q_{\thetar}}(s)$. Differentiating this expectation produces the right-hand side of Eq.~\ref{eq:reverse_log_derivative_identity} and the additional term $\Tau\mathbb E_{q_\theta}[\nabla_\theta\log q_\theta(a\mid s)]$, which vanishes by Eq.~\ref{eq:finite_standard_score_identity}.
\end{proof}

Applying Lemma~\ref{lem:reverse_log_derivative} at every time step while freezing the rollout state distributions and critic gives the finite-horizon actor surrogate
\begin{equation}
\mathcal L_T^{\mathrm{ME}}(\theta;\thetar)
:=\Tau\sum_{t=0}^{T}
\mathbb E_{s_t\sim q_{\thetar}}
\left[
D_{\mathrm{KL}}\!\left(
q_\theta(\cdot\mid s_t)\,\middle\|\,
\pi_{Q,t}^{q_{\thetar}}(\cdot\mid s_t)
\right)
\right].
\label{eq:finite_standard_actor_surrogate}
\end{equation}
At the rollout parameters, its gradient is exactly the trajectory-KL gradient, or equivalently the negative return gradient:
\begin{equation}
\begin{aligned}
\left.\nabla_\theta\mathcal L_T^{\mathrm{ME}}(\theta;\thetar)\right|_{\theta=\thetar}
&=\left.\nabla_\theta\!\left[
\Tau D_{\mathrm{KL}}\!\left(
q_\theta(a_{0:T},s_{0:T+1})
\,\middle\|\,
\pi(a_{0:T},s_{0:T+1})
\right)
\right]\right|_{\theta=\thetar}\\
&=-\left.\nabla_\theta\mathcal J_T(\theta)\right|_{\theta=\thetar}.
\end{aligned}
\label{eq:finite_standard_surrogate_gradient_identity}
\end{equation}
This equality is exact only at the first actor iterate, because subsequent updates continue to use the rollout-frozen state distributions and critic. TRPO motivates optimizing the surrogate for multiple updates subject to a rollout-to-current policy KL constraint, for example
\begin{equation}
\frac{1}{T+1}\sum_{t=0}^{T}
\mathbb E_{s_t\sim q_{\thetar}}
\left[
D_{\mathrm{KL}}\!\left(
q_{\thetar}(\cdot\mid s_t)\,\middle\|\,
q_\theta(\cdot\mid s_t)
\right)
\right]
\leq\delta_{\mathrm{KL}}.
\label{eq:finite_standard_trust_region}
\end{equation}
{Following TRPO, we use a rollout-averaged KL constraint as a practical approximation to the worst-state divergence control appearing in the improvement bound \citep{schulman2015trust}.}

\subsubsection{Infinite-Horizon Discounted Extension}
\label{app:infinite_horizon_gaussian}

In this section, we present two ways to extend the finite-horizon result above to the infinite-horizon discounted setting.

\paragraph{Alternative A: deterministic time weighting.}
For each $t\geq0$, define a time-dependent entropy coefficient and scaled reward
\begin{align}
\Tau_t&:=\gamma^t\Tau,
&
R_t(s_t,a_t)&:=\gamma^tR_\mathrm{env}(s_t,a_t).
\label{eq:time_scaled_terms}
\end{align}
For a finite cutoff $T$, the resulting maximum-entropy return is
\begin{align}
\mathcal J_T(\theta)
&:=\sum_{t=0}^{T}
\mathbb E_{q_\theta}\!\left[
R_t(s_t,a_t)
-\Tau_t\log q_\theta(a_t\mid s_t)
\right]
\nonumber\\
&=\sum_{t=0}^{T}\gamma^t
\mathbb E_{q_\theta}\!\left[
R_\mathrm{env}(s_t,a_t)
-\Tau\log q_\theta(a_t\mid s_t)
\right].
\label{eq:time_scaled_maxent_objective}
\end{align}
Assuming
$\mathbb E_{q_\theta}[\sum_{t=0}^{\infty}\gamma^t
|R_\mathrm{env}(s_t,a_t)-\Tau\log q_\theta(a_t\mid s_t)|]<\infty$,
we may take $T\to\infty$ and obtain the standard discounted maximum-entropy objective~\citep{haarnoja2018soft,levine2018reinforcement}:
\begin{align}
\mathcal J(\theta)
&=\mathbb E_{q_\theta}\!\left[
\sum_{t=0}^{\infty}\gamma^t
\left(
R_\mathrm{env}(s_t,a_t)
-\Tau\log q_\theta(a_t\mid s_t)
\right)
\right].
\label{eq:time_scaled_infinite_objective}
\end{align}
To derive the policy gradient of this maximum-entropy RL objective, let
$c_\theta(s_t,a_t):=\Tau\log q_\theta(a_t\mid s_t)-R_\mathrm{env}(s_t,a_t)$
and $g_t:=\nabla_\theta\log q_\theta(a_t\mid s_t)$.
{We first differentiate at a finite cutoff $T$. Applying the log-derivative trick and then the conditional score identity gives}
\begin{equation}
{
\begin{aligned}
-\nabla_\theta\mathcal J_T(\theta)
&=\mathbb E_{q_\theta}\!\left[
\left(\sum_{j=0}^{T}g_j\right)
\left(\sum_{t=0}^{T}\gamma^t c_\theta(s_t,a_t)\right)
\right]
+\Tau\sum_{t=0}^{T}\gamma^t\mathbb E_{q_\theta}[g_t]\\
&=\sum_{t=0}^{T}\gamma^t
\mathbb E_{q_\theta}\!\left[
c_\theta(s_t,a_t)\sum_{j=0}^{t}g_j
\right].
\end{aligned}
}
\label{eq:time_scaled_gradient_full}
\end{equation}
{The direct entropy derivative vanishes because $\mathbb E_{q_\theta}[g_t]=0$, and causality gives $\mathbb E_{q_\theta}[c_\theta(s_t,a_t)g_j]=0$ for $j>t$. Assuming differentiation under expectation is valid and the differentiated terms admit a locally uniform summable bound, we may take $T\to\infty$ and re-index the remaining causal terms to obtain}
\begin{align}
-\nabla_\theta\mathcal J(\theta)
&=\sum_{j=0}^{\infty}\gamma^j
\mathbb E_{q_\theta}\!\left[
g_j\,
\mathbb E_{q_\theta}\!\left[
\sum_{\ell=0}^{\infty}\gamma^\ell
c_\theta(s_{j+\ell},a_{j+\ell})
\,\middle|\,s_j,a_j
\right]
\right]
\nonumber\\
&=\sum_{j=0}^{\infty}\gamma^j
\mathbb E_{q_\theta}\!\left[
\left(
\Tau\log q_\theta(a_j\mid s_j)
-Q^{q_\theta}(s_j,a_j)
\right)
\nabla_\theta\log q_\theta(a_j\mid s_j)
\right],
\label{eq:time_scaled_soft_pg}
\end{align}
where $Q^{q_\theta}$ is defined by the stationary soft Bellman equations below.

{Let $\rho_t^{q_\theta}(s)$ denote the probability
density of the continuous state $s_t$ evaluated at $s$ under $q_\theta$,
with $s_0\sim p(s_0)$. The normalized discounted state-visitation
density is $d^{q_\theta}(s):=(1-\gamma)\sum_{t=0}^{\infty}\gamma^t
\rho_t^{q_\theta}(s)$.}
Assuming normalizability, define at frozen rollout parameters $\thetar$ the soft policy-improvement target
\begin{align}
\pi_Q^{q_{\thetar}}(a\mid s)
&:=\frac{
\exp\!\left(Q^{q_{\thetar}}(s,a)/\Tau\right)}
{Z_Q^{q_{\thetar}}(s)},
\qquad
Z_Q^{q_{\thetar}}(s)
:=\int_{\mathcal A}
\exp\!\left(Q^{q_{\thetar}}(s,a)/\Tau\right)d a.
\label{eq:infinite_boltzmann_actor}
\end{align}
Applying Lemma~\ref{lem:reverse_log_derivative} to Eq.~\ref{eq:time_scaled_soft_pg} gives the infinite-horizon surrogate loss
\begin{align}
{\mathcal L_{\mathrm{ME}}(\theta;\thetar)}
&:=\frac{\Tau}{1-\gamma}
\mathbb E_{s\sim d^{q_{\thetar}}}
\left[
\Dkl\!\left(
q_\theta(a\mid s)
\,\middle\|\,
\pi_Q^{q_{\thetar}}(a\mid s)
\right)
\right].
\label{eq:infinite_me_surrogate}
\end{align}
At the rollout parameters, $\left.\nabla_\theta{\mathcal L_{\mathrm{ME}}(\theta;\thetar)}\right|_{\theta=\thetar}=-\left.\nabla_\theta\mathcal J(\theta)\right|_{\theta=\thetar}$.

\paragraph{Alternative B: a single KL over geometrically stopped trajectories.}
This construction uses the standard connection between geometric horizons and discounted state visitation~\citep{thakoor2022generalised,borrell2025random}. Let $H\in\{0,1,\ldots\}$ be independent of the MDP trajectory and distributed as
\[
w(h):=\Pr(H=h)=(1-\gamma)\gamma^h,
\qquad
\Pr(H\geq t)=\gamma^t.
\]
Write $\tau_h=(s_0,a_0,\ldots,s_h,a_h,s_{h+1})$.
For a stationary policy $q_\theta(a\mid s)$, define the normalized policy trajectory distribution
\begin{align}
q_\theta(h,\tau_h)
&:=w(h)p(s_0)
\prod_{t=0}^{h}
q_\theta(a_t\mid s_t)
p(s_{t+1}\mid s_t,a_t),
\label{eq:geom_q}
\end{align}
and the corresponding unnormalized reward target and its partition function as
\begin{align}
\widetilde\pi(h,\tau_h)
&:=w(h)p(s_0)
\prod_{t=0}^{h}
p(s_{t+1}\mid s_t,a_t)
\exp\!\left(\frac{R_\mathrm{env}(s_t,a_t)}{\Tau}\right),
\label{eq:geom_target}\\
\mathcal Z
&:=\sum_{h=0}^{\infty}\int
\widetilde\pi(h,\tau_h)\,d\tau_h,
\qquad
\pi:=\frac{\widetilde\pi}{\mathcal Z}.
\label{eq:geom_partition}
\end{align}
{Let $V_{\mathcal A}:=\operatorname{vol}(\mathcal A)$ denote the Lebesgue volume of the action space. The compact-action and bounded-reward assumptions give the explicit sufficient condition}
\begin{equation}
{
M:=V_{\mathcal A}e^{R_{\max}/\Tau},
\qquad \gamma M<1,
\qquad
0<\mathcal Z\leq
(1-\gamma)\sum_{h=0}^{\infty}\gamma^hM^{h+1}
=\frac{(1-\gamma)M}{1-\gamma M}<\infty.
}
\label{eq:geometric_normalizability_condition}
\end{equation}
{This condition can always be enforced without changing the optimizer. For any constant $b$ satisfying $\gamma V_{\mathcal A}\exp((R_{\max}-b)/\Tau)<1$, replace $R_{\mathrm{env}}$ by $R_{\mathrm{env}}^{(b)}:=R_{\mathrm{env}}-b$. Then}
\begin{equation}
{
\mathcal J^{(b)}(\theta)
=\mathcal J(\theta)-\frac{b}{1-\gamma}.
}
\label{eq:geometric_reward_shift}
\end{equation}
{Hence the shift makes the geometric target normalizable while leaving every policy gradient and maximizer unchanged.} The initial-state, geometric-horizon, and raw environment-transition factors are identical in Eqs.~\ref{eq:geom_q} and~\ref{eq:geom_target}. They therefore cancel inside the trajectory log ratio, giving
\begin{align}
\log\frac{q_\theta(h,\tau_h)}
{\pi(h,\tau_h)}
&=
\sum_{t=0}^{h}
\left[
\log q_\theta(a_t\mid s_t)
-\frac{R_\mathrm{env}(s_t,a_t)}{\Tau}
\right]
+\log\mathcal Z.
\label{eq:geom_log_ratio}
\end{align}
To make the sampling distributions explicit, let $\tau_\infty=(s_0,a_0,s_1,a_1,\ldots)\sim q_\theta$ denote an infinite trajectory generated by the policy and environment dynamics, and draw $H\sim w$ independently. For any integrable quantity $x_t=x_t(\tau_\infty)$, the stopped sum satisfies
$\sum_{t=0}^{H}x_t=\sum_{t=0}^{\infty}\mathbf 1_{\{H\geq t\}}x_t$.
Assuming
$\mathbb E_{\substack{H\sim w\\\tau_\infty\sim q_\theta}}[\sum_{t=0}^{\infty}\mathbf 1_{\{H\geq t\}}|x_t|]<\infty$,
Fubini's theorem and the independence of $H$ and $\tau_\infty$ give
{
\begin{align}
  \mathbb E_{\substack{H\sim w\\\tau_\infty\sim q_\theta}}
\!\left[\sum_{t=0}^{H}x_t\right]
&=\sum_{t=0}^{\infty}
\mathbb E_{\substack{H\sim w\\\tau_\infty\sim q_\theta}}
\!\left[\mathbf 1_{\{H\geq t\}}x_t\right]
  \nonumber\\
  &=\sum_{t=0}^{\infty}
\Pr_{H\sim w}(H\geq t)\,
\mathbb E_{\tau_\infty\sim q_\theta}[x_t] =\sum_{t=0}^{\infty}\gamma^t
\mathbb E_{\tau_\infty\sim q_\theta}[x_t].
  \label{eq:geom_discount_identity}
\end{align}
}
The last equality uses the geometric survival probability $\Pr_{H\sim w}(H\geq t)=\gamma^t$; this random-horizon representation of discounted objectives is also stated by \citet[Lemma~2.2]{borrell2025random}. Taking the expectation of Eq.~\ref{eq:geom_log_ratio} under $(H,\tau_H)\sim q_\theta(H,\tau_H)$ and using Eq.~\ref{eq:geom_discount_identity} yields
\begin{align}
\Tau\,\Dkl\!\left(
q_\theta(H,\tau_H)
\,\middle\|\,
\pi(H,\tau_H)
\right)
&=-\mathcal J(\theta)+\Tau\log\mathcal Z,
\label{eq:infinite_traj_kl}\\
\mathcal J(\theta)
&:=\mathbb E_{\tau_\infty\sim q_\theta}\!\left[
\sum_{t=0}^{\infty}\gamma^t
\left(
R_\mathrm{env}(s_t,a_t)
-\Tau\log q_\theta(a_t\mid s_t)
\right)
\right].
\label{eq:discounted_maxent_objective}
\end{align}
Thus, minimizing the single trajectory KL in Eq.~\ref{eq:infinite_traj_kl} is exactly equivalent, up to the parameter-independent additive constant $\Tau\log\mathcal Z$ and a sign, to maximizing the infinite-horizon discounted maximum-entropy RL objective.

\paragraph{Stationary soft value functions and policy gradient.}
Because the MDP, reward, and policy are stationary and the geometric stopping distribution is memoryless, the value functions no longer require a time index. They satisfy the discounted soft Bellman evaluation equations~\citep{levine2018reinforcement}
\begin{align}
Q^{q_\theta}(s,a)
&=R_\mathrm{env}(s,a)
+\gamma\,\mathbb E_{s'\sim p(\cdot\mid s,a)}
[V^{q_\theta}(s')],
\label{eq:infinite_soft_q}\\
V^{q_\theta}(s)
&=\mathbb E_{a\sim q_\theta(\cdot\mid s)}
\left[
Q^{q_\theta}(s,a)
-\Tau\log q_\theta(a\mid s)
\right].
\label{eq:infinite_soft_v}
\end{align}
The causal likelihood-ratio expansion of Eq.~\ref{eq:infinite_traj_kl} gives
\begin{align}
&\nabla_\theta\!\left[
\Tau\Dkl\!\left(
q_\theta(H,\tau_H)
\,\middle\|\,
\pi(H,\tau_H)
\right)
\right]
\nonumber\\[-0.25em]
&\quad=\sum_{t=0}^{\infty}\gamma^t
\mathbb E_{q_\theta}\!\left[
\left(
\Tau\log q_\theta(a_t\mid s_t)
-Q^{q_\theta}(s_t,a_t)
\right)
\nabla_\theta\log q_\theta(a_t\mid s_t)
\right].
\label{eq:infinite_soft_pg}
\end{align}
As in the finite-horizon derivation, the explicit entropy derivative vanishes by the conditional score identity; its dependence through future state visitation remains and is contained in $Q^{q_\theta}$.

Using the normalized discounted state-visitation density $d^{q_\theta}$, Eq.~\ref{eq:infinite_soft_pg} is equivalently
\begin{align}
&\nabla_\theta\!\left[
\Tau\Dkl\!\left(
q_\theta(H,\tau_H)
\,\middle\|\,
\pi(H,\tau_H)
\right)
\right]
\nonumber\\[-0.25em]
&\quad=\frac{1}{1-\gamma}
\mathbb E_{\substack{s'\sim d^{q_\theta}\\
a\sim q_\theta(\cdot\mid s')}}
\left[
\left(\Tau\log q_\theta(a\mid s')-Q^{q_\theta}(s',a)\right)
\nabla_\theta\log q_\theta(a\mid s')
\right].
\label{eq:infinite_soft_pg_occupancy}
\end{align}

\subsection{Data-Processing Inequality and DA-MDP Trajectory KL}
\label{app:infinite_horizon_da_mdp}

This section first derives the data-processing inequality from the KL chain rule for a general latent variable. We then specialize it to Gaussian-policy state--action trajectories, where marginalizing the states introduces $\Delta_{\mathrm{state}}$, and to diffusion policies, where marginalizing the intermediate diffusion variables additionally introduces $\Delta_{\mathrm{diffusion}}$. Finally, we instantiate these identities for the finite-prefix DA-MDP trajectory and derive its stationary discounted objective, Bellman equations, actor surrogate, and learned-forward-kernel correction.

\subsubsection{General Latent-Variable DPI from the KL Chain Rule}
\label{app:dpi_gap_derivation}

\begingroup
Let $X$ be an observed random variable and $Z$ a latent variable, with normalized joint densities $q(X,Z)$ and $\pi(X,Z)$. Assume $q(X,Z)$ is absolutely continuous with respect to $\pi(X,Z)$ and the joint KL is finite. The chain-rule factorizations
\begin{equation}
q(X,Z)=q(X)q(Z\mid X),
\qquad
\pi(X,Z)=\pi(X)\pi(Z\mid X)
\label{eq:general_dpi_factorizations}
\end{equation}
give the pointwise identity
\begin{equation}
\log\frac{q(X,Z)}{\pi(X,Z)}
=
\log\frac{q(X)}{\pi(X)}
+
\log\frac{q(Z\mid X)}{\pi(Z\mid X)}.
\label{eq:general_dpi_log_ratio}
\end{equation}
Taking the expectation under $q(X,Z)$ yields the KL chain rule
\begin{align}
D_{\mathrm{KL}}\!\left(q(X,Z)\,\middle\|\,\pi(X,Z)\right)
={}&
D_{\mathrm{KL}}\!\left(q(X)\,\middle\|\,\pi(X)\right)
\nonumber\\
&+
\mathbb E_{X\sim q(X)}\!\left[
D_{\mathrm{KL}}\!\left(
q(Z\mid X)\,\middle\|\,\pi(Z\mid X)
\right)
\right].
\label{eq:general_kl_chain_rule}
\end{align}
The conditional KL in the second line is nonnegative. Therefore
\begin{equation}
D_{\mathrm{KL}}\!\left(q(X)\,\middle\|\,\pi(X)\right)
\leq
D_{\mathrm{KL}}\!\left(q(X,Z)\,\middle\|\,\pi(X,Z)\right),
\label{eq:general_dpi_from_chain_rule}
\end{equation}
which is the data-processing inequality for the deterministic map $(X,Z)\mapsto X$ that discards $Z$. Its exact gap is the expected conditional KL in Eq.~\ref{eq:general_kl_chain_rule}; equality holds if and only if $q(Z\mid X)=\pi(Z\mid X)$ for $q(X)$-almost every $X$.
\endgroup

\subsubsection{Gaussian-Policy Trajectories and $\Delta_{\mathrm{state}}$}
\label{app:dpi_state_gap_derivation}

\begingroup
For the Gaussian-policy algorithms, the conditional action distribution has the form
\begin{equation}
q_\theta(a_t\mid s_t)
=\mathcal N\!\left(
a_t;\mu_\theta(s_t),
\operatorname{diag}(\sigma_\theta(s_t)^2)
\right),
\label{eq:dpi_gaussian_policy}
\end{equation}
and induces the joint state--action trajectory distribution in Eq.~\ref{eq:finite_standard_policy_trajectory}. To apply Eq.~\ref{eq:general_kl_chain_rule}, set
$X=a_{0:T}$ and $Z=s_{0:T+1}$. This gives
{
\begin{equation}
  D_{\mathrm{KL}}\!\left(
q_\theta(a_{0:T},s_{0:T+1})
\,\middle\|\,
\pi(a_{0:T},s_{0:T+1})
\right) =
D_{\mathrm{KL}}\!\left(
q_\theta(a_{0:T})\,\middle\|\,\pi(a_{0:T})
\right)
+\Delta_{\mathrm{state}},
  \label{eq:dpi_gaussian_state_decomposition}
\end{equation}
}
where
\begin{equation}
\Delta_{\mathrm{state}}
:=
\mathbb E_{a_{0:T}\sim q_\theta(a_{0:T})}\!\left[
D_{\mathrm{KL}}\!\left(
q_\theta(s_{0:T+1}\mid a_{0:T})
\,\middle\|\,
\pi(s_{0:T+1}\mid a_{0:T})
\right)
\right].
\label{eq:dpi_state_gap_derivation}
\end{equation}
Thus, $\Delta_{\mathrm{state}}$ is exactly the information lost when the state trajectory is marginalized and only the action sequence is retained. Although the policy and target joint distributions contain the same environment transition kernel, their conditional state distributions given a complete action sequence need not agree: the policy weights a state path through the state-dependent Gaussian action likelihoods, whereas the target weights it through the exponentiated rewards. The bound in Eq.~\ref{eq:rl_DPI} is tight precisely when these two conditional state-trajectory distributions agree almost surely. The chain-rule result itself does not rely on Gaussianity; Eq.~\ref{eq:dpi_gaussian_policy} identifies the specialization used by the non-diffusion algorithms in this paper.
\endgroup

\subsubsection{Diffusion Trajectories and $\Delta_{\mathrm{diffusion}}$}
\label{app:dpi_diffusion_gap_derivation}

\begingroup
A diffusion policy augments each executed action sequence $a_{0:T}=a_{0:T}^{0}$ with the intermediate diffusion variables $a_{0:T}^{1:K}$. Apply Eq.~\ref{eq:general_kl_chain_rule} with
$X=(a_{0:T},s_{0:T+1})$ and $Z=a_{0:T}^{1:K}$. The resulting chain rule is
\begin{align}
&D_{\mathrm{KL}}\!\left(
q_\theta(a_{0:T}^{0:K},s_{0:T+1})
\,\middle\|\,
\pi_\theta(a_{0:T}^{0:K},s_{0:T+1})
\right)
\nonumber\\
&\quad=
D_{\mathrm{KL}}\!\left(
q_\theta(a_{0:T},s_{0:T+1})
\,\middle\|\,
\pi(a_{0:T},s_{0:T+1})
\right)
+\Delta_{\mathrm{diffusion}},
\label{eq:dpi_diffusion_conditional_decomposition}
\end{align}
where
{\small
\begin{equation}
\Delta_{\mathrm{diffusion}}
:=
\mathbb E_{(a_{0:T},s_{0:T+1})\sim
q_\theta(a_{0:T},s_{0:T+1})}\!\left[
D_{\mathrm{KL}}\!\left(
q_\theta(a_{0:T}^{1:K}\mid a_{0:T},s_{0:T+1})
\,\middle\|\,
\pi_\theta(a_{0:T}^{1:K}\mid a_{0:T},s_{0:T+1})
\right)
\right].
\label{eq:dpi_diffusion_gap_derivation}
\end{equation}
}
Substituting the Gaussian-policy state decomposition in Eq.~\ref{eq:dpi_gaussian_state_decomposition} gives
\begin{align}
&D_{\mathrm{KL}}\!\left(
q_\theta(a_{0:T}^{0:K},s_{0:T+1})
\,\middle\|\,
\pi_\theta(a_{0:T}^{0:K},s_{0:T+1})
\right)
\nonumber\\
&\quad=
D_{\mathrm{KL}}\!\left(
q_\theta(a_{0:T})\,\middle\|\,\pi(a_{0:T})
\right)
+\Delta_{\mathrm{state}}
+\Delta_{\mathrm{diffusion}}.
\label{eq:dpi_gap_exact_derivation}
\end{align}
The two nonnegative terms have distinct meanings. $\Delta_{\mathrm{state}}$ is already present for an ordinary Gaussian policy and measures the conditional state-trajectory mismatch after fixing the executed actions. $\Delta_{\mathrm{diffusion}}$ is introduced only by the diffusion augmentation and measures the additional mismatch between the conditional intermediate diffusion paths after fixing both the executed actions and state trajectory. The augmented DPI bound is tight if and only if both conditional distributions agree almost surely:
\begin{equation}
q_\theta(s_{0:T+1}\mid a_{0:T})
=\pi(s_{0:T+1}\mid a_{0:T}),
\qquad
q_\theta(a_{0:T}^{1:K}\mid a_{0:T},s_{0:T+1})
=\pi_\theta(a_{0:T}^{1:K}\mid a_{0:T},s_{0:T+1}).
\label{eq:dpi_gap_equality_derivation}
\end{equation}
\endgroup

\subsubsection{Finite-Prefix DA-MDP Construction}

We now instantiate the preceding diffusion chain rule for the finite-prefix DA-MDP distributions. Retaining the complete denoising path can only reveal more discrepancy between policy and target than observing the final actions alone.

The augmented policy and target distributions are
{\small
\begin{align}
q_{\mathrm{DA},\theta}^{T}(a_{0:T}^{0:K},s_{0:T+1})
&:=p(s_0)\prod_{t=0}^{T}\!\left[
p(s_{t+1}\mid s_t,a_t^0)q_{\mathrm{prior}}(a_t^K)
\prod_{k=1}^{K}q_\theta(a_t^{k-1}\mid a_t^k,s_t)
\right],
\label{eq:finite_da_policy_trajectory}\\
\pi_{\mathrm{DA},\theta}^{T}(a_{0:T}^{0:K},s_{0:T+1})
&:=\frac{p(s_0)}{\mathcal Z_T}\prod_{t=0}^{T}\!\left[
p(s_{t+1}\mid s_t,a_t^0)
e^{R_{\mathrm{env}}(s_t,a_t^0)/\Tau}
\prod_{k=1}^{K}\pi_\theta(a_t^k\mid a_t^{k-1},s_t)
\right].
\label{eq:finite_da_target_trajectory}
\end{align}
}
Here, $\mathcal Z_T$ is the partition function that normalizes the finite-horizon augmented target.
Every forward kernel is normalized, so marginalizing $a_{0:T}^{1:K}$ from the target recovers the joint state--executed-action target; marginalizing the same variables from the policy recovers its corresponding joint distribution. Applying the data-processing inequality first to $a_{0:T}^{1:K}$ and then to $s_{0:T+1}$ gives
{
\begin{equation}
  \begin{aligned}
  &
D_{\mathrm{KL}}\!\left(q_\theta(a_{0:T})\,\middle\|\,\pi(a_{0:T})\right) \leq
D_{\mathrm{KL}}\!\left(
q_\theta(a_{0:T},s_{0:T+1})
\,\middle\|\,
\pi(a_{0:T},s_{0:T+1})
\right)\\
  &\quad
\leq
D_{\mathrm{KL}}\!\left(
q_{\mathrm{DA},\theta}^{T}(a_{0:T}^{0:K},s_{0:T+1})
\,\middle\|\,
\pi_{\mathrm{DA},\theta}^{T}(a_{0:T}^{0:K},s_{0:T+1})
\right).
  \end{aligned}
  \label{eq:finite_da_two_stage_dpi}
\end{equation}
}
The augmented bound is zero exactly when
$q_{\mathrm{DA},\theta^\star}^{T}(a_{0:T}^{0:K},s_{0:T+1})
=\pi_{\mathrm{DA},\theta^\star}^{T}(a_{0:T}^{0:K},s_{0:T+1})$ almost everywhere. This condition is sufficient for equality of the executed-action marginals, whereas a nonzero augmented KL may still coexist with a good executed-action marginal because the conditional state and intermediate diffusion distributions may still differ.

\subsubsection{Infinite-Horizon Discounted Objective}

We now prove the stationary discounted counterpart of the finite-prefix construction. In simple terms, all denoising decisions used to produce one environment action belong to the same environment time and must therefore receive the same factor $\gamma^t$. As above, all trajectory distributions include $s_0\sim p(s_0)$.

\paragraph{Discounted augmented return.}
For compactness, define the local reverse--forward log ratio and augmented cost
\begin{align}
\ell_{t,k}^{\theta}
&:=\log\frac{q_\theta(a_t^{k-1}\mid a_t^k,s_t)}
{\pi_\theta(a_t^k\mid a_t^{k-1},s_t)},
&
c_{t,k}^{\theta}
&:=\Tau\ell_{t,k}^{\theta}
-\mathbf 1_{\{k=1\}}R_\mathrm{env}(s_t,a_t^0).
\label{eq:infinite_da_local_cost}
\end{align}
The corresponding deterministic time-weighted maximum-entropy return is
\begin{align}
\mathcal J_{\mathrm{DA}}(\theta)
&:=\mathbb E_{q_\theta}\!\left[
\sum_{t=0}^{\infty}\gamma^t
\left(
R_\mathrm{env}(s_t,a_t^0)
-\Tau\sum_{k=1}^{K}\ell_{t,k}^{\theta}
\right)
\right].
\label{eq:infinite_da_objective}
\end{align}
As in Alternative~A of App.~\ref{app:infinite_horizon_gaussian}, this objective can be obtained by assigning $\Tau_t=\gamma^t\Tau$ to every reverse--forward log ratio in environment step $t$ and assigning $R_t=\gamma^tR_\mathrm{env}$ to its terminal denoising phase. We assume the discounted trajectory costs and their parameter derivatives satisfy the integrability condition below. For the finite trajectory prefix $\xi_t=(s_{0:t+1},a_{0:t}^{0:K})$, define
\[
f_t(\theta,\xi_t)
:=q_{\mathrm{DA},\theta}^{t}(\xi_t)\sum_{k=1}^{K}c_{t,k}^{\theta},
\qquad
-\mathcal J_{\mathrm{DA}}(\theta)
=\sum_{t=0}^{\infty}\gamma^t\int f_t(\theta,\xi_t)\,d\xi_t.
\]
At each parameter value under consideration, assume that $f_t$ is continuously differentiable in an open neighborhood $U$, with a parameter-independent integration domain, and that
\[
\sum_{t=0}^{\infty}\gamma^t
\int\sup_{\vartheta\in U}
\left(
|f_t(\vartheta,\xi_t)|
+\|\nabla_{\vartheta}f_t(\vartheta,\xi_t)\|
\right)\,d\xi_t<\infty.
\]
This condition justifies interchanging differentiation, integration, and the discounted sum. The derivative of $f_t$ includes both the trajectory-density derivative and the explicit cost derivative.

\paragraph{A single KL over geometrically stopped DA-MDP trajectories.}
The same objective admits an exact single-KL representation. Let the environment-step terminal index $H$ be independent of the trajectory, with
\begin{align}
w(h):=\Pr(H=h)=(1-\gamma)\gamma^h,
\qquad \Pr(H\geq t)=\gamma^t.
\label{eq:infinite_da_geometric_horizon}
\end{align}
For $\xi_h=(s_{0:h+1},a_{0:h}^{0:K})$, define
\begin{align}
q_{\mathrm{DA},\theta}(h,\xi_h)
&:=w(h)p(s_0)
\prod_{t=0}^{h}\!\left[
q_\mathrm{prior}(a_t^K)
\prod_{k=1}^{K}q_\theta(a_t^{k-1}\mid a_t^k,s_t)
p(s_{t+1}\mid s_t,a_t^0)
\right],
\label{eq:infinite_da_q}\\
\widetilde\pi_{\mathrm{DA},\theta}(h,\xi_h)
&:=w(h)p(s_0)
\prod_{t=0}^{h}\!\left[
e^{R_\mathrm{env}(s_t,a_t^0)/\Tau}
\prod_{k=1}^{K}\pi_\theta(a_t^k\mid a_t^{k-1},s_t)
p(s_{t+1}\mid s_t,a_t^0)
\right],
\label{eq:infinite_da_target}\\
\mathcal Z_{\mathrm{DA}}
&:=\sum_{h=0}^{\infty}\int
\widetilde\pi_{\mathrm{DA},\theta}(h,\xi_h)\,d\xi_h,
\qquad
\pi_{\mathrm{DA},\theta}
:=\frac{\widetilde\pi_{\mathrm{DA},\theta}}
{\mathcal Z_{\mathrm{DA}}}.
\label{eq:infinite_da_partition}
\end{align}
{Because the forward kernels are normalized, integrating over $a_t^{1:K}$ leaves only an integral over the compact executed-action space. Consequently, the sufficient condition $\gamma V_{\mathcal A}e^{R_{\max}/\Tau}<1$ from Eq.~\ref{eq:geometric_normalizability_condition} also guarantees $0<\mathcal Z_{\mathrm{DA}}<\infty$. If it is not satisfied initially, the constant reward shift in Eq.~\ref{eq:geometric_reward_shift} enforces it and changes $\mathcal J_{\mathrm{DA}}$ only by the parameter-independent constant $-b/(1-\gamma)$.} Because each forward kernel is normalized, integrating Eq.~\ref{eq:infinite_da_target} over $a_t^{1:K}$ leaves only the reward potential over $a_t^0$; hence the partition function is independent of the forward-kernel parameters. The initial-state, geometric-horizon, and raw environment-transition factors in Eqs.~\ref{eq:infinite_da_q} and~\ref{eq:infinite_da_target} cancel inside the log ratio. Consequently,
\begin{align}
\log\frac{q_{\mathrm{DA},\theta}(h,\xi_h)}
{\pi_{\mathrm{DA},\theta}(h,\xi_h)}
&=\sum_{t=0}^{h}\!\left[
\log q_\mathrm{prior}(a_t^K)
+\sum_{k=1}^{K}\ell_{t,k}^{\theta}
-\frac{R_\mathrm{env}(s_t,a_t^0)}{\Tau}
\right]
+\log\mathcal Z_{\mathrm{DA}}.
\label{eq:infinite_da_log_ratio}
\end{align}
Using Eq.~\ref{eq:geom_discount_identity} gives the exact identity
\begin{align}
\Tau\Dkl\!\left(
q_{\mathrm{DA},\theta}(H,\xi_H)
\,\middle\|\,
\pi_{\mathrm{DA},\theta}(H,\xi_H)
\right)
&=-\mathcal J_{\mathrm{DA}}(\theta)
+C_{\mathrm{prior}}
+\Tau\log\mathcal Z_{\mathrm{DA}},
\label{eq:infinite_da_trajectory_kl}\\
C_{\mathrm{prior}}
&:=\frac{\Tau}{1-\gamma}
\mathbb E_{a^K\sim q_\mathrm{prior}}
[\log q_\mathrm{prior}(a^K)].
\label{eq:infinite_da_prior_constant}
\end{align}
The prior and normalizer terms are independent of the optimized reverse-policy parameters and, for normalized forward kernels, of the forward-kernel parameters. Minimizing the geometrically stopped augmented-trajectory KL is therefore equivalent, up to parameter-independent additive constants, to maximizing the infinite-horizon discounted DA-MDP objective in Eq.~\ref{eq:infinite_da_objective}.

\subsubsection{Bellman Equations, Occupancy, and Actor Surrogate}

We derive the stationary soft Bellman equations, normalize the discounted augmented-state visitation distribution, and express the policy gradient as a local actor-target KL. In simple terms, these identities show how the global path objective becomes the critic and actor losses evaluated on individual denoising transitions.

\paragraph{Stationary augmented Bellman equations.}
An exact stationary recursion is obtained with the phase-dependent discount
\begin{align}
\bar\gamma(k):=
\begin{cases}
1, & k>1,\\
\gamma, & k=1.
\end{cases}
\label{eq:infinite_da_phase_discount}
\end{align}
Thus, there is no discount between denoising decisions within one chain, and the factor $\gamma$ is applied exactly when $a_t^0$ is executed and the next environment step begins. Let $\tilde s=(s,a^k,k)$, $\tilde a=a^{k-1}$, and let $p_\mathrm{DA}$ denote the augmented transition kernel in Sec.~\ref{sec:MDP}. For a stationary reverse policy $q_\theta$, the soft evaluation equations are
\begin{align}
Q_{\mathrm{DA}}^{q_\theta}(\tilde s,a^{k-1})
&=\mathbf 1_{\{k=1\}}R_\mathrm{env}(s,a^0)
+\bar\gamma(k)\,
\mathbb E_{\tilde s'\sim p_\mathrm{DA}(\cdot\mid\tilde s,a^{k-1})}
[V_{\mathrm{DA}}^{q_\theta}(\tilde s')],
\label{eq:infinite_da_q_bellman}\\
V_{\mathrm{DA}}^{q_\theta}(\tilde s)
&=\mathbb E_{a^{k-1}\sim q_\theta(\cdot\mid\tilde s)}\!\left[
Q_{\mathrm{DA}}^{q_\theta}(\tilde s,a^{k-1})
-\Tau\log\frac{q_\theta(a^{k-1}\mid a^k,s)}
{\pi_\theta(a^k\mid a^{k-1},s)}
\right].
\label{eq:infinite_da_v_bellman}
\end{align}
Starting Eq.~\ref{eq:infinite_da_v_bellman} at $(s_0,a_0^K,K)$ and averaging $s_0\sim p(s_0)$ and $a_0^K\sim q_\mathrm{prior}$ recovers $\mathcal J_{\mathrm{DA}}$ exactly. The prior-density term occurs only in the full trajectory KL and is isolated as the parameter-independent constant in Eq.~\ref{eq:infinite_da_prior_constant}. In particular, the stationary augmented state retains the diffusion phase $k$, but it needs neither an environment-time index nor a finite-horizon countdown.

\paragraph{Discounted occupancy and actor gradient.}
{
Let $\rho_{t,k}^{q_\theta}(s,a^k)$ denote the joint probability
density of $(s_t,a_t^k)$ at denoising phase $k$. The normalized
discounted visitation density at $\tilde s=(s,a^k,k)$ is
$d_{\mathrm{DA}}^{q_\theta}(s,a^k,k)
:=\frac{1-\gamma}{K}\sum_{t=0}^{\infty}\gamma^t
\rho_{t,k}^{q_\theta}(s,a^k)$.
Each $\rho_{t,k}^{q_\theta}(s,a^k)$ integrates to one over the
continuous state and noisy-action variables. Consequently,
\begin{equation}
\begin{aligned}
\sum_{k=1}^K\int\!\int
d_{\mathrm{DA}}^{q_\theta}(s,a^k,k)\,da^k\,ds
&=\frac{1-\gamma}{K}\sum_{t=0}^{\infty}\gamma^t
  \sum_{k=1}^K\underbrace{\int\!\int
  \rho_{t,k}^{q_\theta}(s,a^k)\,da^k\,ds}_{1}\\
&=(1-\gamma)\sum_{t=0}^{\infty}\gamma^t=1.
\end{aligned}
\label{eq:infinite_da_occupancy_normalization}
\end{equation}
}
Moreover, for any integrable function $g$ of the augmented state, substituting the definition of $d_{\mathrm{DA}}^{q_\theta}$ gives
\begin{align}
\sum_{t=0}^{\infty}\gamma^t\sum_{k=1}^{K}
\mathbb E_{q_\theta}[g(\tilde s_{t,k})]
=\frac{K}{1-\gamma}
\mathbb E_{\tilde s\sim d_{\mathrm{DA}}^{q_\theta}}[g(\tilde s)].
\label{eq:infinite_da_occupancy_identity}
\end{align}
Applying this identity to the local policy-gradient integrand, the causal likelihood-ratio argument for a fixed forward kernel gives
{\small
\begin{align}
-\nabla_\theta\mathcal J_{\mathrm{DA}}(\theta)
&=\sum_{t=0}^{\infty}\gamma^t\sum_{k=1}^{K}
\mathbb E_{q_\theta}\!\left[
\left(
\Tau\ell_{t,k}^{\theta}
-Q_{\mathrm{DA}}^{q_\theta}(\tilde s_{t,k},a_t^{k-1})
\right)
\nabla_\theta\log q_\theta(a_t^{k-1}\mid\tilde s_{t,k})
\right]
\nonumber\\
&=\frac{K}{1-\gamma}
\mathbb E_{\substack{\tilde s\sim d_{\mathrm{DA}}^{q_\theta}\\
a^{k-1}\sim q_\theta(\cdot\mid\tilde s)}}\!\left[
\left(
\Tau\ell_\theta(\tilde s,a^{k-1})
-Q_{\mathrm{DA}}^{q_\theta}(\tilde s,a^{k-1})
\right)
\nabla_\theta\log q_\theta(a^{k-1}\mid\tilde s)
\right],
\label{eq:infinite_da_policy_gradient}
\end{align}
}
where $\tilde s=(s',a^k,k)$ and
$\ell_\theta(\tilde s,a^{k-1})
:=\log\!\frac{q_\theta(a^{k-1}\mid a^k,s')}
{\pi(a^k\mid a^{k-1},s')}$. 

At frozen rollout parameters $\thetar$ and for a fixed forward kernel, define the stationary local target
\begin{align}
\pi_{Q,\mathrm{DA}}^{q_{\thetar}}(a^{k-1}\mid\tilde s)
&:=\frac{
\pi_{\thetar}(a^k\mid a^{k-1},{s'})
\exp\!\left(Q_{\mathrm{DA}}^{q_{\thetar}}(\tilde s,a^{k-1})/\Tau\right)}
{Z_{Q,\mathrm{DA}}^{q_{\thetar}}(\tilde s)}.
\label{eq:infinite_da_local_target}
\end{align}
The time-independent actor surrogate is then
\begin{align}
{\mathcal L_{\mathrm{DA}}^{\mathrm{fixed}}(\theta;\thetar)}
&:=\frac{K\Tau}{1-\gamma}
\mathbb E_{\tilde s\sim d_{\mathrm{DA}}^{q_{\thetar}}}\!\left[
\Dkl\!\left(
q_\theta(a^{k-1}\mid\tilde s)
\,\middle\|\,
\pi_{Q,\mathrm{DA}}^{q_{\thetar}}(a^{k-1}\mid\tilde s)
\right)
\right].
\label{eq:infinite_da_actor_surrogate}
\end{align}
At $\theta=\thetar$, its gradient equals Eq.~\ref{eq:infinite_da_policy_gradient}. Thus, the local KL is first-order exact at the rollout parameters even though its occupancy and critic remain frozen during subsequent minibatch updates.

{Here the forward kernel and critic are fixed, so normalizing the local target does not change the actor gradient. The next subsection considers learned forward kernels and uses the unnormalized objective to avoid introducing an additional normalizer gradient that is absent from the trajectory objective.}

\subsubsection{Learned Forward-Kernel Correction}
\label{app:learned_forward_kernel_correction}

We prove that learning the forward diffusion coefficients adds one explicit score term and that the unnormalized local target retains this term without introducing a spurious normalizer gradient. In simple terms, the forward kernel appears inside the reward correction, so it must receive a direct gradient even though actions are sampled from the reverse kernel.

When $\pi_\theta$ is learned, differentiating the local reverse--forward log ratio adds
\begin{equation}
-\Tau\sum_{t=0}^{\infty}\gamma^t\sum_{k=1}^{K}
\mathbb E_{q_\theta}\!\left[
\nabla_\theta\log\pi_\theta(a_t^k\mid a_t^{k-1},s_t)
\right]
\label{eq:infinite_da_learned_forward_gradient}
\end{equation}
to the fixed-forward policy gradient in Eq.~\ref{eq:infinite_da_policy_gradient}. The derivatives of the policy-induced state visitation are already summarized by $Q_{\mathrm{DA}}^{q_\theta}$; Eq.~\ref{eq:infinite_da_learned_forward_gradient} is the only additional explicit derivative.

At frozen rollout parameters, the corresponding gradient-matching surrogate is
\begin{align}
{\mathcal L_{\mathrm{DA}}(\theta;\thetar)}
&:=\frac{K}{1-\gamma}
\mathbb E_{\substack{\tilde s\sim d_{\mathrm{DA}}^{q_{\thetar}}\\
a^{k-1}\sim q_\theta(\cdot\mid\tilde s)}}\!\left[
\Tau\log\frac{q_\theta(a^{k-1}\mid a^k,s')}
{\pi_\theta(a^k\mid a^{k-1},s')}
-Q_{\mathrm{DA}}^{q_{\thetar}}(\tilde s,a^{k-1})
\right].
\label{eq:infinite_da_learned_forward_surrogate}
\end{align}
To see the unnormalized-KL form explicitly, define the local target over $a^{k-1}$ by
\begin{equation}
\widetilde\pi_{Q,\theta}^{q_{\thetar}}(a^{k-1}\mid\tilde s)
:=\pi_\theta(a^k\mid a^{k-1},s')
\exp\!\left(
\frac{Q_{\mathrm{DA}}^{q_{\thetar}}(\tilde s,a^{k-1})}{\Tau}
\right).
\label{eq:infinite_da_unnormalized_local_target}
\end{equation}
This target is generally not normalized as a distribution over $a^{k-1}$. For a normalized reverse kernel $q_\theta(\cdot\mid\tilde s)$, define
{
\begin{equation}
  D_{\mathrm{KL}}^{\mathrm u}\!\left(
q_\theta(\cdot\mid\tilde s)
\,\middle\|\,
\widetilde\pi_{Q,\theta}^{q_{\thetar}}(\cdot\mid\tilde s)
\right) :=\int q_\theta(a^{k-1}\mid\tilde s)
\log\frac{q_\theta(a^{k-1}\mid\tilde s)}
{\widetilde\pi_{Q,\theta}^{q_{\thetar}}(a^{k-1}\mid\tilde s)}
\,d a^{k-1}.
  \label{eq:infinite_da_unnormalized_kl_definition}
\end{equation}
}
Substituting Eq.~\ref{eq:infinite_da_unnormalized_local_target} into this definition gives
\begin{align}
&\Tau D_{\mathrm{KL}}^{\mathrm u}\!\left(
q_\theta(\cdot\mid\tilde s)
\,\middle\|\,
\widetilde\pi_{Q,\theta}^{q_{\thetar}}(\cdot\mid\tilde s)
\right)
\nonumber\\
&\quad=\mathbb E_{a^{k-1}\sim q_\theta(\cdot\mid\tilde s)}\!\left[
\Tau\log\frac{q_\theta(a^{k-1}\mid a^k,s')}
{\pi_\theta(a^k\mid a^{k-1},s')}
-Q_{\mathrm{DA}}^{q_{\thetar}}(\tilde s,a^{k-1})
\right].
\label{eq:infinite_da_unnormalized_kl_expansion}
\end{align}
Hence Eq.~\ref{eq:infinite_da_learned_forward_surrogate} can be written exactly as
\begin{equation}
{\mathcal L_{\mathrm{DA}}(\theta;\thetar)}
=\frac{K\Tau}{1-\gamma}
\mathbb E_{\tilde s\sim d_{\mathrm{DA}}^{q_{\thetar}}}\!\left[
D_{\mathrm{KL}}^{\mathrm u}\!\left(
q_\theta(\cdot\mid\tilde s)
\,\middle\|\,
\widetilde\pi_{Q,\theta}^{q_{\thetar}}(\cdot\mid\tilde s)
\right)
\right].
\label{eq:infinite_da_learned_forward_unnormalized_kl}
\end{equation}
Finally, let $Z_{Q,\theta}(\tilde s):=\int\widetilde\pi_{Q,\theta}^{q_{\thetar}}(a^{k-1}\mid\tilde s)\,d a^{k-1}$ and $p_{Q,\theta}(a^{k-1}\mid\tilde s):=\widetilde\pi_{Q,\theta}^{q_{\thetar}}(a^{k-1}\mid\tilde s)/Z_{Q,\theta}(\tilde s)$. Then
\begin{equation}
\begin{aligned}
&D_{\mathrm{KL}}\!\left(
q_\theta(a^{k-1}\mid\tilde s)
\,\middle\|\,
p_{Q,\theta}(a^{k-1}\mid\tilde s)
\right)\\
&\quad=D_{\mathrm{KL}}^{\mathrm u}\!\left(
q_\theta(a^{k-1}\mid\tilde s)
\,\middle\|\,
\widetilde\pi_{Q,\theta}^{q_{\thetar}}(a^{k-1}\mid\tilde s)
\right)
+\log Z_{Q,\theta}(\tilde s).
\end{aligned}
\label{eq:infinite_da_normalized_vs_unnormalized_kl}
\end{equation}
Because the learned forward kernel makes $Z_{Q,\theta}$ parameter dependent, differentiating the normalized KL would add $\Tau\nabla_\theta\log Z_{Q,\theta}(\tilde s)$, which is absent from the augmented-trajectory gradient. The unnormalized KL retains the required explicit forward-kernel gradient without introducing this additional term.
\subsection{{Monotonic Improvement for the Local DA-MDP Loss}}
\label{app:da_monotonic_improvement}

\begingroup

\newcommand{\daproofstep}[3]{\subsubsection*{{Step~#1: #2}}\label{#3}}
\newcommand{\daproofsubstep}[3]{\paragraph{{Substep~#1: #2.}}\label{#3}}

\noindent\textbf{Statement to be proved.}
{We prove Proposition~\ref{prop:da_return_improvement}
by relating the change in the augmented return of}
Eq.~\ref{eq:kl_decomp_fixed} directly to the change in the local
denoising loss of Eq.~\ref{eq:diffusion_surrogate}. Let $\theta$
be the candidate parameters and $\thetar$ the frozen rollout
parameters, and write $q_{\mathrm{roll}}:=q_{\thetar}$ and
$\pi_{\mathrm{roll}}:=\pi_{\thetar}$. Define
\begin{equation}
\begin{aligned}
\Delta\mathcal J_{\mathrm{DA}}
&:=\mathcal J_{\mathrm{DA}}(\theta)-\mathcal J_{\mathrm{DA}}(\thetar),\\
\Delta\mathcal L_{\mathrm{DA}}
&:=\mathcal L_{\mathrm{DA}}(\theta;\thetar)
  -\mathcal L_{\mathrm{DA}}(\thetar;\thetar),\\
\Delta\mathcal J_{\mathrm{DA}}^{\mathrm{surrogate}}
&:=-\Delta\mathcal L_{\mathrm{DA}}.
\end{aligned}
\label{eq:da_local_improvement_definitions}
\end{equation}
The return and loss differences are candidate minus rollout.
A positive surrogate return gain
$\Delta\mathcal J_{\mathrm{DA}}^{\mathrm{surrogate}}$ corresponds
to a decrease in the local loss and predicts an increase in the true return.
The second argument of the loss keeps its critic and sampling
distribution fixed at the rollout parameters.

We use the exact local rollout value functions from
Eq.~\ref{eq:diffmaxentvaluefunction}, abbreviated as
$Q_{\mathrm{roll}}:=Q_{\mathrm{DA}}^{q_{\mathrm{roll}}}$ and
$V_{\mathrm{roll}}:=V_{\mathrm{DA}}^{q_{\mathrm{roll}}}$.
For $\tilde s=(s,a^k,k)$, define the reverse--forward log ratio and
the local soft advantage by
\begin{align}
\ell_\theta(\tilde s,a^{k-1})
&:=\log\frac{q_\theta(a^{k-1}\mid a^k,s)}
                 {\pi_\theta(a^k\mid a^{k-1},s)},
\label{eq:da_local_log_ratio}\\
A_{\mathrm{roll}}(\tilde s,a^{k-1})
&:=Q_{\mathrm{roll}}(\tilde s,a^{k-1})
 -\Tau\ell_{\mathrm{roll}}(\tilde s,a^{k-1})-V_{\mathrm{roll}}(\tilde s).
\label{eq:da_local_rollout_advantage}
\end{align}
The rollout expectation of this advantage is zero at every
augmented state. The local policy-change KL and expected
forward-log-density change are
\begin{align}
\kappa(\tilde s)
&:=\Dkl\bigl(q_\theta(\cdot\mid\tilde s)\|q_{\mathrm{roll}}(\cdot\mid\tilde s)\bigr),
\label{eq:da_local_policy_change_kl}\\
\bar G_\theta(\tilde s)
&:=\mathbb E_{a^{k-1}\sim q_\theta(\cdot\mid\tilde s)}
 \left[\log\frac{\pi_\theta(a^k\mid a^{k-1},s)}
                    {\pi_{\mathrm{roll}}(a^k\mid a^{k-1},s)}\right].
\label{eq:da_local_forward_log_change}
\end{align}
Define the constants
\begin{equation}
\begin{aligned}
\epsilon_A&:=\sup_{\tilde s,a^{k-1}}|A_{\mathrm{roll}}(\tilde s,a^{k-1})|,
\qquad D_q:=\sup_{\tilde s}\kappa(\tilde s)
 =D_{\mathrm{KL}}^{\max}(q_\theta\|q_{\mathrm{roll}}),\\
\epsilon_G&:=\sup_{\tilde s}|\bar G_\theta(\tilde s)|,
\qquad
B_{K,\gamma}:=\frac{K}{1-\gamma}
 \left(K-1+\frac{2\gamma K}{1-\gamma}\right).
\end{aligned}
\label{eq:da_bound_constants}
\end{equation}
Here $D_q$ is the largest candidate-to-rollout KL across augmented
states, $\epsilon_A$ bounds the rollout advantage, and $\epsilon_G$
bounds the expected change in forward log density. Under the
assumptions below, we prove the KL-based bound
\begin{equation*}
\boxed{\begin{aligned}
\Delta\mathcal J_{\mathrm{DA}}
 &\geq\Delta\mathcal J_{\mathrm{DA}}^{\mathrm{surrogate}}
 -\left({\epsilon_A B_{K,\gamma}}+\frac{K\Tau}{1-\gamma}\right)D_q
 \\&\quad-\Tau B_{K,\gamma}\epsilon_G{\sqrt{D_q/2}}.
\end{aligned}}
\tag{\ref{eq:da_learned_forward_monotonic_bound}}
\end{equation*}
Consequently, the return strictly improves when the surrogate gain
$\Delta\mathcal J_{\mathrm{DA}}^{\mathrm{surrogate}}$ exceeds the displayed penalties.
For fixed forward kernels, $\bar G_\theta=0$ and $\epsilon_G=0$,
leaving a penalty proportional to the reverse-policy KL alone.
\subsubsection{Relating the Bound to Kernel Constraints}
\label{app:da_kernel_constraints}
\paragraph{Direction and averaging of the reverse KL.}
The final bound uses the candidate-to-rollout direction because
the entropy correction $\kappa$ averages the reverse log ratio
under $q_\theta$. Total variation can be bounded using either KL
direction, but the opposite-direction KL alone does not bound
$\kappa$. The theorem uses the supremum over augmented states;
replacing it by a rollout-minibatch average is a practical
relaxation of this worst-state constraint.

\paragraph{Learned forward kernels.}
\label{app:direct_forward_local_constraint}
{Learning the forward coefficients affects both the
surrogate gain and its prediction error. The term $+\Tau\bar G_\theta$
in the surrogate measures the expected increase in forward log density
on candidate reverse samples: increasing $\pi_\theta$ relative to
$\pi_{\mathrm{roll}}$ at these samples raises this contribution,
whereas decreasing it lowers it. The final bound additionally subtracts
$\Tau B_{K,\gamma}\epsilon_G\sqrt{D_q/2}$ to account for evaluating
this change on rollout rather than candidate augmented states.
Since $\epsilon_G=\sup_{\tilde s}|\bar G_\theta(\tilde s)|$, this
penalty is nonnegative regardless of the sign of $\bar G_\theta$.
It vanishes when the forward kernel is fixed or $D_q=0$; in the
latter case, a forward-induced return change is still captured by
the surrogate.}

{Our algorithms encourage small reverse-policy updates
through likelihood-ratio clipping in \dappo{} and reverse-kernel KL
control in \dareppo{} and \dawpo{}. These mechanisms indirectly
target the visitation change responsible for the additional error:
for a fixed $\epsilon_G$, reducing $D_q$ reduces
$\Tau B_{K,\gamma}\epsilon_G\sqrt{D_q/2}$.
The empirical rollout-to-current KL used by the latter methods
(Eq.~\ref{eq:da_reppo_chain_kl}) is a practical proxy.
Thus, the implementation encourages the small-update regime
motivated by the bound without certifying its improvement condition.}

{A complementary direction for future work is to
control $\epsilon_G$ more directly. A sufficient forward-update
constraint is}
\begin{equation}
\epsilon_G\leq\sup_{\tilde s}
 \mathbb E_{a^{k-1}\sim q_\theta(\cdot\mid\tilde s)}
 \left|\log\frac{\pi_\theta(a^k\mid a^{k-1},s)}
                   {\pi_{\mathrm{roll}}(a^k\mid a^{k-1},s)}\right|
 {\leq\delta_\pi}.
\label{eq:da_direct_forward_condition}
\end{equation}
{This bounds the additional penalty by
$\Tau B_{K,\gamma}\delta_\pi\sqrt{D_q/2}$. Estimates on rollout
augmented states with fresh candidate reverse samples could guide
a dedicated forward trust region, although finite-sample checks
do not certify the uniform constraint.}

\subsubsection{{Proof}}
\label{app:da_monotonic_improvement_proof}

\paragraph{Assumptions and notation.}
We consider continuous environment states and latent actions, with
initial-state and environment-transition densities. The candidate
and rollout policies share the same initial-state distribution,
environment dynamics, number $K\geq1$ of diffusion steps, and fixed
prior $q_{\mathrm{prior}}$. The temperature $\Tau>0$ and discount
$0\leq\gamma<1$ are fixed. Assume common support for the reverse and
forward densities, bounded environment rewards, and absolute
integrability of the discounted costs under both policies and of
the candidate-to-rollout reverse and forward log ratios appearing
below. In addition, require
\[
\sup_{\tilde s}|V_{\mathrm{roll}}(\tilde s)|<\infty,
\qquad \epsilon_A<\infty,\quad D_q<\infty,\quad\epsilon_G<\infty,
\]
with the constants defined in Eq.~\ref{eq:da_bound_constants}.
These assumptions justify the expectations below; bounded rollout
values make the terminal telescoping term vanish. These uniform
boundedness assumptions need not hold for Gaussian kernels on
unbounded latent spaces; the proposition provides a conditional
improvement guarantee.
For this theoretical bound, we assume access to the exact value
functions of the frozen rollout policy.
The normalized augmented visitation density $d_{\mathrm{DA}}^{q_\theta}$ is defined
in App.~\ref{app:infinite_horizon_da_mdp}. Its $L^1$ norm integrates the
continuous variables and sums over the discrete diffusion phase.

\paragraph{Proof roadmap.}
The proof follows the maximum-entropy TRPO argument: express both
gains through the same advantage and entropy correction, bound the
advantage error using state visitation, then bound the entropy
correction using $0\leq\kappa\leq D_q$. Diffusion adds intermediate
noisy-action visitation and, when the forward kernels are learned,
the $\bar G_\theta$ term.
The proof first rewrites the two changes exactly, then bounds their
difference.
\begin{enumerate}
\item \textbf{Expand the local-loss change.}
  \hyperref[app:da_proof_advantage_identity]{Substep~1.3} identifies
  the advantage, entropy correction, and forward correction.
  \hyperref[app:da_proof_rollout_average]{Substep~1.4} averages
  them over rollout augmented states.
\item \textbf{Expand the return change.}
  \hyperref[app:da_proof_telescope]{Substep~2.2} telescopes the
  rollout values across both diffusion and environment steps.
  \hyperref[app:da_proof_return_average]{Substep~2.3} obtains the
  candidate augmented-state average of the same terms.
\item \textbf{Bound the advantage and forward-correction errors.}
  \hyperref[app:da_proof_advantage_size]{Substep~3.1} bounds the
  expected rollout advantage. Substeps~3.2--3.5 control the change
  in augmented states, and
  \hyperref[app:da_proof_advantage_combine]{Substep~3.6} converts
  their product into a KL bound.
\item \textbf{Complete the improvement bound.}
  \hyperref[app:da_proof_entropy_error]{Substep~4.1} bounds the
  remaining entropy correction, and
  \hyperref[app:da_proof_combine]{Substep~4.2} combines the bounds.
  Substeps~4.3--4.4 give the fixed-forward case and the sufficient
  condition for return improvement.
\end{enumerate}

\daproofstep{1}{Express the local-loss improvement}{app:da_proof_step1}
We show that the negative loss change is the rollout-state average
of the expected advantage, minus a policy-change KL, plus the
forward correction. This step uses only the loss definition and
the rollout Bellman equation.

\daproofsubstep{1.1}{Restate the loss and evaluate its rollout value}{app:da_proof_surrogate_definition}
Expanding the unnormalized KL in Eq.~\ref{eq:diffusion_surrogate}
gives precisely Eq.~\ref{eq:infinite_da_learned_forward_surrogate}:
\begin{equation}
\mathcal L_{\mathrm{DA}}(\theta;\thetar)
 =\frac{K}{1-\gamma}
 \mathbb E_{\substack{\tilde s\sim d_{\mathrm{DA}}^{q_{\mathrm{roll}}}\\
                     a^{k-1}\sim q_\theta(\cdot\mid\tilde s)}}
 [\Tau\ell_\theta(\tilde s,a^{k-1})-Q_{\mathrm{roll}}(\tilde s,a^{k-1})].
\label{eq:da_local_surrogate_restatement}
\end{equation}
At the rollout parameters, the inner expectation is
$-V_{\mathrm{roll}}(\tilde s)$, because
\begin{equation}
V_{\mathrm{roll}}(\tilde s)
 =\mathbb E_{a^{k-1}\sim q_{\mathrm{roll}}(\cdot\mid\tilde s)}
 [Q_{\mathrm{roll}}(\tilde s,a^{k-1})
  -\Tau\ell_{\mathrm{roll}}(\tilde s,a^{k-1})].
\label{eq:da_local_rollout_bellman}
\end{equation}
Thus $\mathcal L_{\mathrm{DA}}(\thetar;\thetar)
=-\frac{K}{1-\gamma}\mathbb E_{d_{\mathrm{DA}}^{q_{\mathrm{roll}}}}
[V_{\mathrm{roll}}]$.

\daproofsubstep{1.2}{Subtract the two losses}{app:da_proof_surrogate_expand}
Both losses use the same rollout distribution and critic. We can
therefore subtract their inner expectations at each augmented state.
For this calculation, abbreviate the expected local improvement as
\begin{equation}
g_\theta(\tilde s)
 :=\mathbb E_{a^{k-1}\sim q_\theta(\cdot\mid\tilde s)}
 [Q_{\mathrm{roll}}(\tilde s,a^{k-1})
  -\Tau\ell_\theta(\tilde s,a^{k-1})]-V_{\mathrm{roll}}(\tilde s).
\label{eq:da_local_gain}
\end{equation}
The loss difference is
\begin{equation*}
\begin{aligned}
\Delta\mathcal L_{\mathrm{DA}}
 &=\mathcal L_{\mathrm{DA}}(\theta;\thetar)
   -\mathcal L_{\mathrm{DA}}(\thetar;\thetar)\\
 &=\frac{K}{1-\gamma}
   \mathbb E_{\tilde s\sim d_{\mathrm{DA}}^{q_{\mathrm{roll}}}}\!\left[
   \mathbb E_{a^{k-1}\sim q_\theta}
   [\Tau\ell_\theta-Q_{\mathrm{roll}}]+V_{\mathrm{roll}}(\tilde s)\right]\\
 &=-\frac{K}{1-\gamma}
   \mathbb E_{\tilde s\sim d_{\mathrm{DA}}^{q_{\mathrm{roll}}}}
   [g_\theta(\tilde s)].
\end{aligned}
\end{equation*}
The inner expectation is conditional on $\tilde s$; its arguments
are the same as in Eq.~\ref{eq:da_local_surrogate_restatement}.

\daproofsubstep{1.3}{Identify the advantage and the two log-density corrections}{app:da_proof_advantage_identity}
We express $g_\theta$ through the rollout advantage and explicit
changes in the reverse and forward kernels. Recall the rollout
soft-advantage definition from Eq.~\ref{eq:da_local_rollout_advantage},
$A_{\mathrm{roll}}(\tilde s,a^{k-1})
 :=Q_{\mathrm{roll}}(\tilde s,a^{k-1})
 -\Tau\ell_{\mathrm{roll}}(\tilde s,a^{k-1})-V_{\mathrm{roll}}(\tilde s)$.
The first equality below simply substitutes this definition; the
second expands the difference of log ratios:
\begin{equation*}
\begin{aligned}
&Q_{\mathrm{roll}}(\tilde s,a^{k-1})
 -\Tau\ell_\theta(\tilde s,a^{k-1})-V_{\mathrm{roll}}(\tilde s)\\
&\quad=A_{\mathrm{roll}}(\tilde s,a^{k-1})
 -\Tau\bigl[\ell_\theta(\tilde s,a^{k-1})
             -\ell_{\mathrm{roll}}(\tilde s,a^{k-1})\bigr]\\
&\quad=A_{\mathrm{roll}}(\tilde s,a^{k-1})
 -\Tau\log\frac{q_\theta(a^{k-1}\mid a^k,s)}
                   {q_{\mathrm{roll}}(a^{k-1}\mid a^k,s)}
 +\Tau\log\frac{\pi_\theta(a^k\mid a^{k-1},s)}
                   {\pi_{\mathrm{roll}}(a^k\mid a^{k-1},s)}.
\end{aligned}
\end{equation*}
Define $\bar A_\theta(\tilde s)
:=\mathbb E_{a^{k-1}\sim q_\theta(\cdot\mid\tilde s)}
[A_{\mathrm{roll}}(\tilde s,a^{k-1})]$. Taking the same candidate
expectation of the preceding identity gives
\begin{equation}
g_\theta(\tilde s)
 =\bar A_\theta(\tilde s)-\Tau\kappa(\tilde s)
  +\Tau\bar G_\theta(\tilde s).
\label{eq:da_local_gain_components}
\end{equation}
The reverse log ratio has expectation $\kappa$ because the action
is sampled from the candidate kernel. This determines the
candidate-to-rollout direction of the KL. The forward log ratio
instead gives $\bar G_\theta$ as defined in
Eq.~\ref{eq:da_local_forward_log_change}.

\daproofsubstep{1.4}{Average over rollout augmented states}{app:da_proof_rollout_average}
Insert Eq.~\ref{eq:da_local_gain_components} into the loss
difference from Substep~1.2. The predicted return gain is
\begin{equation}
\begin{aligned}
\Delta\mathcal J_{\mathrm{DA}}^{\mathrm{surrogate}}
 =-\Delta\mathcal L_{\mathrm{DA}}
 &=\frac{K}{1-\gamma}
   \mathbb E_{\tilde s\sim d_{\mathrm{DA}}^{q_{\mathrm{roll}}}}
   [g_\theta(\tilde s)]\\
 &=\frac{K}{1-\gamma}
   \mathbb E_{\tilde s\sim d_{\mathrm{DA}}^{q_{\mathrm{roll}}}}
   [\bar A_\theta(\tilde s)-\Tau\kappa(\tilde s)+\Tau\bar G_\theta(\tilde s)].
\end{aligned}
\label{eq:da_learned_forward_surrogate_gain}
\end{equation}

\daproofstep{2}{Express the return improvement and compare the two gains}{app:da_proof_step2}
We now obtain the candidate-state average of the same local gain.
We start from the return itself and insert a telescoping sum of
rollout values. No distribution approximation is made in this step.

\daproofsubstep{2.1}{Restate the return and its initial-value baseline}{app:da_proof_return_difference}
At $\tilde s=(s,a^k,k)$, the regularized one-step reward is
\begin{equation}
r_\theta(\tilde s,a^{k-1})
 :=\mathbf 1_{\{k=1\}}R_{\mathrm{env}}(s,a^0)
   -\Tau\ell_\theta(\tilde s,a^{k-1}).
\label{eq:da_local_regularized_reward}
\end{equation}
Only the final denoising step receives environment reward; each
step receives its reverse--forward log-ratio contribution.
Writing $\tilde s_{t,k}=(s_t,a_t^k,k)$, the return of
Eq.~\ref{eq:infinite_da_objective} is
\begin{equation}
\mathcal J_{\mathrm{DA}}(\theta)
 =\mathbb E_\theta\!\left[
  \sum_{t=0}^\infty\gamma^t\sum_{k=1}^K
  r_\theta(\tilde s_{t,k},a_t^{k-1})\right].
\label{eq:da_regularized_return}
\end{equation}
Here and throughout this proof, $\mathbb E_\theta$ averages over
augmented trajectories generated by $s_0\sim p(s_0)$ and, at each
environment step $t\geq0$, $a_t^K\sim q_{\mathrm{prior}}$,
$a_t^{k-1}\sim q_\theta(\cdot\mid a_t^k,s_t)$ for $k=K,\ldots,1$,
and $s_{t+1}\sim p(\cdot\mid s_t,a_t^0)$. By definition,
$V_{\mathrm{roll}}(\tilde s_{0,K})$ is the expected discounted sum
of $r_{\thetar}$ under the rollout policy, conditional on the
initial augmented state $\tilde s_{0,K}$. Averaging over this state
therefore gives $\mathcal J_{\mathrm{DA}}(\thetar)
=\mathbb E_{s_0\sim p(s_0),\,a_0^K\sim q_{\mathrm{prior}}}
[V_{\mathrm{roll}}(\tilde s_{0,K})]$.
Since both policies share the initial-state distribution and prior,
this expectation is unchanged under the candidate policy. Hence,
\begin{equation}
\begin{aligned}
\Delta\mathcal J_{\mathrm{DA}}
 &=\mathcal J_{\mathrm{DA}}(\theta)
   -\mathcal J_{\mathrm{DA}}(\thetar)\\
 &=\mathbb E_\theta\!\left[
  \sum_{t=0}^\infty\gamma^t\sum_{k=1}^K
  r_\theta(\tilde s_{t,k},a_t^{k-1})
  -V_{\mathrm{roll}}(\tilde s_{0,K})\right].
\end{aligned}
\label{eq:da_return_difference_start}
\end{equation}

\daproofsubstep{2.2}{Telescope the rollout values across both time scales}{app:da_proof_telescope}
Let $\tilde s^+_{t,k}$ be the successor augmented state. For $k>1$
it is $\tilde s_{t,k-1}$; for $k=1$ it is the next environment
state with fresh prior noise, $\tilde s_{t+1,K}$. Recall that
$\bar\gamma(k)=1$ for $k>1$ and $\bar\gamma(1)=\gamma$.
To display the cancellation, write
$V_{t,k}:=V_{\mathrm{roll}}(\tilde s_{t,k})$. Within one chain,
\begin{equation*}
\begin{aligned}
&\sum_{k=1}^K
 [\bar\gamma(k)V_{\mathrm{roll}}(\tilde s^+_{t,k})-V_{t,k}]\\
&\quad=\sum_{k=2}^K(V_{t,k-1}-V_{t,k})
       +\gamma V_{t+1,K}-V_{t,1}
 =\gamma V_{t+1,K}-V_{t,K}.
\end{aligned}
\end{equation*}
Every intermediate noisy-action value cancels. Summing the remaining
terms over environment steps gives, for finite $N$,
\begin{equation*}
\begin{aligned}
\sum_{t=0}^N\gamma^t(\gamma V_{t+1,K}-V_{t,K})
 &=\sum_{t=1}^{N+1}\gamma^t V_{t,K}
   -\sum_{t=0}^N\gamma^t V_{t,K}\\
 &=-V_{0,K}+\gamma^{N+1}V_{N+1,K}.
\end{aligned}
\end{equation*}
Bounded rollout values and $\gamma<1$ make the last term vanish.
We can therefore replace the initial-value subtraction in
Eq.~\ref{eq:da_return_difference_start} by these value differences:
\begin{equation}
\begin{aligned}
\Delta\mathcal J_{\mathrm{DA}}
 =\mathbb E_\theta\!\left[\sum_{t=0}^\infty\gamma^t\sum_{k=1}^K
 \bigl(r_\theta(\tilde s_{t,k},a_t^{k-1})
 +\bar\gamma(k)V_{\mathrm{roll}}(\tilde s^+_{t,k})
 -V_{\mathrm{roll}}(\tilde s_{t,k})\bigr)\right].
\end{aligned}
\label{eq:da_advantage_telescoping}
\end{equation}

\daproofsubstep{2.3}{Identify the local gain and average over candidate states}{app:da_proof_return_average}
Condition on the current augmented state and action, and sample
$\tilde s^+\sim p_{\mathrm{DA}}(\cdot\mid\tilde s,a^{k-1})$.
The rollout Bellman equation uses the same augmented transition
kernel as the candidate, so
\begin{equation}
\mathbb E_{\tilde s^+\mid\tilde s,a^{k-1}}
 [r_\theta(\tilde s,a^{k-1})+\bar\gamma(k)V_{\mathrm{roll}}(\tilde s^+)]
 =Q_{\mathrm{roll}}(\tilde s,a^{k-1})-\Tau\ell_\theta(\tilde s,a^{k-1}).
\label{eq:da_local_bellman_residual}
\end{equation}
Averaging next over $a^{k-1}\sim q_\theta(\cdot\mid\tilde s)$
and subtracting $V_{\mathrm{roll}}(\tilde s)$ gives exactly
$g_\theta(\tilde s)$. Hence
\begin{equation}
\boxed{\Delta\mathcal J_{\mathrm{DA}}
 =\sum_{t=0}^\infty\gamma^t\sum_{k=1}^K
   \mathbb E_\theta[g_\theta(\tilde s_{t,k})]
 =\frac{K}{1-\gamma}
   \mathbb E_{\tilde s\sim d_{\mathrm{DA}}^{q_\theta}}
   [g_\theta(\tilde s)].}
\label{eq:da_learned_forward_pdl}
\end{equation}
The last equality is the discounted-occupancy identity in
Eq.~\ref{eq:infinite_da_occupancy_identity}.

\daproofsubstep{2.4}{Restate both gains and isolate their difference}{app:da_proof_decomposition}
Equations~\ref{eq:da_learned_forward_surrogate_gain}
and~\ref{eq:da_learned_forward_pdl}, with
Eq.~\ref{eq:da_local_gain_components}, have the same integrand:
\begin{equation*}
\begin{aligned}
\Delta\mathcal J_{\mathrm{DA}}^{\mathrm{surrogate}}
 &=\frac{K}{1-\gamma}
   \mathbb E_{\tilde s\sim d_{\mathrm{DA}}^{q_{\mathrm{roll}}}}
   [\bar A_\theta(\tilde s)-\Tau\kappa(\tilde s)+\Tau\bar G_\theta(\tilde s)],\\
\Delta\mathcal J_{\mathrm{DA}}
 &=\frac{K}{1-\gamma}
   \mathbb E_{\tilde s\sim d_{\mathrm{DA}}^{q_\theta}}
   [\bar A_\theta(\tilde s)-\Tau\kappa(\tilde s)+\Tau\bar G_\theta(\tilde s)].
\end{aligned}
\end{equation*}
Their only difference is the augmented-state weighting. The surrogate
uses rollout environment states and noisy actions; the true return uses
those visited by the candidate. Subtract the two identities and
keep the three contributions separate:
\begin{equation}
\Delta\mathcal J_{\mathrm{DA}}
 =\Delta\mathcal J_{\mathrm{DA}}^{\mathrm{surrogate}}
  +\mathcal E_A+\mathcal E_\kappa+\mathcal E_G,
\label{eq:da_learned_forward_true_surrogate_decomposition}
\end{equation}
where
\begin{equation}
\begin{aligned}
\mathcal E_A
 &:=\frac{K}{1-\gamma}
 \left(\mathbb E_{d_{\mathrm{DA}}^{q_\theta}}[\bar A_\theta]
       -\mathbb E_{d_{\mathrm{DA}}^{q_{\mathrm{roll}}}}[\bar A_\theta]\right),\\
\mathcal E_\kappa
 &:=\frac{K\Tau}{1-\gamma}
 \left(\mathbb E_{d_{\mathrm{DA}}^{q_{\mathrm{roll}}}}[\kappa]
       -\mathbb E_{d_{\mathrm{DA}}^{q_\theta}}[\kappa]\right),\\
\mathcal E_G
 &:=\frac{K\Tau}{1-\gamma}
 \left(\mathbb E_{d_{\mathrm{DA}}^{q_\theta}}[\bar G_\theta]
       -\mathbb E_{d_{\mathrm{DA}}^{q_{\mathrm{roll}}}}[\bar G_\theta]\right).
\end{aligned}
\label{eq:da_local_error_terms}
\end{equation}
The sign in $\mathcal E_\kappa$ is reversed because both gains
contain $-\Tau\kappa$. All three terms arise from changing the
state weighting; the local loss already includes both log-density
corrections. We bound the advantage and forward errors in Step~3,
then use the nonnegativity of $\kappa$ in Step~4.

\daproofstep{3}{Bound the advantage and forward-correction errors}{app:da_proof_step3}
We first bound the expected rollout advantage, then the change in
visited augmented states. Multiplying the two bounds gives the
advantage error, just as in the maximum-entropy TRPO proof. The
visitation bound also controls the forward-correction error.

\daproofsubstep{3.1}{Bound the expected rollout advantage}{app:da_proof_advantage_size}
Define the largest reverse-kernel change in total variation by
\begin{equation}
\delta_q:=\sup_{\tilde s}
 D_{\mathrm{TV}}\bigl(q_\theta(\cdot\mid\tilde s),q_{\mathrm{roll}}(\cdot\mid\tilde s)\bigr)
 =\frac12\sup_{\tilde s}\int
 |q_\theta(a^{k-1}\mid\tilde s)-q_{\mathrm{roll}}(a^{k-1}\mid\tilde s)|\,da^{k-1}.
\label{eq:da_local_tv_change}
\end{equation}
The rollout Bellman equation implies
$\mathbb E_{q_{\mathrm{roll}}}[A_{\mathrm{roll}}\mid\tilde s]=0$.
Subtract this zero from the candidate expectation to introduce the
difference between the two kernels:
\begin{equation}
\begin{aligned}
|\bar A_\theta(\tilde s)|
 &=\left|\int(q_\theta-q_{\mathrm{roll}})(a^{k-1}\mid\tilde s)
            A_{\mathrm{roll}}(\tilde s,a^{k-1})\,da^{k-1}\right|\\
 &\leq\epsilon_A\int
 |q_\theta(a^{k-1}\mid\tilde s)-q_{\mathrm{roll}}(a^{k-1}\mid\tilde s)|\,da^{k-1}
 \leq2\epsilon_A\delta_q.
\end{aligned}
\label{eq:da_expected_advantage_bound}
\end{equation}
It remains to control the distributions used to average this
quantity. An augmented state contains both an environment state
and a noisy action, so both must be included.

\daproofsubstep{3.2}{Bound the noisy-action marginals at each diffusion step}{app:da_proof_noisy_actions}
For $i\in\{\theta,\mathrm{roll}\}$, let
$q_i^{(k)}(a^k\mid s)$ be the marginal density of the noisy action
at phase $k$ when a chain is generated at environment state $s$.
These marginals satisfy
\begin{equation}
q_i^{(K)}=q_{\mathrm{prior}},
\qquad q_i^{(k-1)}(a^{k-1}\mid s)
 =\int q_i^{(k)}(a^k\mid s)q_i(a^{k-1}\mid a^k,s)\,da^k.
\label{eq:da_denoising_marginal_recursion}
\end{equation}
{Fix an environment state $s$. The $L^1$ distance
between the two noisy-action densities is their integrated absolute
difference:}
\begin{equation*}
{
\|q_\theta^{(k)}(\cdot\mid s)-q_{\mathrm{roll}}^{(k)}(\cdot\mid s)\|_1
 :=\int\!\left|q_\theta^{(k)}(a^k\mid s)
                  -q_{\mathrm{roll}}^{(k)}(a^k\mid s)\right|\,da^k.}
\end{equation*}
{The integral is over all values of $a^k$, while
$s$ remains fixed. This distance equals twice the total variation
distance and lies in $[0,2]$. We suppress the common conditioning
on $s$ in the norms below. To compare the marginals after one
denoising step, add and subtract the rollout marginal propagated
through the candidate kernel. The triangle inequality gives}
\begin{equation*}
\begin{aligned}
\|q_\theta^{(k-1)}-q_{\mathrm{roll}}^{(k-1)}\|_1
&\leq\int\!\int|q_\theta^{(k)}(a^k\mid s)-q_{\mathrm{roll}}^{(k)}(a^k\mid s)|
                 q_\theta(a^{k-1}\mid a^k,s)\,da^{k-1}da^k\\
&\quad+\int q_{\mathrm{roll}}^{(k)}(a^k\mid s)
       \int|q_\theta(a^{k-1}\mid a^k,s)-q_{\mathrm{roll}}(a^{k-1}\mid a^k,s)|
       \,da^{k-1}da^k\\
&\leq\|q_\theta^{(k)}-q_{\mathrm{roll}}^{(k)}\|_1+2\delta_q.
\end{aligned}
\end{equation*}
{In the first term, integrating the candidate kernel
over $a^{k-1}$ gives one, leaving the previous marginal distance.
In the second term, the inner integral is at most $2\delta_q$
by Eq.~\ref{eq:da_local_tv_change}; averaging it against the
normalized rollout marginal preserves this bound. Thus, each
denoising step adds at most $2\delta_q$ to the distance.}

{We now apply induction as the diffusion index
decreases from $K$ to $0$. The base case is $k=K$: both marginals
equal $q_{\mathrm{prior}}$, so their distance is zero. For the
induction step, suppose the distance at phase $k$ is at most
$2(K-k)\delta_q$. The preceding one-step inequality then gives}
\begin{equation*}
{
\|q_\theta^{(k-1)}-q_{\mathrm{roll}}^{(k-1)}\|_1
 \leq2(K-k)\delta_q+2\delta_q
 =2\bigl(K-(k-1)\bigr)\delta_q.}
\end{equation*}
{Hence, after $K-k$ denoising steps the distance
is at most $2(K-k)\delta_q$. Since $\delta_q$ bounds the kernel
change at every augmented state, this holds uniformly in $s$:}
\begin{equation}
\sup_s\|q_\theta^{(k)}(\cdot\mid s)-q_{\mathrm{roll}}^{(k)}(\cdot\mid s)\|_1
 \leq2(K-k)\delta_q,\qquad 0\leq k\leq K.
\label{eq:da_denoising_marginal_bound}
\end{equation}
In particular, the executed-action marginal at $k=0$ differs by
at most $2K\delta_q$ in $L^1$.

\daproofsubstep{3.3}{Express the discounted environment-state distribution}{app:da_proof_state_identity}
We next track how the changed executed actions affect future
environment states. The environment transition under policy $q_i$ is
\begin{equation}
P_i(s'\mid s):=\int q_i^{(0)}(a^0\mid s)p(s'\mid s,a^0)\,da^0.
\label{eq:da_induced_state_transition}
\end{equation}
Let $p_0$ be the common initial-state density and $\rho_t^i$ the
density of $s_t$ under $q_i$, so $\rho_0^i=p_0$ and
$\rho_{t+1}^i=\rho_t^iP_i$. Here
$(fP_i)(s'):=\int f(s)P_i(s'\mid s)\,ds$ denotes propagation
through one environment transition. Write
$d_i(s):=(1-\gamma)\sum_{t=0}^\infty\gamma^t\rho_t^i(s)$
for the normalized discounted state-visitation density $d^{q_i}$.
Separating its initial term and shifting the remaining index gives
\begin{equation}
\begin{aligned}
d_i
 &=(1-\gamma)p_0+(1-\gamma)\sum_{t=1}^\infty\gamma^t\rho_t^i\\
 &=(1-\gamma)p_0+\gamma
   \left[(1-\gamma)\sum_{u=0}^\infty\gamma^u\rho_u^i\right]P_i
 =(1-\gamma)p_0+\gamma d_iP_i.
\end{aligned}
\label{eq:da_occupancy_fixed_point}
\end{equation}
The bracketed quantity is the defining series for $d_i$ itself.
Subtracting the candidate and rollout identities cancels the
initial-state term. Adding and subtracting $\gamma d_{\mathrm{roll}}P_\theta$
then yields
\begin{equation}
\begin{aligned}
d_\theta-d_{\mathrm{roll}}
 &=\gamma(d_\theta P_\theta-d_{\mathrm{roll}}P_{\mathrm{roll}})\\
 &=\gamma(d_\theta-d_{\mathrm{roll}})P_\theta
   +\gamma d_{\mathrm{roll}}(P_\theta-P_{\mathrm{roll}}).
\end{aligned}
\label{eq:da_occupancy_difference_identity}
\end{equation}
The first term propagates an existing state-distribution difference;
the second introduces a difference through the changed actions.

\daproofsubstep{3.4}{Bound the environment-state difference}{app:da_proof_state_bound}
Two normalization properties bound the terms in
Eq.~\ref{eq:da_occupancy_difference_identity}. First, propagating
any signed density $f$ through a Markov kernel cannot increase its
$L^1$ norm:
\begin{equation*}
\|fP_\theta\|_1
 \leq\int\!\int|f(s)|P_\theta(s'\mid s)\,ds\,ds'
 =\int|f(s)|\,ds=\|f\|_1.
\end{equation*}
Second, integrating against the normalized environment transition
cannot increase the executed-action difference. Equation~\ref{eq:da_denoising_marginal_bound}
therefore implies
\begin{equation*}
\begin{aligned}
\int|P_\theta(s'\mid s)-P_{\mathrm{roll}}(s'\mid s)|\,ds'
 &\leq\int\!\int|q_\theta^{(0)}(a^0\mid s)-q_{\mathrm{roll}}^{(0)}(a^0\mid s)|
                p(s'\mid s,a^0)\,da^0ds'\\
 &=\|q_\theta^{(0)}(\cdot\mid s)-q_{\mathrm{roll}}^{(0)}(\cdot\mid s)\|_1
 \leq2K\delta_q.
\end{aligned}
\end{equation*}
Averaging over the normalized $d_{\mathrm{roll}}$ preserves this
bound. Taking norms in Eq.~\ref{eq:da_occupancy_difference_identity}
thus gives
\begin{equation*}
\|d_\theta-d_{\mathrm{roll}}\|_1
 \leq\gamma\|d_\theta-d_{\mathrm{roll}}\|_1+2\gamma K\delta_q.
\end{equation*}
Moving the first term to the left and dividing by $1-\gamma$ yields
\begin{equation}
\boxed{\|d_\theta-d_{\mathrm{roll}}\|_1
 \leq\frac{2\gamma K}{1-\gamma}\delta_q.}
\label{eq:da_occupancy_perturbation}
\end{equation}
This is the discounted state-visitation argument used in TRPO
\citep[Theorem~1]{schulman2015trust}, with the executed-action change
bounded through the $K$ denoising kernels.

\daproofsubstep{3.5}{Combine environment-state and noisy-action differences}{app:da_proof_augmented_bound}
The augmented-state visitation density factorizes as
\begin{equation}
d_{\mathrm{DA}}^{q_i}(s,a^k,k)
 =\frac1K d_i(s)q_i^{(k)}(a^k\mid s),\qquad k=1,\ldots,K,
\label{eq:da_augmented_occupancy_factorization}
\end{equation}
This follows directly by conditioning each
term in the defining sum for $d_{\mathrm{DA}}^{q_i}$ on $s_t$: the conditional
noisy-action marginal is $q_i^{(k)}$, independent of environment time.
The factor $1/K$ assigns equal weight to the phases.

For each phase, split the joint-density difference as
$(d_\theta-d_{\mathrm{roll}})q_\theta^{(k)}
+d_{\mathrm{roll}}(q_\theta^{(k)}-q_{\mathrm{roll}}^{(k)})$.
The triangle inequality and normalization of $q_\theta^{(k)}$ give
\begin{equation}
\begin{aligned}
\|d_{\mathrm{DA}}^{q_\theta}-d_{\mathrm{DA}}^{q_{\mathrm{roll}}}\|_1
 &\leq\|d_\theta-d_{\mathrm{roll}}\|_1
   +\frac1K\sum_{k=1}^K\int d_{\mathrm{roll}}(s)
     \|q_\theta^{(k)}(\cdot\mid s)-q_{\mathrm{roll}}^{(k)}(\cdot\mid s)\|_1\,ds\\
 &\leq\frac{2\gamma K}{1-\gamma}\delta_q
       +\frac{2}{K}\sum_{k=1}^K(K-k)\delta_q\\
 &=\left(\frac{2\gamma K}{1-\gamma}+K-1\right)\delta_q.
\end{aligned}
\label{eq:da_augmented_occupancy_perturbation}
\end{equation}
The last equality uses $\sum_{k=1}^K(K-k)=K(K-1)/2$.
The first contribution comes from future environment states;
the second comes from the noisy actions inside the current chain.
For $K=1$, the latter vanishes because the only noisy input is the
shared prior.

\daproofsubstep{3.6}{Combine the bounds and express the errors in KL}{app:da_proof_advantage_combine}
We now multiply the augmented-state distance by the magnitude of
the function being averaged. The visitation estimate from
Eq.~\ref{eq:da_augmented_occupancy_perturbation} and the definition
of $B_{K,\gamma}$ give
\begin{equation*}
\frac{K}{1-\gamma}
\|d_{\mathrm{DA}}^{q_\theta}-d_{\mathrm{DA}}^{q_{\mathrm{roll}}}\|_1
\leq B_{K,\gamma}\delta_q.
\end{equation*}
The expected advantage is also proportional to $\delta_q$ by
Eq.~\ref{eq:da_expected_advantage_bound}. Their product is therefore
proportional to $\delta_q^2$. {Using Pinsker's
inequality with natural logarithms,
$D_{\mathrm{TV}}(p,q)^2\leq\tfrac12\Dkl(p\|q)$, gives}
\begin{equation}
{\delta_q^2\leq\frac{D_q}{2},
\qquad \delta_q\leq\sqrt{D_q/2}.}
\label{eq:da_local_tv_to_kl}
\end{equation}
The KL direction here is candidate to rollout, as in the definition
of $\kappa$. Applying the inequality
$|\int f h|\leq\|f\|_1\sup|h|$ to the errors in
Eq.~\ref{eq:da_local_error_terms} yields
\begin{equation}
\begin{aligned}
|\mathcal E_A|
 &\leq\frac{K}{1-\gamma}
   \|d_{\mathrm{DA}}^{q_\theta}-d_{\mathrm{DA}}^{q_{\mathrm{roll}}}\|_1
   \sup_{\tilde s}|\bar A_\theta(\tilde s)|
 \leq2\epsilon_A B_{K,\gamma}\delta_q^2
 {\leq\epsilon_A B_{K,\gamma}D_q},\\
|\mathcal E_G|
 &\leq\frac{K\Tau}{1-\gamma}
   \|d_{\mathrm{DA}}^{q_\theta}-d_{\mathrm{DA}}^{q_{\mathrm{roll}}}\|_1
   \epsilon_G
 \leq\Tau B_{K,\gamma}\epsilon_G\delta_q
 {\leq\Tau B_{K,\gamma}\epsilon_G\sqrt{D_q/2}}.
\end{aligned}
\label{eq:da_learned_forward_occupancy_error_bound}
\end{equation}
The first line is the usual TRPO mechanism: a small policy change
limits both the expected advantage and the change in visited
states. Diffusion changes the visitation coefficient through the
intermediate noisy actions. The second line accounts for learned
forward kernels and vanishes when those kernels stay fixed.

\daproofstep{4}{Complete the KL-based return-improvement bound}{app:da_proof_step4}
We bound the remaining entropy-correction error using the
nonnegativity of the local KL, then combine it with Step~3.
This is the same final step as in the maximum-entropy TRPO proof.

\daproofsubstep{4.1}{Bound the entropy-correction error}{app:da_proof_entropy_error}
The contribution from $-\Tau\kappa$ is $\mathcal E_\kappa$ in
Eq.~\ref{eq:da_local_error_terms}. Since
$0\leq\kappa(\tilde s)\leq D_q$ at every augmented state and both
visitation distributions are normalized, both averages of
$\kappa$ lie in $[0,D_q]$. Their difference is therefore at least
$-D_q$, so
\begin{equation}
\mathcal E_\kappa
 =\frac{K\Tau}{1-\gamma}
 \left(\mathbb E_{d_{\mathrm{DA}}^{q_{\mathrm{roll}}}}[\kappa]
       -\mathbb E_{d_{\mathrm{DA}}^{q_\theta}}[\kappa]\right)
 \geq-\frac{K\Tau}{1-\gamma}D_q.
\label{eq:da_entropy_correction_bound}
\end{equation}
This bounds a difference of entropy corrections already present
in both gains. It does not introduce a new term into the actor loss.

\daproofsubstep{4.2}{Combine the advantage, entropy, and forward bounds}{app:da_proof_combine}
Insert Eqs.~\ref{eq:da_learned_forward_occupancy_error_bound}
and~\ref{eq:da_entropy_correction_bound} into the exact decomposition
from Substep~2.4:
\begin{equation}
\begin{aligned}
\Delta\mathcal J_{\mathrm{DA}}
 &=\Delta\mathcal J_{\mathrm{DA}}^{\mathrm{surrogate}}+\mathcal E_A+\mathcal E_\kappa+\mathcal E_G\\
 &\geq\Delta\mathcal J_{\mathrm{DA}}^{\mathrm{surrogate}}
       -|\mathcal E_A|-\frac{K\Tau}{1-\gamma}D_q-|\mathcal E_G|\\
 &\geq\Delta\mathcal J_{\mathrm{DA}}^{\mathrm{surrogate}}
       -\left({\epsilon_A B_{K,\gamma}}+\frac{K\Tau}{1-\gamma}\right)D_q\\
 &\quad-\Tau B_{K,\gamma}\epsilon_G{\sqrt{D_q/2}}.
\end{aligned}
\label{eq:da_learned_forward_monotonic_bound}
\end{equation}
This proves the statement at the beginning of the section. The
first penalty combines the advantage and entropy-correction
errors and is proportional to the reverse-kernel KL. The second
controls the interaction between reverse-policy change and
forward-log-density change. No supremum of $g_\theta$ remains
in the result.

\daproofsubstep{4.3}{Recover the fixed-forward maximum-entropy bound}{app:da_proof_fixed_forward}
When $\pi_\theta=\pi_{\mathrm{roll}}$, we have
$\bar G_\theta=0$ and $\epsilon_G=0$. Substituting the definition
of $B_{K,\gamma}$ makes the KL coefficient explicit:
\begin{equation}
\boxed{\begin{aligned}
\Delta\mathcal J_{\mathrm{DA}}
 &\geq\Delta\mathcal J_{\mathrm{DA}}^{\mathrm{surrogate}}\\
 &\quad-\left(\frac{{2\gamma K^2\epsilon_A}}{(1-\gamma)^2}
        +\frac{{K(K-1)\epsilon_A}+K\Tau}{1-\gamma}\right)
 D_{\mathrm{KL}}^{\max}(q_\theta\|q_{\mathrm{roll}}).
\end{aligned}}
\label{eq:da_fixed_forward_kl_bound}
\end{equation}
For $K=1$, this becomes
\begin{equation}
\Delta\mathcal J_{\mathrm{DA}}
 \geq\Delta\mathcal J_{\mathrm{DA}}^{\mathrm{surrogate}}
 -\left(\frac{{2\gamma\epsilon_A}}{(1-\gamma)^2}
        +\frac{\Tau}{1-\gamma}\right)
 D_{\mathrm{KL}}^{\max}(q_\theta\|q_{\mathrm{roll}}),
\label{eq:da_single_step_kl_bound}
\end{equation}
{which is the single-step maximum-entropy bound
with the sharper Pinsker constant.} For
$K>1$, the term proportional to $K(K-1)$ accounts for changing
noisy-action distributions within each chain; the $K^2$ term
accounts for their accumulated effect on future environment states.

\daproofsubstep{4.4}{{Sufficient Condition for Return Improvement}}{app:da_proof_interpretation}
With learned forward kernels, a sufficient condition for strict
return improvement is
\begin{equation}
\Delta\mathcal J_{\mathrm{DA}}^{\mathrm{surrogate}}
 >\left({\epsilon_A B_{K,\gamma}}+\frac{K\Tau}{1-\gamma}\right)D_q
  +\Tau B_{K,\gamma}\epsilon_G{\sqrt{D_q/2}}
\quad\Longrightarrow\quad
\Delta\mathcal J_{\mathrm{DA}}>0.
\label{eq:da_learned_forward_update_condition}
\end{equation}
For fixed forward kernels, only the KL penalty remains. In words,
the surrogate return gain must exceed the bound on its prediction
error. A small KL alone does not imply improvement, and failure
of this sufficient condition does not imply that an update is
harmful. Both the surrogate and the true return evaluate the same
local advantage and log-density corrections; the penalties bound
the effect of using rollout rather than candidate augmented states.

\endgroup

\subsection{Maximum-Entropy Wasserstein Policy Optimization}
\label{app:WPO_deriv}

This section derives ME-WPO as an alternative update for the per-state maximum-entropy surrogate objective. We first compute its corresponding $2$-Wasserstein gradient flow in policy-distribution space, then express this flow as a score-matching loss and project it onto the parametric policy family using an inverse-Fisher preconditioner.

\paragraph{Functional derivative.}
Fix a state $s$ and the rollout-frozen critic $Q^{q_{\thetar}}$. Expanding the per-state actor-target KL gives, up to a constant independent of $q$,
\begin{equation}
\mathcal L_{\mathrm{ME}}(q,s)
=
\int q(a\mid s)
\left[\log q(a\mid s)-\alpha Q^{q_{\thetar}}(s,a)\right]da
+\mathrm{const}.
\end{equation}
Its functional derivative is
\begin{equation}
\frac{\delta\mathcal L_{\mathrm{ME}}(q,s)}{\delta q}(a)
=
\log q(a\mid s)-\alpha Q^{q_{\thetar}}(s,a)+\mathrm{const}.
\label{eq:func_deriv_reverseKL}
\end{equation}
The constant contains both the target log-partition function and the $+1$ produced by differentiating $q\log q$; it disappears under the action gradient.

\paragraph{Wasserstein gradient flow.}
To keep the projection calculation compact, define the spatial residual and its gradient by
\begin{equation}
\phi_t(a,s)
:=\log q_t(a\mid s)-\alpha Q^{q_{\thetar}}(s,a),
\qquad
v_t(a,s):=\nabla_a\phi_t(a,s).
\label{eq:me_wpo_velocity_definition}
\end{equation}
The $2$-Wasserstein gradient flow that minimizes $\mathcal L_{\mathrm{ME}}$ is \citep{benamou2000computational}
\begin{equation}
\partial_t q_t(a\mid s)
=
\nabla_a\!\cdot\!\left(
q_t(a\mid s)
v_t(a,s)
\right)
\label{eqn:wasserstein_gradient_flow}
\end{equation}
Equivalently, the continuity-equation velocity is
$-v_t(a,s)$: the value term transports actions toward larger $Q^{q_{\thetar}}$, while the entropy term spreads probability mass. We write $v_\theta(a,s):=v_t(a,s)|_{q_t=q_\theta}$ below and substitute its full expression only at the end of the calculation.

\paragraph{Projection to a parametric policy.}
Define the instantaneous density change at the current policy by
$\dot q_\theta(a\mid s):=\left.\partial_tq_t(a\mid s)\right|_{q_t=q_\theta}$.
The Wasserstein flow prescribes the desired infinitesimal density
$q_\theta+\dot q_\theta\,\mathrm dt$. In contrast, a small parameter update produces
$q_{\theta+\Delta\theta}=q_\theta+(\nabla_\theta q_\theta)^\top\Delta\theta+o(\|\Delta\theta\|)$.
Because the parametric policy family generally cannot represent the Wasserstein density change exactly, we choose $\Delta\theta$ such that its induced density change best approximates $\dot q_\theta\,\mathrm dt$. The remaining approximation error is
$\dot q_\theta\,\mathrm dt-(\nabla_\theta q_\theta)^\top\Delta\theta$.
Measuring this infinitesimal residual by the local KL geometry gives the projection used by \citet{neklyudov2023wasserstein,pfau2025wasserstein}. More broadly, this is the classical information-geometric construction of projecting an infinite-dimensional density evolution onto the tangent space of a finite-dimensional statistical manifold; the local KL induces the Fisher metric \citep{amari1998natural}, as in projection-filter methods \citep{brigo1998differential,brigo1999approximate}. Concretely, the projection minimizes
\begin{align}
&\mathbb E_{s\sim d^{q_{\thetar}}}
\left[
D_{\mathrm{KL}}\!\left(
q_\theta(\cdot\mid s)
\,\middle\|\,
q_\theta(\cdot\mid s)
+\dot q_\theta(\cdot\mid s)\,\mathrm dt
-\bigl(\nabla_\theta q_\theta(\cdot\mid s)\bigr)^\top\Delta\theta
\right)
\right]
\nonumber\\[-0.25em]
&\qquad\approx
\frac{1}{2}
\begin{bmatrix}\mathrm dt\\-\Delta\theta\end{bmatrix}^{\!\top}
\begin{bmatrix}
\mathcal F_{tt} & \mathcal F_{t\theta}^{\top}\\
\mathcal F_{t\theta} & \mathcal F_{\theta\theta}
\end{bmatrix}
\begin{bmatrix}\mathrm dt\\-\Delta\theta\end{bmatrix}.
\label{eq:me_wpo_local_projection}
\end{align}
This form follows from the second-order expansion of the KL divergence, whose curvature is the Fisher information matrix. Its three blocks are
\begin{align}
\mathcal F_{tt}
&:=
\mathbb E_{\substack{s\sim d^{q_{\thetar}}\\a\sim q_\theta(\cdot\mid s)}}
\left[
\left(\frac{\dot q_\theta(a\mid s)}{q_\theta(a\mid s)}\right)^2
\right],
\nonumber\\
\mathcal F_{t\theta}
&:=
\mathbb E_{\substack{s\sim d^{q_{\thetar}}\\a\sim q_\theta(\cdot\mid s)}}
\left[
\nabla_\theta\log q_\theta(a\mid s)
\frac{\dot q_\theta(a\mid s)}{q_\theta(a\mid s)}
\right],
\nonumber\\
\mathcal F_{\theta\theta}
&:=
\mathbb E_{\substack{s\sim d^{q_{\thetar}}\\a\sim q_\theta(\cdot\mid s)}}
\left[
\nabla_\theta\log q_\theta(a\mid s)
\nabla_\theta\log q_\theta(a\mid s)^\top
\right].
\label{eq:me_wpo_fisher}
\end{align}
Here, $\mathcal F_{tt}$ measures the size of the unrestricted density flow, $\mathcal F_{t\theta}$ measures how that flow aligns with changes realizable by the policy parameters, and $\mathcal F_{\theta\theta}$ is the usual policy Fisher matrix. Minimizing Eq.~\ref{eq:me_wpo_local_projection} with respect to $\Delta\theta$ gives
\begin{equation}
\Delta\theta^\star
=
\mathcal F_{\theta\theta}^{-1}\mathcal F_{t\theta}\,\mathrm dt.
\label{eq:me_wpo_parametric_projection}
\end{equation}
\paragraph{Evaluation of the mixed Fisher block.}
Substituting the density flow from Eq.~\ref{eqn:wasserstein_gradient_flow} gives
\begin{align}
\mathcal F_{t\theta}
&=
\mathbb E_{s\sim d^{q_{\thetar}}}
\Bigg[
\int \nabla_\theta\log q_\theta(a\mid s)
\nabla_a\!\cdot\!\Bigg(
q_\theta(a\mid s)
v_\theta(a,s)
\Bigg)da
\Bigg].
\label{eq:me_wpo_mixed_block_flow}
\end{align}
For a fixed state, expanding the divergence and using
$q_\theta(a\mid s)\nabla_\theta\log q_\theta(a\mid s)=\nabla_\theta q_\theta(a\mid s)$ gives
{
\begin{equation}
  \mathcal F_{t\theta}(s)
={}
\int \nabla_\theta q_\theta(a\mid s)
\nabla_a\!\cdot v_\theta(a,s)\,da +\int \nabla_\theta\log q_\theta(a\mid s)
v_\theta(a,s)^{\!\top}
\nabla_a q_\theta(a\mid s)\,da.
  \label{eq:me_wpo_divergence_expansion}
\end{equation}
}
Integration by parts in the first term yields
{
\begin{align}
  &\int \nabla_\theta q_\theta(a\mid s)
\nabla_a\!\cdot v_\theta(a,s)\,da =
\int_{\partial\Omega}
\nabla_\theta q_\theta(a\mid s)
v_\theta(a,s)^{\!\top}n\,dS
  \nonumber\\
  &\qquad-
\int
\nabla_a\!\left(\nabla_\theta q_\theta(a\mid s)\right)
v_\theta(a,s)\,da.
  \label{eq:me_wpo_integration_by_parts}
\end{align}
}
Here, $\Omega=\mathbb R^N$, and the boundary integral vanishes when $q_\theta$ and its parameter derivatives decay sufficiently fast. Expanding
{
\begin{equation}
  \nabla_a\!\left(\nabla_\theta q_\theta(a\mid s)\right)
={}
\nabla_\theta\log q_\theta(a\mid s)
\nabla_aq_\theta(a\mid s)^\top +q_\theta(a\mid s)
\nabla_\theta\nabla_a\log q_\theta(a\mid s)
  \label{eq:me_wpo_score_product_rule}
\end{equation}
}
shows that the first term on the right cancels the second integral in Eq.~\ref{eq:me_wpo_divergence_expansion}. Consequently,
\begin{align}
\mathcal F_{t\theta}
={}&-
\mathbb E_{\substack{s\sim d^{q_{\thetar}}\\
a\sim q_\theta(\cdot\mid s)}}
\Bigg[
\nabla_\theta\nabla_a\log q_\theta(a\mid s)
v_\theta(a,s)
\Bigg].
\label{eq:flow_identity}
\end{align}
Substituting $v_\theta(a,s)=\nabla_a[\log q_\theta(a\mid s)-\alpha Q^{q_{\thetar}}(s,a)]$ recovers the mixed-derivative identity used by WPO. In the corresponding first-order surrogate, $\nabla_\theta$ acts only on the policy score $\nabla_a\log q_\theta(a\mid s)$; the sampled action, critic, and score residual multiplying this mixed derivative are held fixed.

Multiplying the reverse-KL objective by $\Tau$ gives the maximum-entropy RL objective used in this paper. Its detached-sample Fisher-divergence surrogate is
\begin{equation}
\widehat{\mathcal L}_{\mathrm{ME\text{-}WPO}}(\theta)
:=
\frac{\Tau}{2}
\mathbb E_{\substack{s\sim d^{q_{\thetar}}\\
a\sim q_{\thetastop}(\cdot\mid s)}}
\left[
\left\|
\nabla_a\!\left(
\log q_\theta(a\mid s)
-\frac{1}{\Tau}Q^{q_{\thetar}}(s,a)
\right)
\right\|^2
\right].
\label{eq:wpo_surrogate_rl}
\end{equation}
Indeed, holding both the sampled action and the critic fixed gives
\begin{equation}
\nabla_\theta
\widehat{\mathcal L}_{\mathrm{ME\text{-}WPO}}(\theta)
=
\mathbb E_{\substack{s\sim d^{q_{\thetar}}\\
a\sim q_{\thetastop}(\cdot\mid s)}}
\Bigg[
\nabla_\theta\nabla_a\log q_\theta(a\mid s)
\,
\nabla_a\!\left(
\Tau\log q_\theta(a\mid s)-Q^{q_{\thetar}}(s,a)
\right)
\Bigg].
\label{eq:me_wpo_parameter_gradient}
\end{equation}
The exact continuous-time projected parameter flow is therefore
\begin{equation}
\dot\theta
=
-\mathcal F_{\theta\theta}^{-1}
\nabla_\theta
\widehat{\mathcal L}_{\mathrm{ME\text{-}WPO}}(\theta).
\label{eq:reverseKL_wpo_update}
\end{equation}
The inverse Fisher may be computed exactly for small policy classes or approximated by a diagonal, block-diagonal, or structured preconditioner. Omitting the preconditioner recovers ordinary gradient descent on the same Fisher-divergence objective.

\paragraph{Diagonal-Gaussian specialization.}
For the Gaussian policy used by ME-WPO, write
\begin{equation}
q_\theta(a\mid s)
=\mathcal N\!\left(
a\,;\,\mu_\theta(s),
\operatorname{diag}\!\left(\sigma_\theta(s)^2\right)
\right),
\label{eq:me_wpo_gaussian_policy_derivation}
\end{equation}
Its action score and its scores in the local distribution coordinates $(\mu,\sigma)$ are
\begin{align}
\nabla_a\log q_\theta(a\mid s)
&=-\operatorname{diag}\!\left(\sigma_\theta(s)^{-2}\right)
\left(a-\mu_\theta(s)\right),
\label{eq:me_wpo_gaussian_action_score}\\
\nabla_\mu\log q_\theta(a\mid s)
&=\operatorname{diag}\!\left(\sigma_\theta(s)^{-2}\right)
\left(a-\mu_\theta(s)\right),
\nonumber\\
\nabla_\sigma\log q_\theta(a\mid s)
&=-\sigma_\theta(s)^{-1}
+\left(a-\mu_\theta(s)\right)^{\odot 2}
\odot\sigma_\theta(s)^{\odot(-3)},
\label{eq:me_wpo_gaussian_parameter_scores}
\end{align}
where powers and products involving vectors are element-wise. Consequently, the score residual appearing in Eq.~\ref{eq:wpo_surrogate_rl} becomes
\begin{equation}
\nabla_a\!\left(
\log q_\theta(a\mid s)-\frac{1}{\Tau}Q^{q_{\thetar}}(s,a)
\right)
=-\operatorname{diag}\!\left(\sigma_\theta(s)^{-2}\right)
\left(a-\mu_\theta(s)\right)
-\frac{1}{\Tau}\nabla_aQ^{q_{\thetar}}(s,a).
\label{eq:me_wpo_gaussian_residual}
\end{equation}
The negative of this residual is the continuity-equation velocity for the unscaled reverse KL. Thus, the policy-score contribution spreads samples away from high-density regions, whereas the critic contribution moves them toward increasing value.

At a fixed state, the exact Fisher matrix in the output coordinates $(\mu,\sigma)$ is block diagonal:
{\small
\begin{equation}
\mathcal F_{(\mu,\sigma)}
=
\begin{pmatrix}
\operatorname{diag}(\sigma_\theta(s)^{-2}) & 0\\
0 & 2\operatorname{diag}(\sigma_\theta(s)^{-2})
\end{pmatrix},
\qquad
\mathcal F_{(\mu,\sigma)}^{-1}
=
\begin{pmatrix}
\operatorname{diag}(\sigma_\theta(s)^2) & 0\\
0 & \tfrac12\operatorname{diag}(\sigma_\theta(s)^2)
\end{pmatrix}.
\label{eq:me_wpo_gaussian_fisher_derivation}
\end{equation}
}
The corresponding preconditioned output-coordinate gradients are therefore
\begin{equation}
\widetilde\nabla_{\mu}\widehat{\mathcal L}_{\mathrm{ME\text{-}WPO}}
=\sigma_\theta(s)^2\odot
\nabla_{\mu}\widehat{\mathcal L}_{\mathrm{ME\text{-}WPO}},
\qquad
\widetilde\nabla_{\sigma}\widehat{\mathcal L}_{\mathrm{ME\text{-}WPO}}
=\frac12\sigma_\theta(s)^2\odot
\nabla_{\sigma}\widehat{\mathcal L}_{\mathrm{ME\text{-}WPO}}.
\label{eq:me_wpo_gaussian_preconditioner_derivation}
\end{equation}
For a neural policy, these exact local Gaussian blocks provide the tractable natural-gradient approximation of \citet{pfau2025wasserstein} to the full parameter Fisher.

\subsection{Derivation of Diffusion-Augmented Wasserstein Policy Optimization}
\label{app:diff_WPO_derivation}

We now derive the \dawpo{} objective for a DA-MDP transition, first for a learnable reverse diffusion kernel and then for a learnable forward diffusion kernel. Fix an augmented state $\tilde s_{\tilde t}=(s_t,a_t^k,k)$ and the rollout-frozen critic $Q_{\mathrm{DA}}^{q_{\thetar}}$. To avoid carrying the full DA-MDP notation through every intermediate step, define
\begin{equation}
\begin{aligned}
x&:=a_t^{k-1},
&q_\theta(x)&:=q_\theta(a_t^{k-1}\mid a_t^k,s_t),\\[-0.1em]
\pi_\theta(x)&:=\pi_\theta(a_t^k\mid a_t^{k-1},s_t),
&Q(x)&:=Q_{\mathrm{DA}}^{q_{\thetar}}(\tilde s_{\tilde t},a_t^{k-1}).
\end{aligned}
\label{eq:da_wpo_compact_notation}
\end{equation}
{We derive the updates by transporting the reverse density and the local forward-target shape at fixed critic and temperature, then projecting these density-space flows onto the kernel parameters.} Assuming $0<Z_\theta<\infty$, we define
\begin{equation}
\widetilde p_\theta(x)
:=\pi_\theta(x)\exp\!\left(\alpha Q(x)\right),
\qquad
Z_\theta:=\int\widetilde p_\theta(x)\,dx,
\qquad
\bar p_\theta(x):=\frac{\widetilde p_\theta(x)}{Z_\theta}.
\label{eq:da_wpo_local_target}
\end{equation}
With $\alpha=1/\Tau$, the {normalized functional} is
{
\begin{equation}
  \begin{aligned}
  \mathcal E_{\tilde s_{\tilde t}}(q_\theta,\bar p_\theta)
&:=\Tau D_{\mathrm{KL}}\!\left(
q_\theta\,\middle\|\,\bar p_\theta
\right) =\Tau\int q_\theta(x)
\log\frac{q_\theta(x)}{\bar p_\theta(x)}\,dx\\
  &=\int q_\theta(x)
\left[
\Tau\log\frac{q_\theta(x)}{\pi_\theta(x)}-Q(x)
\right]dx
+\Tau\log Z_\theta.
  \end{aligned}
  \label{eq:da_wpo_unnormalized_functional}
\end{equation}
}
{
To distinguish normalization from the choice of transport geometry, for a positive density $\rho$ with $Z[\rho]:=\int\rho(x)\,dx\in(0,\infty)$ define
\begin{equation}
\begin{aligned}
\mathcal F^{\mathrm u}(q,\rho)
&:=\Tau\int q(x)\log\frac{q(x)}{\rho(x)}\,dx,\\
\mathcal G(q,\rho)
&:=\Tau D_{\mathrm{KL}}\!\left(q\,\middle\|\,\rho/Z[\rho]\right)
=\mathcal F^{\mathrm u}(q,\rho)+\Tau\log Z[\rho].
\end{aligned}
\label{eq:da_wpo_normalized_unnormalized_relation}
\end{equation}
For $\rho=\widetilde p_\theta$, $\mathcal F^{\mathrm u}$ is the local unnormalized functional in Eq.~\ref{eq:diffusion_surrogate}, and $\mathcal G$ equals $\mathcal E_{\tilde s_{\tilde t}}$ above. Normalization leaves the target's spatial score and its mixed spatial--parameter derivative unchanged:
\begin{equation}
\nabla_x\log\bar p_\theta=\nabla_x\log\widetilde p_\theta,
\qquad
\nabla_\theta\nabla_x\log\bar p_\theta
=\nabla_\theta\nabla_x\log\widetilde p_\theta.
\label{eq:da_wpo_normalized_score_invariance}
\end{equation}
Because $Z_\theta$ is independent of $x$, it disappears from the spatial derivatives above.
}

\subsubsection{Learnable Reverse Diffusion Kernel}
We first vary $q$ while holding $\bar p$ fixed. For $q_\epsilon=q+\epsilon\eta$ with $\int\eta(x)dx=0$,
\begin{equation}
\left.\frac{d}{d\epsilon}
\mathcal E_{\tilde s_{\tilde t}}(q_\epsilon,\bar p)
\right|_{\epsilon=0}
=\Tau\int
\left(\log\frac{q(x)}{\bar p(x)}+1\right)
\eta(x)dx.
\label{eq:da_wpo_reverse_variation}
\end{equation}
Thus, the functional derivative of Eq.~\ref{eq:da_wpo_unnormalized_functional} is
\begin{equation}
\frac{\delta\mathcal E_{\tilde s_{\tilde t}}}{\delta q}
(x)
=\Tau\log\frac{q(x)}{\bar p(x)}+\mathrm{const}.
\label{eq:da_wpo_reverse_first_variation}
\end{equation}
Hence its minimizing $2$-Wasserstein flow, with $\lambda$ denoting flow time, is
\begin{equation}
\partial_\lambda q_\lambda(x)
=\Tau\nabla_x\!\cdot\!\left[
q_\lambda(x)
\nabla_x\log\frac{q_\lambda(x)}{\bar p_{\thetastop}(x)}
\right].
\label{eq:da_wpo_reverse_flow}
\end{equation}
For compactness, define the reverse-kernel score residual
\begin{equation}
r_\theta^q(x)
:=\nabla_x\log\frac{q_\theta(x)}{\bar p_{\thetastop}(x)}.
\label{eq:da_wpo_reverse_residual}
\end{equation}
The residual is available without evaluating the normalizer because
\begin{equation}
{\small
\begin{aligned}
\nabla_x\log\frac{q_\theta(x)}{\bar p_{\thetastop}(x)}
&=\nabla_x\!\left[\log q_\theta(x)-\log\pi_{\thetastop}(x)
-\alpha Q(x)+\log Z_{\thetastop}\right]\\
&=\nabla_x\!\left[\log q_\theta(x)-\log\pi_{\thetastop}(x)-\alpha Q(x)\right].
\end{aligned}
}
\label{eq:da_wpo_reverse_normalizer_cancellation}
\end{equation}
Applying the projection in Eqs.~\ref{eq:me_wpo_local_projection}--\ref{eq:me_wpo_parametric_projection}, and using the integration-by-parts identity in Eq.~\ref{eq:flow_identity}, gives
\begin{align}
\mathcal F_{\lambda\theta}^{q}
&:=\int
\nabla_\theta\log q_\theta(x)
\left.\partial_\lambda q_\lambda(x)
\right|_{q_\lambda=q_\theta}
dx
\nonumber\\
&=-\Tau
\mathbb E_{x\sim q_{\thetastop}}
\left[
\nabla_\theta\nabla_x\log q_\theta(x)\,
r_\theta^q(x)
\right]
=-\nabla_\theta\mathcal L_q(\theta).
\label{eq:da_wpo_reverse_mixed_block}
\end{align}
Restoring the DA-MDP variables, the projected reverse-kernel flow is therefore represented by
\begin{equation}
\mathcal L_q(\theta)
:=\frac{\Tau}{2}
\mathbb E_{a_t^{k-1}\sim q_{\thetastop}(\cdot\mid a_t^k,s_t)}
\left[
\left\|
\nabla_{a_t^{k-1}}
\left(
\log\frac{q_\theta(a_t^{k-1}\mid a_t^k,s_t)}
{\pi_{\thetastop}(a_t^k\mid a_t^{k-1},s_t)}
-\alpha Q_{\mathrm{DA}}^{q_{\thetar}}
(\tilde s_{\tilde t},a_t^{k-1})
\right)
\right\|^2
\right].
\label{eq:da_wpo_reverse_loss}
\end{equation}
Here $\nabla_\theta$ acts only on $\nabla_x\log q_\theta(x)$; the forward target, sampled $x$, and critic are detached.

\subsubsection{Learnable Forward Diffusion Kernel}

\paragraph{Normalization preserves the Wasserstein flow.}
{
\textbf{Eq.~\ref{eq:da_wpo_normalized_score_invariance}} establishes that normalization leaves the target's spatial score and its mixed spatial--parameter derivative unchanged. For the forward-target flow, the corresponding invariance follows from the functional-derivative calculation below.

We hold $q$, the critic, and the temperature fixed and consider mass-preserving spatial transport of the forward target. Assume sufficient regularity, finite second moments, and vanishing boundary flux for the Wasserstein flows and integrations by parts below. Comparing the two functionals on the same density variable $\rho$ and with the same transport metric gives
\begin{equation}
\begin{aligned}
\frac{\delta\mathcal F^{\mathrm u}}{\delta\rho}(x)
&=-\Tau\frac{q(x)}{\rho(x)},
&\frac{\delta\mathcal G}{\delta\rho}(x)
&=-\Tau\frac{q(x)}{\rho(x)}+\frac{\Tau}{Z[\rho]},\\
\nabla_x\frac{\delta\mathcal F^{\mathrm u}}{\delta\rho}
&=\nabla_x\frac{\delta\mathcal G}{\delta\rho}.
\end{aligned}
\label{eq:da_wpo_forward_normalization_invariance}
\end{equation}
The added first-variation term is spatially constant. Consequently, both functionals induce the same flow $\partial_\lambda\rho=\nabla_x\!\cdot(\rho\nabla_x(\delta\mathcal F^{\mathrm u}/\delta\rho))$, which preserves $\int\rho\,dx$ under the stated boundary condition. Thus, for the following calculation, we use the normalized target $\bar p_\theta=\widetilde p_\theta/Z_\theta$ and its Wasserstein metric.
The normalizer affects ordinary parameter differentiation, as discussed in App.~\ref{app:learned_forward_kernel_correction}, but does not affect the mass-preserving spatial flows considered here. The resulting projected transport update is therefore distinct from the ordinary parameter gradient of the local objective.
}

\paragraph{Derivation.}
We now vary the normalized target through $\pi_\theta$. For $\bar p_\epsilon=\bar p+\epsilon\eta$ with $\int\eta(x)\,dx=0$, Eq.~\ref{eq:da_wpo_unnormalized_functional} gives
\begin{equation}
\left.\frac{d}{d\epsilon}
\mathcal E_{\tilde s_{\tilde t}}(q_{\thetastop},\bar p_\epsilon)
\right|_{\epsilon=0}
=-\Tau\int
q_{\thetastop}(x)
\frac{\eta(x)}{\bar p(x)}dx.
\label{eq:da_wpo_forward_variation}
\end{equation}
Therefore,
\begin{equation}
\frac{\delta\mathcal E_{\tilde s_{\tilde t}}}
{\delta\bar p}(x)
=-\Tau\frac{q_{\thetastop}(x)}{\bar p(x)}.
\label{eq:da_wpo_forward_first_variation}
\end{equation}
The corresponding spatial transport of the target shape is
\begin{equation}
\partial_\lambda\bar p_\lambda(x)
=-\Tau\nabla_x\!\cdot\!\left[
\bar p_\lambda(x)
\nabla_x\frac{q_{\thetastop}(x)}{\bar p_\lambda(x)}
\right].
\label{eq:forward_flow}
\end{equation}
Equivalently, defining
\begin{equation}
v_\lambda^\pi(x)
:=\Tau\nabla_x\frac{q_{\thetastop}(x)}{\bar p_\lambda(x)},
\qquad
\partial_\lambda\bar p_\lambda
+\nabla_x\!\cdot(\bar p_\lambda v_\lambda^\pi)=0,
\label{eq:da_wpo_forward_velocity}
\end{equation}
makes explicit that the forward-target shape is transported by the velocity $v_\lambda^\pi$. Because the critic is frozen,
\begin{equation}
\nabla_\theta\nabla_x\log\bar p_\theta(x)
=\nabla_\theta\nabla_x\!\left[
\log\pi_\theta(x)+\alpha Q(x)-\log Z_\theta
\right]
=\nabla_\theta\nabla_x\log\pi_\theta(x).
\label{eq:da_wpo_forward_target_mixed_derivative}
\end{equation}
The middle equality displays the cancellation directly: although $\nabla_\theta\log Z_\theta$ need not vanish, its derivative with respect to $x$ is zero.
Define the forward-kernel residual by
\begin{equation}
r_\theta^\pi(x)
:=\nabla_x\log\frac{q_{\thetastop}(x)}{\bar p_\theta(x)}.
\label{eq:da_wpo_forward_residual}
\end{equation}
We can now project Eq.~\ref{eq:forward_flow} onto the forward-kernel parameter block. Substitution of the flow, one integration by parts, and the identity
$\bar p\,\nabla_x(q/\bar p)=q\,\nabla_x\log(q/\bar p)$ give the full mixed-block calculation
\begin{align}
\mathcal F_{\lambda\theta}^{\pi}
&:=\int
\nabla_\theta\log\bar p_\theta(x)
\left.\partial_\lambda\bar p_\lambda(x)
\right|_{\bar p_\lambda=\bar p_{\thetastop}}
dx
\nonumber\\
&=-\Tau\int
\nabla_\theta\log\bar p_\theta(x)
\nabla_x\!\cdot\!\left[
\bar p_{\thetastop}(x)
\nabla_x\frac{q_{\thetastop}(x)}{\bar p_{\thetastop}(x)}
\right]dx
\nonumber\\
&=\Tau\int
\bar p_{\thetastop}(x)
\nabla_\theta\nabla_x\log\bar p_\theta(x)
\nabla_x\frac{q_{\thetastop}(x)}{\bar p_{\thetastop}(x)}dx
\nonumber\\
&=\Tau\int q_{\thetastop}(x)
\nabla_\theta\nabla_x\log\pi_\theta(x)
\nabla_x\log\frac{q_{\thetastop}(x)}{\bar p_{\thetastop}(x)}dx
\nonumber\\
&=\Tau\mathbb E_{x\sim q_{\thetastop}}
\left[
\nabla_\theta\nabla_x\log\pi_\theta(x)\,
r_\theta^\pi(x)
\right]
=-\nabla_\theta\mathcal L_\pi(\theta).
\label{eq:da_wpo_forward_mixed_block}
\end{align}
The last equality follows because $q_{\thetastop}$ is detached and $\nabla_\theta r_\theta^\pi=-\nabla_\theta\nabla_x\log\pi_\theta$. Restoring the DA-MDP variables, the forward-kernel loss is
\begin{equation}
\mathcal L_\pi(\theta)
:=\frac{\Tau}{2}
\mathbb E_{a_t^{k-1}\sim q_{\thetastop}(\cdot\mid a_t^k,s_t)}
\left[
\left\|
\nabla_{a_t^{k-1}}
\left(
\log\frac{q_{\thetastop}(a_t^{k-1}\mid a_t^k,s_t)}
{\pi_\theta(a_t^k\mid a_t^{k-1},s_t)}
-\alpha Q_{\mathrm{DA}}^{q_{\thetar}}
(\tilde s_{\tilde t},a_t^{k-1})
\right)
\right\|^2
\right].
\label{eq:da_wpo_forward_loss}
\end{equation}
Here $\nabla_\theta$ acts only on $\nabla_x\log\pi_\theta(x)$; the reverse-kernel sample and score and the critic gradient are held fixed.

\vspace{\runinheadingspace}\noindent\textbf{Why the parameter blocks are disjoint.}
The two variations above are independent for the sampler in App.~\ref{app:vp_sde}. Although both kernels use the global parameter vector $\theta=(\Psi,\beta_K,\ldots,\beta_0)$, for the local transition pair at reverse step $k$, $q_\theta(x)$ depends only on $(\Psi,\beta_k)$, whereas $\pi_\theta(x)$ depends only on $\beta_{k-1}$. Hence, the paired kernels have no shared learnable coordinate: varying the reverse kernel leaves the paired forward kernel fixed, and vice versa. The same coefficient can occur in the neighboring transition pair, and the sum over all $(t,k)$ automatically accumulates its reverse- and forward-kernel contributions. If a paired forward and reverse kernel shared a parameter, the two variations would be coupled and the projection would require the corresponding cross-Fisher blocks; the independent derivations above could then not simply be combined.

\vspace{\runinheadingspace}\noindent\textbf{Combining the forward- and reverse-kernel gradients.}
Because the reverse- and forward-kernel gradients have disjoint support in $\theta$, Eqs.~\ref{eq:da_wpo_reverse_mixed_block} and~\ref{eq:da_wpo_forward_mixed_block} combine as
\begin{equation}
\nabla_\theta\mathcal L_{\mathrm{DA:WPO}}(\theta)
=
\nabla_\theta\mathcal L_q(\theta)
+
\nabla_\theta\mathcal L_\pi(\theta),
\label{eq:da_wpo_combined_block_gradient}
\end{equation}
where each derivative acts only through its respective kernel score. Because those parameter blocks are disjoint, the two gradients are also obtained by differentiating the single local loss
{\eqsizesmall
\begin{equation}
\mathcal L_{\mathrm{DA:WPO}}(\theta\mid\tilde s_{\tilde t})
=\frac{\Tau}{2}
\mathbb E_{a_t^{k-1}\sim q_{\thetastop}(\cdot\mid a_t^k,s_t)}
\left[
\left\|
\nabla_{a_t^{k-1}}
\left(
\log\frac{q_\theta(a_t^{k-1}\mid a_t^k,s_t)}
{\pi_\theta(a_t^k\mid a_t^{k-1},s_t)}
-\alpha Q_{\mathrm{DA}}^{q_{\thetar}}
(\tilde s_{\tilde t},a_t^{k-1})
\right)
\right\|^2
\right].
\label{eq:da_wpo_compact_combined_loss}
\end{equation}
}
Finally, averaging the local loss over rollout augmented states gives
\begin{equation}
\mathcal L_{\mathrm{DA:WPO}}(\theta)
=
\mathbb E_{\tilde s_{\tilde t}\sim d_{\mathrm{DA}}^{q_{\thetar}}}
\left[
\mathcal L_{\mathrm{DA:WPO}}(\theta\mid\tilde s_{\tilde t})
\right].
\label{eq:da_wpo_appendix_loss}
\end{equation}
This is Eq.~\ref{eq:diffmaxent_wpo_main}. With the sampled action and critic held fixed, its gradient is
{\eqsizetiny
\begin{align}
\nabla_\theta\mathcal L_{\mathrm{DA:WPO}}(\theta)
={}&
\mathbb E_{\substack{
\tilde s_{\tilde t}\sim d_{\mathrm{DA}}^{q_{\thetar}}\\
a_t^{k-1}\sim q_{\thetastop}(\cdot\mid a_t^k,s_t)}}
\Bigg[
\left(
\nabla_\theta\nabla_{a_t^{k-1}}
\log\frac{q_\theta(a_t^{k-1}\mid a_t^k,s_t)}
{\pi_\theta(a_t^k\mid a_t^{k-1},s_t)}
\right)
\nonumber\\[-2pt]
&\hspace{7mm}\times
\nabla_{a_t^{k-1}}
\left(
\Tau\log\frac{q_\theta(a_t^{k-1}\mid a_t^k,s_t)}
{\pi_\theta(a_t^k\mid a_t^{k-1},s_t)}
-Q_{\mathrm{DA}}^{q_{\thetar}}
(\tilde s_{\tilde t},a_t^{k-1})
\right)
\Bigg].
\label{eq:da_wpo_derivation_gradient}
\end{align}
}
The first factor contains the reverse-score derivative and the negative forward-score derivative, so Eq.~\ref{eq:da_wpo_derivation_gradient} combines both projected kernel gradients without differentiating through the complete diffusion chain.

\vspace{\runinheadingspace}\noindent\textbf{Parametric projection.}
The inverse-Fisher projection is obtained by applying Eqs.~\ref{eq:me_wpo_local_projection}--\ref{eq:me_wpo_parametric_projection} separately to the disjoint parameter blocks of $q_\theta(a_t^{k-1}\mid a_t^k,s_t)$ and $\pi_\theta(a_t^k\mid a_t^{k-1},s_t)$. In practice, constructing a Fisher matrix over all neural-network parameters would be prohibitively expensive. We therefore use the adjusted variance-preserving Gaussian preconditioner derived in Eqs.~\ref{eq:da_wpo_vp_fisher_block}--\ref{eq:da_wpo_adjusted_vp_gradient}; without this option, \dawpo{} applies the ordinary gradient in Eq.~\ref{eq:da_wpo_derivation_gradient}.

\clearpage
\section{{Common Hyperparameters}}
\label{app:common_hyperparameters}

{Tables~\ref{tab:common_mujoco_hyperparameters} and~\ref{tab:common_maniskill_hyperparameters} summarize the common training configurations for MuJoCo and ManiSkill. Exact configurations for each experiment, including task-specific settings and ablation overrides, are provided with the accompanying code.}

\subsection{{MuJoCo}}
\label{app:common_mujoco_hyperparameters}

{Table~\ref{tab:common_mujoco_hyperparameters} records the common JAX continuous-control presets. The PyTorch action-transform and G1 diffusion-step experiments use the settings specified in Apps.~\ref{app:action_transform_entropy_ablation} and~\ref{app:diffusion_step_details}, respectively.}

\begin{table}[H]
    \centering
    \small
    \caption{{Common MuJoCo hyperparameters for the JAX training presets.}}
    \label{tab:common_mujoco_hyperparameters}
    {\begin{tabular}{lccc}
        \toprule
        Parameter & \dappo{} & \dareppo{} & \dawpo{} \\
        \midrule
        \multicolumn{4}{l}{\textbf{Rollout and training budget}} \\
        Environment interactions & \multicolumn{3}{c}{$50$M} \\
        Parallel environments & $2048$ & $1024$ & $1024$ \\
        Environment steps per rollout, per env. & \multicolumn{3}{c}{$64$} \\
        Local transitions per rollout, per env. & \multicolumn{3}{c}{$512$} \\
        Minibatches per epoch & \multicolumn{3}{c}{$64$} \\
        Epochs per rollout & $12$ & $8$ & $8$ \\
        Discount $\gamma$ / trace parameter $\lambda$ & \multicolumn{3}{c}{$0.999/0.98$} \\
        Observation / reward normalization & \multicolumn{3}{c}{Yes / no} \\
        \midrule
        \multicolumn{4}{l}{\textbf{Diffusion policy}} \\
        Control MLP hidden layers $\times$ width & \multicolumn{3}{c}{$3\times256$} \\
        Activation / control LayerNorm & \multicolumn{3}{c}{GeLU / no} \\
        Time-encoder hidden / output width & \multicolumn{3}{c}{$32/16$} \\
        Diffusion steps $K$ / prior std. & \multicolumn{3}{c}{$8/3.0$} \\
        Base time step / initial friction & \multicolumn{3}{c}{$0.125/0.25$} \\
        Learned per-step time increments & \multicolumn{3}{c}{Yes} \\
        Learned friction & Yes & Yes & No \\
        Environment-action clipping bound & $1.0$ & $0.999$ & $1.0$ \\
        \midrule
        \multicolumn{4}{l}{\textbf{Critic}} \\
        Critic target & GAE ($V$) & TD-$\lambda$ ($Q$) & TD-$\lambda$ ($Q$) \\
        Hidden width / LayerNorm & \multicolumn{3}{c}{$512$ / yes} \\
        Encoder / head linear layers & \multicolumn{3}{c}{$2/2$} \\
        Activation / loss & \multicolumn{3}{c}{SiLU / HL-Gauss} \\
        HL-Gauss bins & $151$ & $151$ & $301$ \\
        Critic support & \multicolumn{3}{c}{$[-100,200]$} \\
        Auxiliary-loss weight & $0.35$ & $0.10$ & $0.05$ \\
        \midrule
        \multicolumn{4}{l}{\textbf{Optimization and regularization}} \\
        Optimizer / learning-rate schedule & \multicolumn{3}{c}{Adam / constant} \\
        Actor and critic learning rate & $10^{-4}$ & $10^{-3}$ & $10^{-3}$ \\
        Maximum gradient norm & \multicolumn{3}{c}{$0.5$} \\
        PPO clipping $\epsilon$ / chain-KL bound & $0.2$ & $0.04$ & $0.02$ \\
        Temperature & Fixed: $2\times10^{-4}$ & Learned & Learned \\
        Initial learned temperature & --- & $0.01$ & $0.01$ \\
        Target-entropy factor $c$ & --- & $-5$ & $-8$ \\
        Temperature / KL-dual learning rate & --- & $3\times10^{-4}$ & $6\times10^{-4}$ \\
        Initial KL multiplier & --- & $0.3$ & $0.3$ \\
        \bottomrule
    \end{tabular}}
\end{table}

\noindent {The effective rollout length is $128\times\tfrac12=64$ environment steps, and the configured eight minibatches become $8K=64$ partitions of local transitions. For entropy tuning, $\mathcal H_{\mathrm{tar}}=\mathcal H[q_{\mathrm{prior}}]+cd$, where $d$ is the action dimension. The presets request approximately $50$ periodic evaluations over training; the continuous-control comparison uses the evaluation curves described in App.~\ref{app:continuous_control_protocol}.}

\clearpage
\subsection{{PushT Hyperparameters}}
\label{app:common_maniskill_hyperparameters}

{Table~\ref{tab:common_maniskill_hyperparameters} records the DA:REPPO PushT configuration in Fig.~\ref{fig:pusht_motion_and_metrics}; the Gaussian baseline and matched evaluation protocol are described in App.~\ref{app:pusht_multimodality}.}
\begin{table}[H]
    \centering
    \small
    \caption{{Hyperparameters of the unchunked DA:REPPO PushT experiment.}}
    \label{tab:common_maniskill_hyperparameters}
    {\begin{tabular}{@{}ll@{}}
        \toprule
        Parameter & Value\\
        \midrule
        Parallel environments / policy steps per rollout & $1024/128$\\
        {Training budget / episode horizon} & {$250.35$M interactions / $100$ steps}\\
        Action chunk size / diffusion steps & $1/8$\\
        Epochs / minibatches per epoch & $8/64$\\
        Discount / trace parameter & $0.9/0.98$\\
        Actor / critic hidden dimensions & $(512,512,512)/(512,512)$\\
        Actor architecture / activation & Residual MLP + LayerNorm / GeLU\\
        Time-embedding frequencies & $32$\\
        Prior standard deviation / base time step & $2.5/0.125$\\
        Time grid / learned friction & Cosine / yes\\
        Learned time increments / noise scale & No / no\\
        Action transform & $\tanh$\\
        Critic loss / bins / support & HL-Gauss / $301$ / $[-10,15]$\\
        Auxiliary-loss weight & $0.05$\\
        Optimizer / network learning rate & Adam / $3\cdot10^{-4}$\\
        Maximum gradient norm / chain-KL bound & $0.5/0.08$\\
        Initial temperature / entropy factor & $0.01/-4$\\
        Entropy-reference standard deviation & $3.0$\\
        Temperature / KL-dual learning rate & $2.12\cdot10^{-4}$ each\\
        Initial KL multiplier / forward-kernel constraint & $0.3$ / disabled\\
        Observation / running reward normalization & Yes / no\\
        \bottomrule
    \end{tabular}}
\end{table}

\noindent{Training uses the task's normalized dense reward and bootstraps from successful final observations. The entropy target is $d[\tfrac12\log(2\pi e\,3^2)-4]$, with $d$ the action dimension. Its reference standard deviation differs from the sampling prior's $2.5$. The evaluation protocol and results are specified in App.~\ref{app:pusht_multimodality}.}

\end{document}